\documentclass[10pt,journal,compsoc]{IEEEtran}

\usepackage{orcidlink}
\usepackage{listings}
\usepackage[utf8]{inputenc} 
\usepackage[T1]{fontenc}    
\usepackage{hyperref}       
\usepackage{url}            
\usepackage{booktabs}       
\usepackage{amsfonts}       
\usepackage{nicefrac}       
\usepackage{microtype}      
\usepackage{amsmath,amsthm,amssymb}
\usepackage{bbold}
\usepackage{fancyvrb} 
\usepackage{float}  
\VerbatimFootnotes    
 
\usepackage[numbers]{natbib}

\usepackage{graphicx}
\usepackage{color}
\usepackage{comment,color}
\usepackage{bm}
\usepackage{wrapfig}
\usepackage{adjustbox}
\usepackage{array}

\usepackage{newtxtext}
\usepackage{newtxmath}

\usepackage{bookmark}

\usepackage{caption}
\usepackage{subcaption}
\usepackage{float}  

\usepackage{autobreak}
\usepackage{mathtools}

\usepackage{tabularx}

\usepackage{algorithm}
\usepackage{algpseudocode}
\usepackage{booktabs} 

\usepackage{placeins} 

\algdef{SE}[DOWHILE]{Do}{doWhile}{\algorithmicdo}[1]{\algorithmicwhile\ #1}%

\makeatletter
\newcommand{\algohline}{\par\nobreak\kern2pt\noindent\hrulefill\kern2pt\par\nobreak}
\makeatother

\newcommand{\SC}{ T }

\newtheorem{example}{Example}

\newtheorem{proposition}{Proposition} 
\newtheorem{remark}{Remark}
\newtheorem{corollary}{Corollary}
\newtheorem{definition}{Definition}

\begin{document}

\title{Variable~Selection in Maximum~Mean~Discrepancy for Interpretable~Distribution~Comparison}

\author{Kensuke Mitsuzawa\orcidlink{0009-0004-4793-1168},
        Motonobu Kanagawa\orcidlink{0000-0002-3948-8053},
        Stefano Bortoli,
        Margherita Grossi,
        Paolo Papotti\orcidlink{0000-0003-0651-4128}
\IEEEcompsocitemizethanks{
\IEEEcompsocthanksitem Kensuke Mitsuzawa is with Laboratoire Jean-Alexandre Dieudonné, Université Côte d'Azur, Parc Valrose, 28 Avenue Valrose 06108, Nice, France. (Work done while affiliated with EURECOM.) \protect\\ E-mail: kensuke.mitsuzawa@unice.fr \\

\IEEEcompsocthanksitem Motonobu Kanagawa and Paolo Papotti are with the Data Science Department, EURECOM, 450 Route des Chappes, 06410 Biot, France. \protect\\
E-mail: \{motonobu.kanagawa,paolo.papotti\}@eurecom.fr \\

\IEEEcompsocthanksitem Stefano Bortoli is with Microblink, Trg Drage Iblera 10, 10000, Zagreb,  Croatia.
(Work done while affiliated with Huawei Munich Research Center.)
\protect\\
Email: stefano.bortoli@microblink.com \\

\IEEEcompsocthanksitem Margherita Grossi is with the Intelligent Cloud Technologies Laboratory, Huawei Munich Research Center, Riesstraße 25, 80992 München, Germany. \protect\\
Email: margherita.grossi@huawei.com 
}
}


\IEEEtitleabstractindextext{
\begin{abstract}

We study two-sample variable selection: identifying variables that discriminate between the distributions of two sets of data vectors. Such variables help scientists understand the mechanisms behind dataset discrepancies. Although domain-specific methods exist (e.g., in medical imaging, genetics, and computational social science), a general framework remains underdeveloped.
We make two separate contributions.
(i) We introduce a mathematical notion of the \emph{discriminating set of variables}: the largest subset containing no variables whose marginals are identical across the two distributions and independent of the remaining variables. We prove this set is uniquely defined and establish further properties, making it a suitable ground truth for theory and evaluation.
(ii) We propose two methods for two-sample variable selection that assign weights to variables and optimise them to maximise the power of a kernel two-sample test while enforcing sparsity to downweight redundant variables. To select the regularisation parameter—unknown in practice, as it controls the number of selected variables—we develop two data-driven procedures to balance recall and precision. Synthetic experiments show improved performance over baselines, and we illustrate the approach on two applications using datasets from water-pipe and traffic networks.

\end{abstract}
}

\maketitle

\IEEEraisesectionheading{\section{Introduction} \label{sec:introduction}}

\IEEEPARstart{I}dentifying the discrepancies between two given datasets is a fundamental task in many fields involving statistics, machine learning, and pattern analysis~\cite[e.g.,][]{NIPS2016_5680522b, Sutherland2016,lopez-paz2017revisiting,muandet2021counterfactual}. 

This paper focuses on the identification of discrepancies between two datasets appearing in their {\em variables} (or {\em features}): we will call this problem {\em two-sample variable selection}. Specifically, suppose that two sets of data vectors, all having the same number of variables, are respectively generated from two unknown probability distributions. The problem is to select the variables that discriminate between the two distributions using the given datasets. 
 The discriminating variables can serve as interpretable indicators of discrepancies between two datasets and offer insights into the mechanisms underlying them.

\begin{figure}[t]
  \begin{tabular}{cc}
  \begin{minipage}[t]{0.49\hsize}
      \centering
      \includegraphics[scale=0.45]{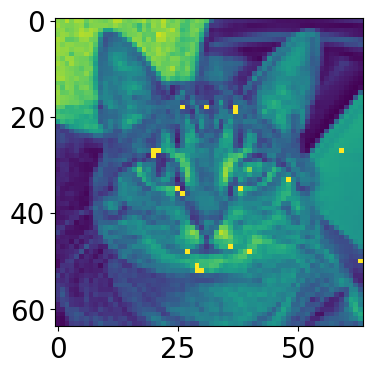}
  \end{minipage} &

  \begin{minipage}[t]{0.49\hsize}
      \centering
      \includegraphics[scale=0.45]{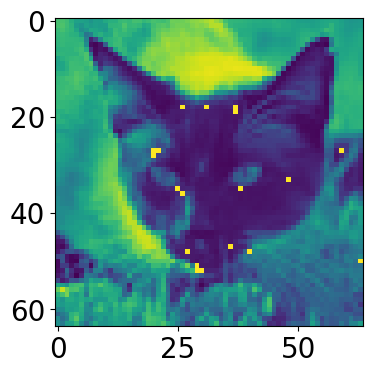}
  \end{minipage} \\
 \hline \hline
  \begin{minipage}[t]{0.49\hsize}
      \centering
      \includegraphics[scale=0.45]{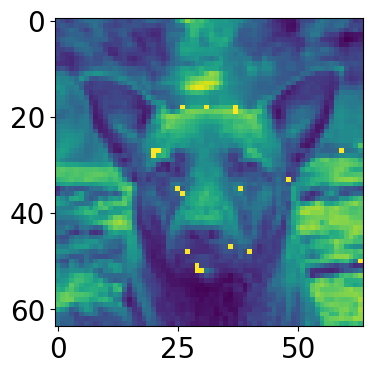}
  \end{minipage} &
  \begin{minipage}[t]{0.49\hsize}
      \centering
      \includegraphics[scale=0.45]{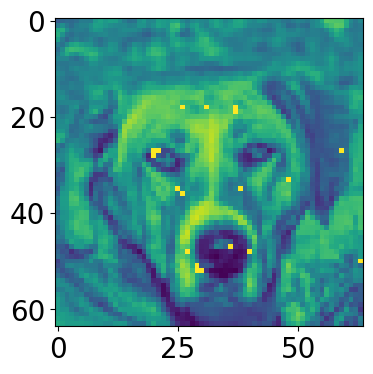}
  \end{minipage}
  \end{tabular}
  \caption{
  Illustration of discriminating variables = pixels (yellow dots) selected by one of the proposed methods (Algorithm~\ref{alg-enhanced-stability-selection}), applied to two sets of face images of cats and dogs. Two example images from each set is shown. 
  See Appendix~\ref{sec:cats-dogs-demo} for details.
  }
  \label{fig:dog-cat-afhq-cv-agg-masked}
\end{figure}

For example, given two datasets of images, the variables may be defined as different pixels or patches, and the problem is to identify the pixels that discriminate between the distributions of the two datasets.  For illustration, Figure~\ref{fig:dog-cat-afhq-cv-agg-masked} shows a toy example where two datasets consist of face images of cats and dogs (only two from each set are shown). The discriminating variables (pixels) selected by one of the proposed methods (Algorithm~\ref{alg-enhanced-stability-selection}) are indicated by yellow dots and positioned at key face parts that distinguish cats from dogs.

The two-sample variable selection problem naturally appears in various fields.  
Suppose two datasets consist of medical images—one from healthy subjects and the other from patients with a certain disease—and that voxels are treated as variables. Then the discriminating variables indicate possible effects of the disease~\cite[e.g.,][]{ashburner2000voxel,nichols2002nonparametric,smith2006tract}. 
In genomics, each data vector may represent a subject’s measured expression levels of different genes (variables), and two datasets may come from subjects with and without a certain disease.  Differential expression analysis~\cite[e.g.,][]{robinson2010edger,love2014moderated} aims to estimate the effects of disease by identifying genes whose expression levels differ significantly between the two datasets. 
In computational social science and natural language processing, two-sample variable selection arises in identifying words (variables) that discriminate between two sets of text data from different groups, such as political parties, genders, or age cohorts~\cite[e.g.,][]{monroe2008fightin,schwartz2013personality}.

Although two-sample variable selection has been addressed in domain-specific ways in various fields, its development as a general methodological framework is comparatively new. Hido et al.~\cite{hido2008unsupervised} and Lopez-Paz \& Oquab~\cite{lopez-paz2017revisiting} study classifier-based two-sample tests, where a classifier is trained to distinguish between the two datasets and its predictive accuracy is used as the test statistic; the learned feature weights are then interpreted qualitatively to suggest variables driving the distributional differences.
Yamada et al.~\cite{yamada2018post} and Lim et al.~\cite{pmlr-v108-lim20a} develop post-selection inference methods based on the maximum mean discrepancy (MMD)~\cite{JMLR:v13:gretton12a}, a kernel-based distance between distributions. Their approaches select variables whose univariate marginal distributions differ, but, by construction, they miss discrepancies arising solely from multivariate structure. 
Hara et al.~\cite{hara2017consistent} and Zhen et al.~\cite{zheng2020nonparametric} compute pairwise divergences between bivariate marginals (e.g., Kullback–Leibler or Wasserstein), form a variable-by-variable matrix, and select variables associated with large entries. Mueller \& Jaakkola~\cite{mueller2015principal} estimate a linear projection that maximises the one-dimensional Wasserstein distance between the projected datasets; the projection coefficients then serve as variable importances for selection. 
Sutherland et al.~\cite{Sutherland2016}, and similarly Wang et al.~\cite{wang2023variable}, learn automatic relevance determination (ARD)–style variable weights by maximising the statistical power of the MMD two-sample test.\footnote{A separate line of work performs {\em sample selection}: it identifies locations in the input space where the two distributions differ most. Some methods estimate the two densities directly~\cite{duong2009highest,duong2013local}, others compare kernel mean embeddings to locate discrepant points~\cite{NIPS2015_0fcbc61a,Jitkrittum2016,NIPS2016_5680522b}, and others frame it as regression to find points most easily discriminated as coming from one distribution or the other~\cite{cazals:hal-01245408,kim2019global}. These approaches target points, not variables, and therefore address a different problem from the one studied here.} 

\subsection*{Contributions}

We make two separate contributions to the emerging field of two-sample variable selection, one mathematical and one methodological. \\[2pt]

\noindent
{\bf Mathematical.}
A fundamental problem in the literature is that the notion of ``variables discriminating between two distributions" has not been rigorously defined, thus making the target set for selection obscure. The only exception is Hara et al.~\cite[Problem 1]{hara2017consistent}, who define a target set of variables whose removal equalises the marginal distributions of the remaining variables, but adding any variable in the target set back to the remaining variables makes their distributions different.  Such a target set, however, fails to include variables whose discrepancies across the two distributions arise solely from their correlation structures (see Remark~\ref{remark:hara-def} for details). In this sense, their definition does not capture all possible discrepancies between two distributions.

Our first contribution is to introduce a formal mathematical definition of the {\em discriminating set of variables} (Definition~\ref{def:subset}).
We define the discriminating set as the largest subset containing no ``redundant'' variables whose marginals are identical across the two distributions and independent of the remaining variables.  We prove that the discriminating set is unique, so all the variables are uniquely divided into the discriminating and redundant variables (Proposition~\ref{prop:problem-formulation}). Moreover, variables whose discrepancies arise solely from their correlation structure, if they exist, are included in the discriminating set, thus resolving the issue in the definition of Hara et al.~\cite{hara2017consistent}. 
Therefore, the discriminating set can serve as a suitable ground truth for theory and evaluation, laying the foundation for the new field of two-sample variable selection. \\[2pt]

\noindent
{\bf Methodological.}
We develop two practical algorithms for two-sample variable selection based on MMD~\cite{JMLR:v13:gretton12a}, a kernel-based distance that quantifies the dissimilarity between two distributions. We adopt an ARD kernel, in which a weight parameter is assigned to each variable to model its relevance~\cite{Sutherland2016}.  These weights are optimised to maximise the estimated MMD test power to make the MMD most sensitive to distributional differences,  while penalising redundant weights via sparsity-inducing $\ell_1$ regularisation~\cite{tibshirani96regression}.  The optimised weights are then used for selecting discriminating variables.

The question is how to select the regularisation parameter, which specifies the strength of regularisation and indirectly determines the number of discriminating variables, and is unknown in practice. Too strong regularisation may lead to low coverage of true discriminating variables, while too weak (or no) regularisation may lead to many false discoveries.  This is a common problem in existing methods for two-sample variable selection, where either the regularisation strength or the number of discriminating variables is treated as a hyperparameter
\cite[e.g.,][]{yamada2018post,pmlr-v108-lim20a,mueller2015principal,wang2023variable}.

We propose two data-driven methods for addressing regularisation parameter selection, aimed at balancing recall (high coverage) and precision (low false discovery).  \begin{enumerate}
    \item The first method (Algorithm~\ref{alg-hyperparameter_selection}) selects a regularisation level that maximises the estimated MMD power (a proxy for high recall) while yielding a low permutation test p-value for the selected variables (a proxy for high precision).

 \item The second method (Algorithm~\ref{alg-enhanced-stability-selection}) extends the first method by aggregating the results across candidate levels to compute variable-importance scores. This method is more sample-efficient than the first one, thus performing better with small samples, albeit at a higher computational cost.
\end{enumerate}
Since these strategies require only an estimate of test power and a permutation test on the selected variables, the same principles could also guide hyperparameter selection in other two-sample selection methods, not just those based on MMD. Synthetic experiments show that both strategies work well, with the aggregation approach performing best, especially in small-sample settings.

~ \\
This paper is organised as follows. Section~\ref{sec:background} offers necessary background on MMD, ARD weights, and test power maximisation. 
Section~\ref{sec:problem-formulation} defines the discriminating set of variables and proves relevant properties, such as its uniqueness.  
MMD is also analysed with respect to discriminating variables, and it is argued that regularisation is necessary to eliminate certain obviously redundant variables. 
Section~\ref{sec:proposed-methods} introduces regularised test power optimisation and the two proposed methods for addressing the regularisation parameter selection problem. 
Section~\ref{sec:empirical_assessment} describes experiments on synthetic controlled datasets. Our approach is demonstrated in experiments on a water-leakage pipe dataset in Section~\ref{sec:simulation_water_pipe_network} and on a traffic dataset with blocked lanes in Section~\ref{sec:simulated_traffic_dataset_body}. 
Supplementary materials offer proofs for mathematical results and additional experiments.

\section{Preliminaries}
\label{sec:background}

We provide the necessary background on MMD in Section \ref{sec:mmd-def} and optimisation of ARD weights to maximise the MMD test power in Section \ref{sec:mmd-opt}.

\subsection{Maximum Mean Discrepancy (MMD)}
\label{sec:mmd-def}

Maximum Mean Discrepancy (MMD) is a (kernel–dependent) distance between
probability distributions that quantifies how different two distributions are
\cite{JMLR:v13:gretton12a}. 
Let $k$ be a positive definite kernel on the $D$-dimensional Euclidean space $\mathbb{R}^D$, which measures
similarity between two input vectors (e.g., linear, polynomial, Gaussian, or Laplace
kernels; see \cite{SchSmo02}). Given such a kernel, the squared MMD between two probability
distributions $P$ and $Q$ on $\mathbb{R}^D$ is defined as
\begin{equation}
  \begin{split}
    {\rm MMD}_k^2(P, Q) :=\ & \mathbb{E}_{X, X' \sim P}[k(X, X')] + \mathbb{E}_{Y, Y' \sim Q}[k(Y, Y')] \\
    &  - 2\mathbb{E}_{X \sim P,\ Y \sim Q}[k(X, Y)], \label{eq:mmd-pq}
  \end{split}
\end{equation}
where $X$ and $X'$ are independent random vectors following $P$, and $Y$ and $Y'$ are those following $Q$.
MMD compares average similarity \emph{within} each distribution to average
similarity \emph{across} the two.  

For any kernel, $\mathrm{MMD}_k^2(P,Q)\ge 0$ and equals $0$ when $P=Q$.
If the kernel is \emph{characteristic} \cite{Fukumizu2008}, then
$\mathrm{MMD}_k^2(P,Q)=0$ holds \emph{if and only if} $P=Q$; equivalently,
$\mathrm{MMD}_k^2(P,Q)>0$ whenever $P\ne Q$. Gaussian and Laplace kernels are
characteristic \cite{SriGreFukSchetal10}.  

Given independently and identically distributed (i.i.d.) samples ${\bf X} :=\{X^1,\dots,X^n\} \stackrel{i.i.d.}{\sim} P$ and
${\bf Y} :=\{Y^1,\dots,Y^m\} \stackrel{i.i.d.}{\sim} Q$ of sizes $n$ and $m$, respectively, replace the expectations in
\eqref{eq:mmd-pq} by empirical averages to obtain the unbiased $U$-statistic
estimator
\begin{equation} 
  \label{eq:mmd-unbiased-est}
  \begin{split}
    & \widehat{\rm MMD}^{2}_U( {\bf X}, {\bf Y} ) := \frac{1}{n (n-1)} \sum_{1 \leq i \ne i' \leq n} k(X^i, X^{i'}) \\
    & + \frac{1}{m(m-1)} \sum_{1 \leq j \ne j' \leq m} k(Y^j, Y^{j'})  - \frac{2}{nm} \sum_{i=1}^n \sum_{j=1}^m k(X^i, Y^j).
  \end{split}
\end{equation}
This estimator converges to the true MMD at the usual parametric rate~\cite[Theorem 10]{JMLR:v13:gretton12a}.
When the kernel is characteristic, the estimated MMD serves as a valid
two-sample test statistic: larger values indicate greater distributional
difference, smaller values greater similarity.

\subsection{Automatic Relevance Detection (ARD)}
\label{sec:mmd-opt}

We summarise the variable–selection approach of Sutherland et al.~\citep[Section 4]{Sutherland2016} based on an ARD (dimension-wise) kernel 
\begin{align} 
  \label{eq:ard-kernel-matrix}
  k(x, y) &=  \exp \left(- \frac{1}{D} \sum^D_{d=1} \frac{ a_{d}^2 (x_{d} - y_{d})^2}{\gamma_{d}^2} \right),  \\
& x := (x_1, \dots, x_D)^\top \in \mathbb{R}^D,\ \   y := (y_1, \dots, y_D)^\top \in \mathbb{R}^D, \nonumber
\end{align}
where $a_1,\dots,a_D \ge 0$ are ARD weights and $\gamma_1,\dots,\gamma_D>0$ are
length scales. Larger $a_d$ increases the influence of differences in the $d$-th
variable on $k(x,y)$; $\gamma_d$ rescales the $d$-th coordinate (chosen via the
per-variable median heuristic; see Appendix~\ref{sec:length-scale-selection}).

Sutherland et al.~\citep{Sutherland2016} choose the ARD weights $a_1, \dots, a_D$ to maximise an asymptotic approximation to test power \citep{gretton2012optimal}, the objective function being
\begin{equation} \label{eq:430}
    \ell \big( a_1, \dots, a_D \big) := \frac{ \widehat{\rm MMD}^{2}_U( {\bf X}, {\bf Y} ) }{\sqrt{\widehat{V}_{n,m} ( {\bf X}, {\bf Y}  ) + C}} ~,
\end{equation}
where the numerator is the unbiased MMD estimate in \eqref{eq:mmd-unbiased-est} and $C \geq 0$ is a small constant.
$\widehat{V}_{n,m}({\bf X},{\bf Y})$ in the denominator is an unbiased estimate of the variance
of $\widehat{\rm MMD}^{2}_U ({\bf X},{\bf Y})$ with respect to the sampling of ${\bf X},{\bf Y}$.
For $n=m$ (the form we use \citep{pmlr-v119-liu20m}),
\begin{equation} \label{eq:var-unbiased-est}
    \widehat{V}_{n,m} ({\bf X}, {\bf Y}) := \frac{4}{n^3} \sum_{i=1}^n \left( \sum_{j=1}^n H_{ij} \right)^2 - \frac{4}{n^4} \left(  \sum_{i, j = 1}^n H_{ij} \right)^2, 
\end{equation}
with $H_{i,j} := k(X^i, X^j) + k(Y^i, Y^j) - k(X^i, Y^j) - k(Y^i, X^j)$.

Both the numerator and denominator in \eqref{eq:430} depend on the ARD kernel.
Increasing $\widehat{\rm MMD}^{2}_U ({\bf X},{\bf Y})$ favors weights that enlarge the separation between the two samples; decreasing $\widehat{V}_{n,m}({\bf X},{\bf Y})$ favors weights that stabilize the estimator across hypothetically resampled datasets. The objective balances these two effects.

The constant $C$ prevents instability when $\widehat{V}_{n,m}$ is very small; we
follow Liu et al.~\cite{pmlr-v119-liu20m} and set $C=10^{-8}$ in experiments.

\section{Discriminating Variables}
\label{sec:problem-formulation}

We introduce a formal mathematical definition of discriminating variables between two probability distributions. Section~\ref{sec:notation-disc} summarises the notation, Section~\ref{sec:def-discriminating-var} defines a discriminating set of variables and analyses its properties, Section~\ref{sec:examples-disc} offers examples, and Section~\ref{sec:theory} analyses MMD with respect to discriminating variables.

\subsection{Notation} \label{sec:notation-disc}
We fix the notation.
Let $\{1, \dots, D\}$ be the set of indices for the $D$ variables ($D \geq 1$).
Let $S \subseteq \{1,\dots,D\}$ be a subset. The complement of $S$, written $S^{c}$, is the set of all indices in $\{1,\dots,D\}$ that are not in $S$. More generally, for two sets $S$ and $U$, the notation $S \backslash U$ denotes the set of all elements in $S$ that are not in $U$. 
Denote by $|S|$ the number of elements in $S$.

For a $D$-dimensional vector $a$, let $a_S$ denote the $|S|$-dimensional subvector consisting of the components of $a$ in $S$.

For a probability distribution $P$ on the $D$-dimensional Euclidean space $\mathbb{R}^{D}$, we write $P_{S}$ for its marginal distribution on the variables in $S$, and $P_{S^{c}}$ for the marginal on the variables in the complement. The same notation $Q_{S}$ and $Q_{S^{c}}$ is used for another distribution $Q$.

For any disjoint subsets $S,U \subseteq \{1,\dots,D\}$, write $P_S \otimes P_U$ for the product distribution of the marginals $P_S$ and $P_U$.

Recall that if the marginal of $P$ on $S \cup U$ equals this product, i.e., $P_{S \cup U} = P_S \otimes P_U$, then for a random vector $(X_S, X_U) \sim P_{S \cup U}$ the subvectors $X_S$ and $X_U$ are (statistically) independent.

\subsection{Defining Discriminating Variables}
\label{sec:def-discriminating-var}

We define the set of discriminating variables between two probability distributions as that satisfying the following two conditions: 
\begin{enumerate}
    \item \textbf{No redundant subset:} None of its subsets have identical distributions under both probability distributions and are independent of the remaining variables within each distribution. Such subsets are redundant and excluded.  
    \item \textbf{Maximality:} The set is as large as possible while still satisfying the first condition; adding any further variable would make the condition fail.  
\end{enumerate}
The formal definition is as follows.

\begin{definition}[Discriminating Set of Variables] \label{def:subset}
Let $P$ and $Q$ be probability distributions on $\mathbb{R}^D$.
A subset  $S \subset \{1, \dots, D\}$ of variable indices is called a {\bf discriminating set}, if it satisfies the following: 
\begin{enumerate}
    \item \textbf{No redundant subset:} There is no non-empty subset  $U \subset S$  such that  
    \begin{align*}
    &  P_U = Q_U, \quad P_S = P_U \otimes P_{S \backslash U}, \quad Q_S = Q_U \otimes Q_{S \backslash U}.
    \end{align*}
    
    \item \textbf{Maximality:} There is no strict superset of $S$ that satisfies  condition 1).
\end{enumerate}

\end{definition}

The following proposition shows that a discriminating set of variables exists {\em uniquely}: remove the largest set of variables whose marginal distributions are identical under both distributions and that are independent of the remaining variables within each distribution; the variables that remain after this removal form the unique discriminating set.

\begin{proposition} \label{prop:problem-formulation}
For any probability distributions $P$ and $Q$ on $\mathbb{R}^D$, a discriminating subset $S \subset \{1, \dots, D\}$ satisfying Definition~\ref{def:subset} exists uniquely. 
Moreover, let $U \subset \{1, \dots, D\}$ be the largest subset on which $P$ and $Q$ have identical marginals and which are independent of the remaining variables:
\[
P_U = Q_U , \quad P = P_U \otimes P_{\{1, \dots, D\}\setminus U}, \quad 
Q = Q_U \otimes Q_{\{1, \dots, D\}\setminus U}.
\]
Then the discriminating set is given by its complement, $$S = \{1, \dots, D\} \setminus U,$$ and we obtain the decomposition
\begin{equation} \label{eq:decomposition}
    P = P_S \otimes P_{\{1, \dots, D\}\setminus S}, \qquad
Q = Q_S \otimes Q_{\{1, \dots, D\}\setminus S}.
\end{equation}
 
\end{proposition}

\begin{proof}
    Appendix \ref{sec:prop-problem-form}.
\end{proof}

This proposition establishes that any two distributions can be decomposed into two parts: the variables on which their marginal distributions coincide and are independent of the rest, and the complementary variables; the latter form the unique discriminating set, on which the marginal distributions differ. In this way, all discrepancies are precisely confined to a uniquely specified set of discriminating variables.

Lastly, we show that the discriminating set is the empty set if and only if the two distributions coincide. 
\begin{corollary} \label{coro:p-equal-q}
    For probability distributions $P$ and $Q$ on $\mathbb{R}^D$, the discriminating set is the empty set if and only if $P = Q$.
\end{corollary}
\begin{proof}
    Appendix~\ref{sec:proof-p-equal-q}.
\end{proof}

\subsection{Examples} \label{sec:examples-disc}

We provide a few examples; other examples can be found in Section~\ref{sec:data_generation_process}.

\begin{example} \label{ex:discriminating-set2}

Let \(D=2\) with variables \(\{1,2\}\). Assume
\[
P_{\{1\}} \neq Q_{\{1\}}, \qquad
P_{\{2\}} = Q_{\{2\}},
\]
and that under both \(P\) and \(Q\) the two variables are independent,
$P = P_{\{1\}} \otimes P_{\{2\}}$ and $Q = Q_{\{1\}} \otimes Q_{\{2\}}$.
In this case \(S=\{1\}\) is the \emph{discriminating set}:  
(i) variable \(1\) is necessary because its marginal distributions differ;  
(ii) variable \(2\) is redundant since its distribution is the same under both \(P\) and \(Q\) and is independent of variable \(1\).

\end{example}

\begin{example} \label{ex:discriminating-set1}

Let \(D=2\) with variables \(\{1,2\}\). 
Assume identical univariate marginals $P_{\{1\}} = Q_{\{1\}}$ and $P_{\{2\}} = Q_{\{2\}}$, and that the two variables are independent under \(P\) but dependent under \(Q\):
\[
P = P_{\{1\}} \otimes P_{\{2\}},
\quad
Q \neq Q_{\{1\}} \otimes Q_{\{2\}}.
\]
 In this case \(S=\{1,2\}\) is the \emph{discriminating set}:
(i) no singleton subset is redundant, because although \(P_{\{j\}}=Q_{\{j\}}\) and
\(P_S = P_{\{j\}}\otimes P_{S\setminus\{j\}}\), we have
\(Q_S \neq Q_{\{j\}} \otimes Q_{S\setminus\{j\}}\) for \(j\in\{1,2\}\);
(ii) \(S\) is maximal in \(D=2\).
Equivalently, \emph{both variables are necessary} to capture the dependence difference between \(P\) and \(Q\).

\end{example}

\begin{remark} \label{remark:hara-def}
 Hara et al.~\cite[Problem 1]{hara2017consistent} define a target set of variables \(S\) such that the
marginal distributions of \(P\) and \(Q\) on the variables excluding \(S\) are
identical, and adding back any single variable from \(S\) makes the two
distributions differ again.
Formally, they require \(S \subset \{1,\dots,D\}\) such that
\begin{align*}
& P_{\{1,\dots,D\}\setminus S} = Q_{\{1,\dots,D\}\setminus S}
\quad\text{and} \\
& P_{\{d\}\cup(\{1,\dots,D\}\setminus S)} \neq Q_{\{d\}\cup(\{1,\dots,D\}\setminus S)},
\quad \forall d \in S .    
\end{align*}
This definition fails for Example~\ref{ex:discriminating-set1}.
Removing variable 1 leaves matching marginals on variable 2
        \((P_{\{2\}}=Q_{\{2\}})\), and adding variable 1 back yields \(P\neq Q\),
        so \(S=\{1\}\) is accepted (symmetrically, \(S=\{2\}\) is also accepted).
 However, \(S=\{1,2\}\) is not allowed: removing both variables leaves no variables (trivially matching), and adding back only one at a time keeps the marginals equal \((P_{\{1\}}=Q_{\{1\}},\, P_{\{2\}}=Q_{\{2\}})\), so the condition fails.
This is problematic because the discrepancy between \(P\) and \(Q\) lies in their
\emph{joint} dependence structure and appears only when both variables are
considered together; the appropriate discriminating set is therefore \(S=\{1,2\}\),
which Hara’s definition cannot capture.

\end{remark}

\subsection{MMD and Discriminating Variables}
\label{sec:theory}

Lastly, we analyse theoretical properties of MMD with respect to discriminating variables when using a generalised ARD kernel defined below.  
Based on this, we argue that regularisation is necessary when optimising ARD weights to eliminate clearly redundant variables.

Let $\phi_d:\mathbb{R}\to\mathbb{R}$ be one–dimensional positive definite functions
(for $d=1,\dots,D$) and let $a_1,\dots,a_D\ge 0$ be ARD weights. Define a generalised ARD kernel as
\begin{align} \label{eq:gen-ARD-kernel}
    & k(x,x') := \prod_{d=1}^D \phi_d \left( \sqrt{ a_d^2 (x_d - x_d')^2  / D} \right), \\
    & x=(x_1,\dots,x_D)^\top,\quad x'=(x'_1,\dots,x'_D)^\top\in\mathbb{R}^D. \nonumber
\end{align}
Choosing $\phi_d(r)=\exp(-r^2/\gamma_d^2)$ recovers the Gaussian ARD kernel in
\eqref{eq:ard-kernel-matrix}; choosing $\phi_d(r)=\exp(-|r|/\gamma_d)$ yields a
Laplace-type ARD kernel.

For any subset $S\subset\{1,\dots,D\}$, denote the restriction the generalised ARD kernel
\eqref{eq:gen-ARD-kernel} to the variables in $S$ by
\[
k_S(x_S,x_S') := \prod_{d\in S}\phi_d\!\left(\sqrt{a_d^2(x_d-x_d')^2/D}\right),
\quad x_S,x_S'\in\mathbb{R}^{|S|}.
\]
Let $U:=\{1,\dots,D\}\setminus S$ be its complement and define the restricted kernel $k_U$ on it analogously.

Proposition~\ref{prop:max-mmd-weights} below states that, with the generalised ARD kernel,
MMD is factorised into the MMD on the discriminating variables $S$ and a constant from the redundant ones $U$; moreover, maximising over all ARD weights reduces to maximising over the weights on the discriminating variables.

\begin{proposition} \label{prop:max-mmd-weights}
For probability distributions $P$ and $Q$ on $\mathbb{R}^D$, let $S \subset \{1, \dots, D\}$ be their discriminating variables (Definition~\ref{def:subset}) and be $U := \{1, \dots, D\} \backslash S$ be the redundant ones. 
Then, the MMD \eqref{eq:mmd-pq} between $P$ and $Q$ using the generalised ARD kernel $k$ in \eqref{eq:gen-ARD-kernel} is written as
\begin{equation} \label{eq:MMD-decomp-629}
{\rm MMD}_k^2(P,Q) = {\rm MMD}_{k_S}^2(P_S,Q_S) ~ \mathbb{E}[ k_U(X_U, X'_U)] ,
\end{equation}
where $X_U, X'_U \stackrel{i.i.d.}{\sim} P_U$.  Moreover, its maximum over the ARD weights is given by
\begin{align} \label{eq:MMD-decomp-max-630}
    \max_{a \in \mathbb{R}^D} {\rm MMD}_k^2(P,Q) =  \max_{a_S \in \mathbb{R}^{|S|}} {\rm MMD}_{k_S}^2(P,Q) ~  \prod_{d \in U} \phi_d \left( 0 \right)  . 
\end{align}

\end{proposition}
\begin{proof}
     Appendix \ref{sec:proof-mmd-max-weights}. 
\end{proof}

From this result, it follows that the MMD-maximising ARD weights for any redundant variables with nonzero variances are zero, as summarised below.

\begin{proposition}\label{eq:prop-MMD-weights_maximizer}
For probability distributions $P$ and $Q$ on $\mathbb{R}^D$, let $S \subset \{1, \dots, D\}$ be their discriminating variables (Definition~\ref{def:subset}) and $U := \{1, \dots, D\} \backslash S$ be the redundant ones. 
Let $k$ be the generalised ARD kernel \eqref{eq:gen-ARD-kernel} such that each $\phi_d$ is strictly maximised at $0$, i.e., $\phi_d(0)>\phi_d(r)$ for all $r>0$.
Let $a^* = (a_1^*, \dots, a_d^*)^\top$ be ARD weights that maximise the MMD:
$$
a^*\in\arg\max_{a\in\mathbb{R}^D}{\rm MMD}_k^2(P,Q).
$$
Then 
\[
a^*_d = 0 ~~~~  \text{if} ~~ d\in U ~~ \text{and} ~~ \mathrm{Var}(P_d)=\mathrm{Var}(Q_d)>0,  
\]
 where $\mathrm{Var}(P_d)$ and $\mathrm{Var}(Q_d)$ denote the variances of the marginal distributions $P_d$ and $Q_d$ of the $d$-th variable, respectively.
\end{proposition}

\begin{proof}
  See  Appendix \ref{sec:proof-mmd-max-weights-maximizer}. 
\end{proof}

This result shows that the MMD-maximising ARD weights of redundant variables are zero {\em if} their variances are positive. 
On the contrary, if their variances are zero, i.e., their marginal distributions are degenerate, then the MMD-maximising ARD weights may not necessarily be zero, even if they are redundant. 
This is because for redundant variables with degenerate marginals, the ARD weights do not influence the kernel value, as summarised below.

\begin{example} \label{ex:redundant-variables}
If a redundant variable $d \in U$ has zero variance, it always takes the same value, say $\xi \in \mathbb{R}$, for the two distributions $P$ and $Q$. 
The $d$-th component of the ARD kernel~\eqref{eq:gen-ARD-kernel} evaluated on samples of $P$ and/or $Q$ then becomes a constant:
\begin{align*}
& \phi_d( \sqrt{a_d^2 (\xi_d -\xi_d)^2} )  = \phi_d(0).
\end{align*} 
Thus, the ARD weight $a_d$ does not affect the kernel output. This implies that, in this case, the ARD weight can take an arbitrary value to achieve the maximum MMD, even if this variable is clearly redundant and should be removed.

\end{example}

\begin{figure}[t]
  \begin{minipage}[b]{0.45\linewidth}
    \centering
    \includegraphics[width=40mm]{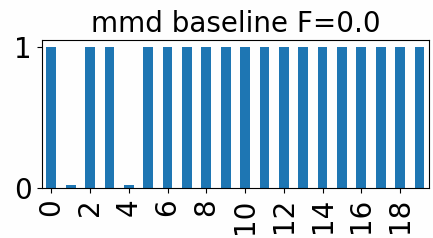}
  \end{minipage}%
\hfill
  \begin{minipage}[b]{0.45\linewidth}
    \centering
    \includegraphics[width=40mm]{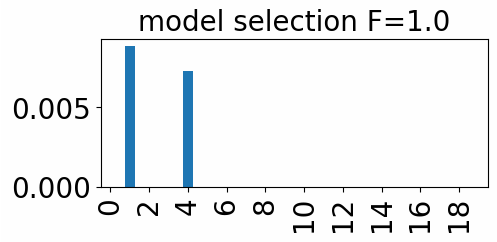}
  \end{minipage}
  \caption{
    Optimised ARD weights without regularisation (Left, \cite{Sutherland2016}) and with regularisation (Right, Algorithm~\ref{alg-hyperparameter_selection}).
    Here, $S = \{1, 4\}$ are the true discriminating variables for distinguishing $P$ and $Q$, and $U = \{0, 2, 3, 5, \dots, 19\}$ are the redundant variables with zero-variance marginal distributions. 
    Without regularisation, the redundant variables' ARD weights do not change from their initial value of $1$, burying the weights of the discriminating variables. With regularisation, these redundant variables are successfully eliminated.  The details of the setting are described in Section~\ref{sec:synthetic_data_assessment} (``Redundant Dirac'').   
  }
  \label{fig:example-regularization-effect}
\end{figure}

Example~\ref{ex:redundant-variables} illustrates that certain clearly redundant variables cannot be identified by maximising the MMD for the ARD weights.
The same issue occurs with maximising the test power objective~\eqref{eq:430}, because the objective value is independent of the ARD weights for the redundant variables with zero variance, as it depends on the ARD weights only through the kernel. 
This is illustrated in Figure~\ref{fig:example-regularization-effect}, where the ARD weights on the redundant variables with zero variance remain their initial values and do not become zero after optimisation. 
To resolve the issue, the optimisation objective should include a regularisation term that shrinks redundant variables. This motivates our regularised objective introduced next.

\section{Proposed Methods}
\label{sec:proposed-methods}

This section describes the two proposed methods for two-sample variable selection. 
Section~\ref{sec:regularization} introduces a regularised MMD power optimisation problem for ARD weights.  
Section~\ref{sec:regularized_model_selection} describes the core idea of our regularisation parameter selection algorithm. Section~\ref{sec:stability-selection} extends it to an aggregation algorithm that combines the results from different regularisation parameters.

\label{sec:proposed}

\subsection{Regularisation for Variable Selection}
\label{sec:regularization}

Motivated by the discussion in Section~\ref{sec:theory}, we propose to optimise the ARD weights by minimising the negative logarithm of the MMD test power objective~\eqref{eq:430},  which is equivalent to maximising the test power, plus a regularisation term given by the $\ell_1$ norm of the ARD weights: 

\begin{equation}
\label{eq:regularized-opt}
\min_{a_1, \dots, a_D \in \mathbb{R}}\; -\log\!\big(\ell(a_1,\dots,a_D)\big)\;+\;\lambda\sum_{d=1}^D |a_d|,
\end{equation}
where $\lambda\ge 0$  is a regularisation parameter.\footnote{After the initial version of our paper appeared in November 2023 on arXiv ({\tt 2311.01537v1}), an independent work by Adamer et al.~\cite{adamer2024biomarker} appeared in July 2024 and used essentially the same optimisation approach as \eqref{eq:regularized-opt} in biomarker identification; a key difference from our work is that they did not address the issue of regularisation parameter selection.} 
When $\lambda=0$, the problem~\eqref{eq:regularized-opt} reduces to the maximisation of the test power \eqref{eq:430}.

The $\ell_1$ norm regularisation, known as Lasso~\cite{tibshirani96regression}, shrinks redundant weights and also encourages sparsity: many of the optimised weights become zero, thus a standard choice in variable selection~\cite{hastie2015statistical}.    
Figure~\ref{fig:example-regularization-effect} shows how it resolves the issue of unregularised optimisation discussed in Section~\ref{sec:theory}.

The regularisation parameter $\lambda$ in \eqref{eq:regularized-opt} controls sparsity. If it is too large, many ARD weights collapse to zero (false negatives); if too small, redundant variables retain nonzero weights (false positives). See
Appendix~\ref{sec:appendix-syn2-result-parameter-choice} for empirical evidence.
We develop two methods for addressing this issue, detailed in
Sections~\ref{sec:regularized_model_selection} and \ref{sec:stability-selection}.

\label{sec:variable_selection}
\begin{remark}[Variable selection from optimised weights]
 Given the optimised ARD weights $a^*=(a^*_1,\dots,a^*_D)$ from \eqref{eq:regularized-opt},
we select variables whose weights exceed a \emph{data-driven} threshold. 
A fixed threshold performs poorly because weight scales vary widely across datasets. 
Instead, we set the threshold as the smallest local minimum of the histogram of $\{a^*_d\}$. 
(Details are in Appendix~\ref{app:variable_selection}.)
\end{remark}

\label{sec:hyperparameter_range_method}
\begin{remark}[Setting candidate regularization parameters] 
\label{remark:reg-candidates}
We search $\lambda$ over a data-driven range. The lower bound is a small constant (default $0.01$). 
The upper bound is the smallest value at which either only one variable is selected or the selected set stops changing when $\lambda$ increases. 
We then form an equally spaced grid between these bounds. 
(Details are in Appendix~\ref{sec:reg-param-range-selection}.)    
\end{remark}

\subsection{Regularisation Parameter Selection Algorithm}
\label{sec:regularized_model_selection}

\begin{algorithm}[t]
  \caption{Discriminating Variable Selection with Regularisation Parameter Selection. (``{\tt model-selection}'')}
  \label{alg-hyperparameter_selection}
  \textbf{Input:} 
  A set $\Lambda$ of candidate regularisation parameters;  two sets of data vectors   ${\bf X} = \{ X^1, \dots, X^n \} \subset \mathbb{R}^D$ and ${\bf Y} = \{Y^1,..., Y^m\} \subset \mathbb{R}^{D}$;   training-validation split ratio $\rho_{\rm train} \in (0, 1)$ (default value $0.5$).

 \algohline
  \textbf{Output:} 
  A set $\hat{S}_{\lambda^*} \subset \{1, \dots, D
  \}$ of variables selected with the best regularisation parameter  $\lambda^* \in \Lambda$.

  \algohline 
  \begin{algorithmic}[1]
      \State Randomly split the two datasets $({\bf X}, {\bf Y})$ into ``training'' sets $({\bf X}_{\rm train}, {\bf Y}_{\rm train})$ and ``validation'' sets $({\bf X}_{\rm val}, {\bf Y}_{\rm val})$ by ratio $\rho_{\rm train}$.
      \ForAll {$\lambda \in \Lambda$}
        
        \State 
        -- Solve the regularised MMD test power optimisation problem \eqref{eq:regularized-opt} on training sets $({\bf X}_{\rm train}, {\bf Y}_{\rm train})$ with regularisation parameter $\lambda$ to obtain the optimised ARD weights $a_{\lambda} \in \mathbb{R}^D$, from which a set of variables $\hat{S}_{\lambda} \subset \{1, \dots, D\}$  is selected. 

        \State 
        -- Let $\ell_{\rm val}(a_\lambda) > 0$ be the MMD test power objective~\eqref{eq:430} for ARD weights $a_\lambda$ on validation sets $({\bf X}_{\rm val}, {\bf Y}_{\rm val})$.
       

       \State
        -- Let $0 \leq p_\lambda \leq 1$ be the $p$-value from a permutation two-sample test using only the selected variables $\hat{S}_{\lambda}$ on validation sets $({\bf X}_{\rm val}, {\bf Y}_{\rm val})$, with the sliced Wasserstein distance$^*$ as a test statistic. (*Any other test statistic may be used here.)   
    \EndFor

    \If{there exists $\lambda \in \Lambda$ with $p_\lambda < 0.05$} 
    \State  $\lambda^* = \arg\max_{\lambda \in \Lambda } \{ \ell_{\rm val}(a_\lambda) \mid p_{\lambda} < 0.05 \}$. 
    \Else \State 
    $\lambda^* = \arg\min_{\lambda \in \Lambda} p_\lambda$. 
    \EndIf
\end{algorithmic}
\end{algorithm}

 Ideally, variable selection should attain both
 \begin{itemize}
     \item {\em High recall}: many true discriminating variables are selected ({\em high coverage});   
    \item {\em High precision}: the fraction of redundant variables among all the selected variables is small ({\em low false discovery}).  
 \end{itemize}
Recall and precision are in a trade-off relationship: high recall does not imply high precision, and vice versa. For example, if all variables, both discriminating and redundant, are selected, recall is highest, but precision is low. If only one variable is selected and it is a discriminating variable, precision is highest, but recall is low, because many other discriminating variables are not selected.

Our overarching idea is to select the regularisation parameter $\lambda$ that balances recall and precision.  A practical challenge is, of course, that both recall and precision are unknown, as the true discriminating variables are unknown. 

We address this issue by using 1) {\em MMD test power} and 2) {\em the p-value of a permutation two-sample test on the selected variables}, both evaluated on held-out data, as proxies to recall and precision. More precisely, we select a regularisation constant $\lambda$ such that:
\begin{enumerate}
    \item  
The optimised ARD weights make the MMD two-sample test have {\em high power}, i.e., a high probability of distinguishing the two distributions when they differ. This requires {\em the ARD weights to be supported on many discriminating variables, i.e., high recall}. 

\item
 A permutation two-sample test based {\em only} on the selected variables (without the ARD weights) yields a {\em low $p$-value}, i.e., a high discrepancy from the null hypothesis of identical distributions. 
    This requires the selected variables {\em not to contain many redundant variables, i.e. high precision}, because the inclusion of redundant variables, without weights, adds more ``noise'' to ``signals'' carried in the discriminating variables. 
\end{enumerate}
We thus select a regularisation parameter that yields high MMD test power and a small $p$-value for a permutation two-sample test applied to the selected variables, thereby balancing recall and precision. 

Algorithm~\ref{alg-hyperparameter_selection} implements the above idea:
\begin{enumerate}
    \item First, split the data into two sets, one for optimising the ARD weights (``training'') and one for evaluating the test power objective and running a permutation two-sample test (``validation''). 
    \item Using each candidate regularisation parameter $\lambda$, solve the regularised test power maximisation problem~\eqref{eq:regularized-opt} to obtain the optimised ARD weights $a_\lambda$, based on which variables $\hat S_\lambda$ are selected.

    \item Compute the MMD test power objective $\ell_{\rm val}(a_\lambda)$ on the validation data using the optimised ARD weights. 
    
    \item Run a permutation two-sample test on the validation data restricted to the selected variables $\hat S_\lambda$ to compute the $p$-value~$p_\lambda$ (we use the sliced Wasserstein distance~\cite{bonneel2014sliced} as the test statistic for speed, but any multivariate two-sample statistic may be used). 
    \item Finally, choose the regularisation parameter $\lambda$ with the largest MMD test power objective $\ell_{\rm val}(a_\lambda)$ among those with the $p$-values $p_\lambda$ are less than a threshold (here $0.05$ is used); if no candidate meets this threshold, choose the $\lambda$ with the smallest $p_\lambda$.
\end{enumerate}

\subsection{Aggregation Algorithm}
\label{sec:stability-selection}

\begin{algorithm}[t]
  \caption{Discriminating Variable Selection via Aggregation. (``{\tt CV-aggregation}'')}
  \label{alg-enhanced-stability-selection}
    \label{alg-stability-selection}
  \textbf{Input:} 
  A set $\Lambda$ of candidate regularisation parameters;  two sets of data vectors   ${\bf X} = \{ X^1, \dots, X^n \} \subset \mathbb{R}^D$ and ${\bf Y} = \{Y^1,..., Y^m\} \subset \mathbb{R}^{D}$;   training-validation split ratio $\rho_{\rm train} \in (0, 1)$ (default value $0.5$); the number $K$ of cross-validation splits. 
  
  \algohline
  \textbf{Output:} A set $\hat{S} \subset \{1, \dots, D\}$ of selected variables and a variable importance score vector $\hat{\Pi} \in \mathbb{R}^D$.  
  \algohline 
  \begin{algorithmic}[1]
      \ForAll {$\lambda \in \Lambda$} 
        \ForAll {$i \in \{1, ... ,K\}$}
      
        \State
        -- Randomly split the two datasets $({\bf X}, {\bf Y})$ into ``training'' sets $({\bf X}_{\rm train}^{i}, {\bf Y}_{\rm train}^{i})$ and ``validation'' sets $( {\bf X}_{\rm val}^{i}, {\bf Y}_{\rm val}^{i})$ by ratio~$\rho_{\rm train}$.
        
       \State 
        -- Solve the regularised MMD test power optimisation problem \eqref{eq:regularized-opt} on training sets $({\bf X}_{\rm train}^i, {\bf Y}_{\rm train}^i)$ with regularisation parameter $\lambda$ to obtain the optimised ARD weights $a_{\lambda}^i \in \mathbb{R}^D$, from which a set of variables $\hat{S}_{\lambda}^i \subset \{1, \dots, D\}$  is selected.


        \State 
         -- Let $\ell_{\rm val}(a_\lambda^i) > 0$ be the MMD test power objective~\eqref{eq:430} for ARD weights $a_\lambda^i$ on validation sets $({\bf X}_{\rm val}^i, {\bf Y}_{\rm val}^i)$.
        


       \State 
        -- Let $0 \leq p_\lambda^i \leq 1$ be the $p$-value from a permutation two-sample test using only the selected variables $\hat{S}_{\lambda}^i$ on validation sets $({\bf X}_{\rm val}^i, {\bf Y}_{\rm val}^i)$, with the sliced Wasserstein distance$^*$ as a test statistic. (*Any other test statistic may be used here.) 
        \State -- Let $\tilde{a}_\lambda^i := a_\lambda^i / \left( \max_{d \in \{1, \dots, D\} }  a_{\lambda, d}^i \right) \in \mathbb{R}^D$ be the normalised ARD weights. 
        


          

          
        \EndFor

      \State -- Let $\hat{ \Pi }_\lambda := \frac{1}{K} \sum_{i=1}^K \mathbb{I}( p_\lambda^i < 0.05 ) \ell_{\rm val} (a_\lambda^i ) \tilde{a}_\lambda^i \in \mathbb{R}^D$ be the variable-importance score vector for the regularisation parameter $\lambda$. ($\mathbb{I}(\cdot )$ is the indicator function.)
      \EndFor

      \State Let $\hat{\Pi} := \frac{1}{ |\Lambda| } \sum_{ \lambda \in \Lambda } \hat{\Pi}_\lambda \in \mathbb{R}^D$ be the final variable importance score vector, based on which a set of variables $\hat{S} \subset \{1, \dots, D\}$ is selected. 
\end{algorithmic}

\end{algorithm}

Our second algorithm, described in Algorithm~\ref{alg-enhanced-stability-selection}, extends Algorithm~\ref{alg-hyperparameter_selection} by incorporating cross-validation and aggregating results across candidate regularisation parameters, thereby making more efficient use of data and performing more stably with smaller sample sizes, albeit at a higher computational cost.\footnote{Appendix~\ref{sec:appendix-syn2-result-parameter-choice} compares different possible aggregation strategies, in which the one in Algorithm~\ref{alg-enhanced-stability-selection} is shown to perform the best.}  
The idea of aggregation is inspired by the stability selection algorithm~\cite{stability-selection} in high-dimensional statistics, while details differ substantially. 
Algorithm~\ref{alg-enhanced-stability-selection} performs the following:
\begin{enumerate}
    \item For each candidate regularisation parameter $\lambda$, compute a variable-importance score vector $\hat{ \Pi }_\lambda$ as a weighted average of the (normalised) ARD weight vectors obtained for different ``training-validation'' splits of data. 
   The weight in each split is the MMD power objective on the validation data if a permutation two-sample test using the selected variables yields a $p$-value smaller than a threshold, say 0.05,  and zero otherwise  (see Algorithm~\ref{alg-hyperparameter_selection}).

    \item  Compute the final score vector $\hat{\Pi}$, based on which variable selection shall be performed, as the average of the score vectors computed for different regularisation parameters.
\end{enumerate}

The importance of each candidate regularisation parameter is essentially quantified by the MMD power objective (proxy for recall/coverage), thresholded by the $p$-value for the selected variables (proxy for precision); this is what Algorithm~\ref{alg-hyperparameter_selection} does.  Thus, the final score vector of Algorithm~\ref{alg-enhanced-stability-selection} is essentially the weighted average of the normalised ARD weight vectors across candidate regularisation parameters, each weighted by its importance.

\section{Synthetic Data Experiments}
\label{sec:empirical_assessment}

Our methods are assessed against baseline methods on synthetic datasets, where the ground-truth discriminating variables are known for evaluation. 
Only selected results are reported, and more can be found in the supplementary materials. 
We explain the methods to be compared in Section \ref{sec:assessment-opt-configuration} and data settings and evaluation criteria in Section~\ref{sec:data_generation_process}. We discuss the results in Section~\ref{sec:synthetic-exp-results}.

\subsection{Methods}
\label{sec:assessment-opt-configuration}

We compare Algorithm~\ref{alg-hyperparameter_selection} (``\texttt{model-selection}'') and Algorithm~\ref{alg-enhanced-stability-selection} (``\texttt{CV-aggregation}'')  against three baselines:
\begin{itemize}
  \item \texttt{mmd-baseline} (Sutherland \cite[Sec.~5]{Sutherland2016}): optimise the ARD weights by maximising \eqref{eq:430} (equivalently, solve \eqref{eq:regularized-opt} with $\lambda=0$), then select variables from the optimised weights.
  \item \texttt{mskernel-star} (Lim et al.~\cite{pmlr-v108-lim20a}):  post-selection inference using  MMDs on univariate marginals.
  We use the authors’ implementation.\footnote{\url{https://github.com/jenninglim/multiscale-features}. “star” denotes that the number of selected variables is provided.} This method requires the number of variables to select; we set it to the ground-truth value,  a significant (unfair) advantage over other methods. 
  \item \texttt{regression-baseline}: classifier two-sample test using $\ell_1$-regularised logistic regression.\footnote{\url{https://scikit-learn.org/stable/modules/generated/sklearn.linear_model.LogisticRegression.html}.} Variables are selected from the learned coefficients. The regularisation strength is tuned by grid search (inverse parameter on the range $[0.1, 4.0]$) with 5-fold cross-validation.

  \item \texttt{mmd-tuning-best-F1}: the regularised MMD-power optimisation approach~\eqref{eq:regularized-opt} in which the regularisation parameter $\lambda$ is tuned to maximise the F score (defined below). This oracle variant is not implementable in practice—its F score requires the ground-truth variables—but serves as a best-case benchmark for \eqref{eq:regularized-opt}.
\end{itemize}

For kernel-based methods, we use the Gaussian ARD kernel~\eqref{eq:ard-kernel-matrix} with
per-variable length-scales set by the median heuristic (Appendix~\ref{sec:length-scale-selection}).
We optimise \eqref{eq:regularized-opt} for the ARD weights using the Adam optimiser with PyTorch’s
\texttt{ReduceLROnPlateau}\footnote{\url{https://pytorch.org/docs/stable/generated/torch.optim.lr_scheduler.ReduceLROnPlateau.html}}
scheduler (initial learning rate $0.01$, adaptively reduced based on the objective). All ARD
weights are initialised to $1$. Early stopping is triggered if, over 100 consecutive epochs,
the range between the maximum and minimum objective values (evaluated on both training and
validation data) drops below $10^{-3}$; we cap training at 99{,}999 epochs.
For \texttt{model-selection} and \texttt{CV-aggregation}, the set
$\Lambda$ of candidate regularisation parameters are generated as in Remark~\ref{remark:reg-candidates}; for
\texttt{CV-aggregation} we use $K=10$ splits. $p$-values based on the sliced Wasserstein
distance are computed with the \texttt{POT} package~\cite{flamary2021pot}.

\label{sec:discrepancy}
\label{sec:synthetic_data_assessment}

\subsection{Data Settings and Evaluation Criteria} 
\label{sec:data_generation_process}

Two probability distributions $P$ and $Q$ on $\mathbb{R}^D$ are defined such that their marginal distributions $P_S$ and $Q_S$ on the discriminating set $S \subset \{1, \dots, D\}$ differ while those on the remaining variables are identical (see Proposition~\ref{prop:problem-formulation}): 
$$
P = P_S \otimes P_{ \{1, \dots, D\} \backslash S }, \quad Q = Q_S \otimes P_{ \{1, \dots, D\} \backslash S }, \quad P_S \neq Q_S.
$$
We set $P$ as the uncorrelated Gaussian distribution with zero means and unit variances, so are the marginals $P_S$ and $P_{ \{1, \dots, D\} \backslash S }$. 
We consider the following different setups for the marginal $Q_S$:

\begin{enumerate}
    \item {\bf Shifted means}: 
    $Q_S$ is the uncorrelated Gaussian with means~$0.5$ and unit variances, i.e., a mean-shifted version of $P_S$.
    
    \item {\bf Wider variances}: 
    $Q_S$ is the uncorrelated Gaussian with zero means and variances~$1.5$, larger than those of $P_S$.
    
    \item  {\bf Narrower variances}: 
    $Q_S$ is the uncorrelated Gaussian with zero means and variances $0.5$, smaller than those of $P_S$.

    \item {\bf Laplace distribution}: 
    $Q_S$ is the uncorrelated Laplace distribution with zero means and unit variances. It thus has the same first two moments as $P_S$ but differs in its higher-order moments.

    \item {\bf Correlated Gaussian}: 
    A random vector from $Q_S$ has exactly the same components, each of which is marginally Gaussian with zero mean and unit variance. Thus, $Q_S$ and $P_S$ have the same univariate marginal distributions, but their correlation structures differ.


\end{enumerate}
We also consider the following setting corresponding to that discussed in Section \ref{sec:theory}. 
 \begin{enumerate} \setcounter{enumi}{5}
     \item {\bf Redundant Dirac}:
     $P_S$ and $Q_S$ are the same as the ``Shifted means" setting; they are both uncorrelated Gaussian with unit variances, while the means of $Q_S$ are shifted from $P_S$. 
The identical marginals $P_{\{1, \dots, D \} \backslash S} = Q_{\{1, \dots, D \} \backslash S}$ on the remaining variables are the zero-mean Dirac distribution, i.e., the remaining variables always take the value zero.
     
 \end{enumerate}

The discriminating set $S \subset \{1,\dots, D\}$ is randomly chosen.
We set the total number $D$ of variables to $20$ and the number of discriminating variables to $2$; see the supplementary materials for results on other settings, where qualitatively similar results are reported. 
We generate two i.i.d.\ samples  from $P$ and $Q$, each of size $n=200$. 
The task is to identify the discriminating set from these samples. 

We evaluate a set $\hat{S} \subset \{1, \dots, D \}$ of selected variables using three criteria: precision (Pre), recall (Re), and F score, defined as

\begin{equation*}
   {\rm Pr} = \frac{|\hat{S} \cap S|}{|\hat{S}|}, \hspace{2ex} 
  {\rm Re} = \frac{|\hat{S} \cap S|}{|S|} , \hspace{2ex}
   F = \frac{2 \times {\rm Pr} \times {\rm Re}}{{\rm Pr} + {\rm Re}} .
  \label{eq:f-score}
\end{equation*}
The precision is the ratio of the true discriminating variables among the selected variables, the recall is the ratio of the selected discriminating variables among all the discriminating variables, and the F score is their harmonic mean;  higher values 
indicate better variable selection performance. 
We repeat the above procedure 10 times independently, and compute averages and standard deviations of the criteria.

\subsection{Results}
\label{sec:synthetic-exp-results}

\begin{figure}[t]

  \centering
  \includegraphics[scale=0.35]{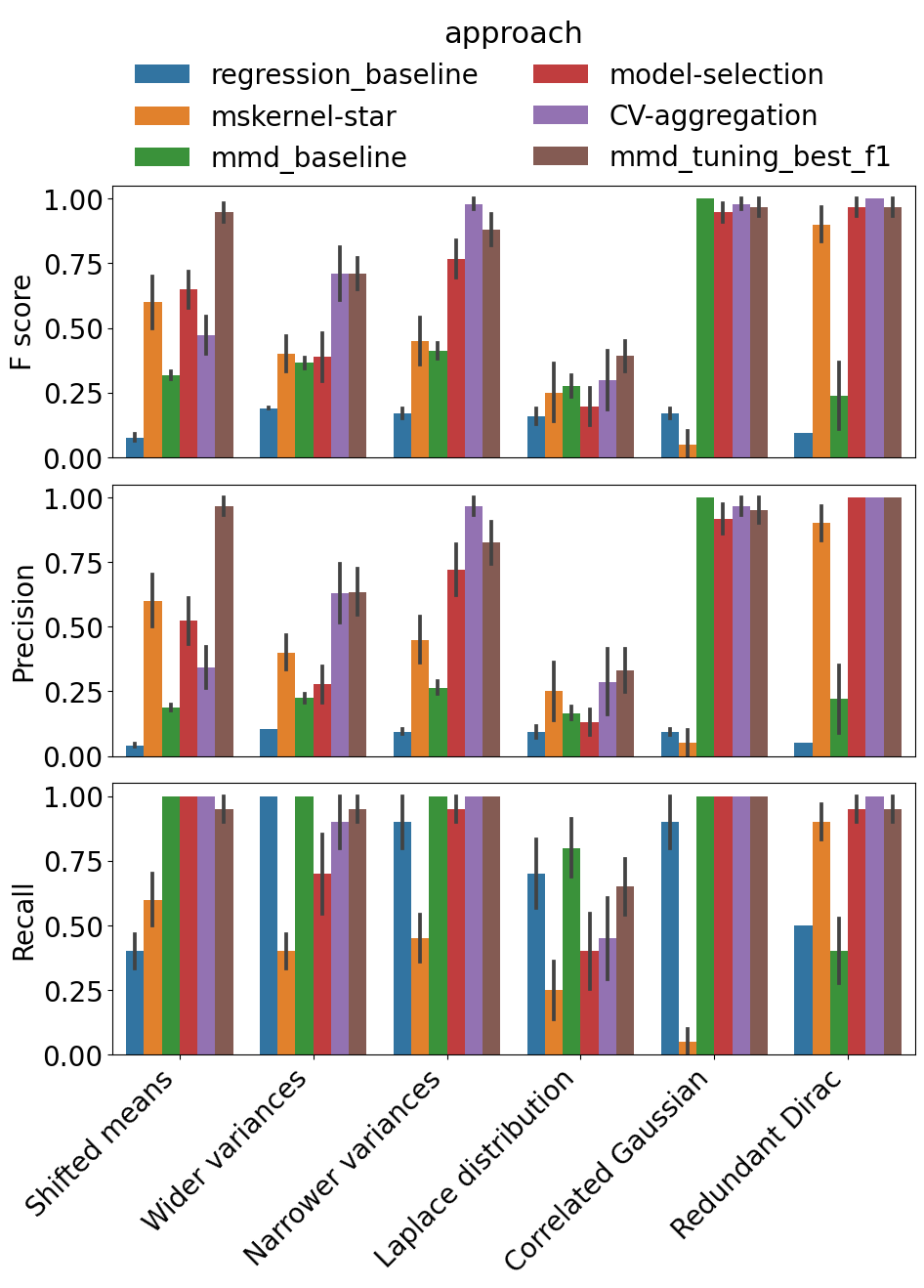}

  \caption{
Synthetic data experiment results (Section~\ref{sec:synthetic_data_assessment}).  The top, middle, and bottom panels report F score, precision, and recall, respectively. 
Groups correspond to the six settings. 
Bars show mean \(\pm\) standard deviation over 10 runs.}
    \label{fig:synthetic2-20-comparison}
\end{figure}

\begin{figure}[t]
  \centering
  \includegraphics[scale=0.28]{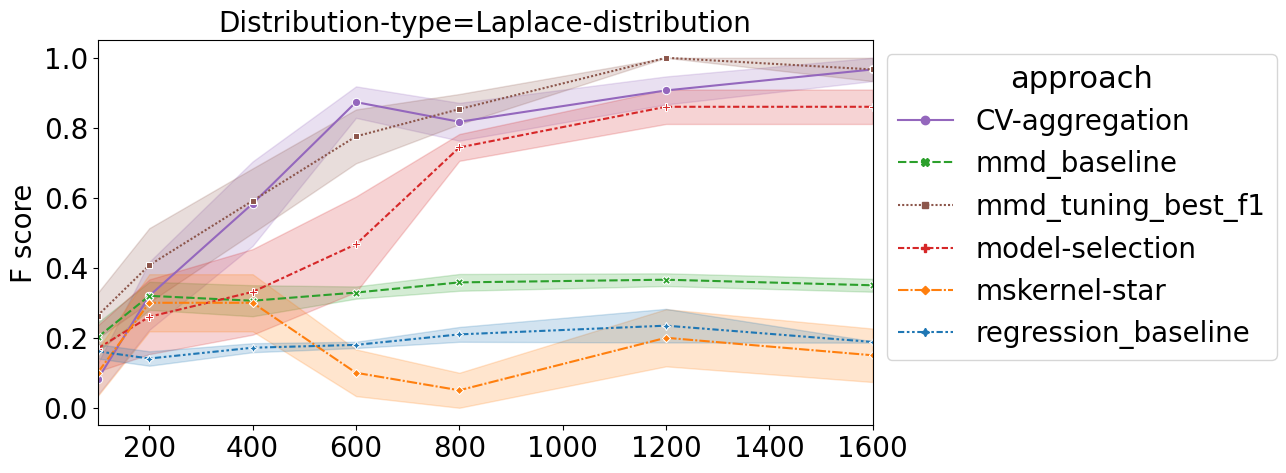}
  \caption{
  F scores for different sample sizes in the Laplace distribution setting in Section \ref{sec:synthetic_data_assessment}. The horizontal axis indicates sample sizes. For each sample size and each method, the confidence interval shows the standard deviation of the F scores over 10 
  experiments. 
  }
    \label{fig:synthetic2-sample-size-comparison}
\end{figure}

Figure~\ref{fig:synthetic2-20-comparison} summarises the results. The following observations can be made, which generally support the validity of Algorithm~\ref{alg-hyperparameter_selection} ({\tt model-selection}) and Algorithm~\ref{alg-enhanced-stability-selection} ({\tt CV-aggregation}): 
\begin{itemize}
    \item 
    {\tt model-selection} and {\tt CV-aggregation} attain relatively high F scores across different setups by balancing recall and precision, in comparison with the three baselines ({\tt regression-baseline}, {\tt mskernel-star} and {\tt mmd-baseline}). This supports the validity of the proposed approaches.  

    \item 
    For most setups, {\tt CV-aggregation} attains F scores comparable with (or sometimes even higher than) those of {\tt mmd-tuning-best-f1},  whose regularisation parameter was tuned to maximise the F score (thus not implementable in practice) to indicate the best possible performance for the regularised test power optimisation approach~\eqref{eq:regularized-opt}.     This suggests the effectiveness of aggregating the results from different regularisation parameters. 

    \item 
    {\tt mskernel-star} compares univariate marginals on each variable using MMD separately, so it should perform well for the setups other than 5) correlated Gaussian, given also that the true number of discriminating variables is given to this method in the experiments (again, this is an unfair advantage over others). 
    Our methods attain higher or comparable F-values in these setups, even though they know neither the true number of discrimination variables nor that discrepancies exist only in the univariate marginals. 

    \item 
    For the 5) correlated Gaussian setup, where discrepancies appear only in the correlation structures, no true discriminating variables are selected by {\tt mskernel-star}. This is reasonable as it only compares univariate marginals. Our methods and {\tt mmd-baseline} successfully select discriminating variables in this setup, demonstrating that the ARD-weight optimisation approach can identify discrepancies in correlation structures.

    \item 
    For the 6) redundant Dirac setup,  {\tt mmd-baseline} fails to remove the clearly redundant variables that deterministically take value zero for the two distributions, thus missing the discriminating variables.
    This failure is due to the lack of regularisation that penalises redundancy; see Section~\ref{sec:theory}. Our methods performing well in this setup demonstrates the necessity of regularisation in the ARD weight optimisation approach and the validity of the proposed methods for regularisation parameter selection and aggregation.

    \item For the 4) Laplace setup, discrepancies in the discriminating variables occur not in their means or variances but in their higher-order moments, so larger samples are needed to identify the discrepancies than in the other setups. Figure~\ref{fig:synthetic2-sample-size-comparison} shows the F scores of each method for different sample sizes.  The F scores of our methods increase with larger samples, suggesting that they can detect subtle discrepancies with sufficiently large samples. Notably, {\tt CV-aggregation} needs substantially fewer samples to attain the same level of F score than {\tt model-selection} (e.g., an F score of about 0.87 is attained by {\tt CV-aggregation} at a sample size of 600 and by {\tt model-selection} at a size of 1,200), demonstrating the sample efficiency of {\tt CV-aggregation}.
\end{itemize}

\section{Application to Water-Pipe Leakage Data}

\label{sec:simulation_water_pipe_network}

The current and next sections describe two demonstrations of two-sample variable selection using Algorithm~\ref{alg-enhanced-stability-selection} ({\tt CV-aggregation}), which performed best across different settings in Section~\ref{sec:empirical_assessment}. Its configuration follows Section~\ref{sec:empirical_assessment}. 

\begin{figure}[th]
    \centering
    \includegraphics[width=0.8\linewidth]{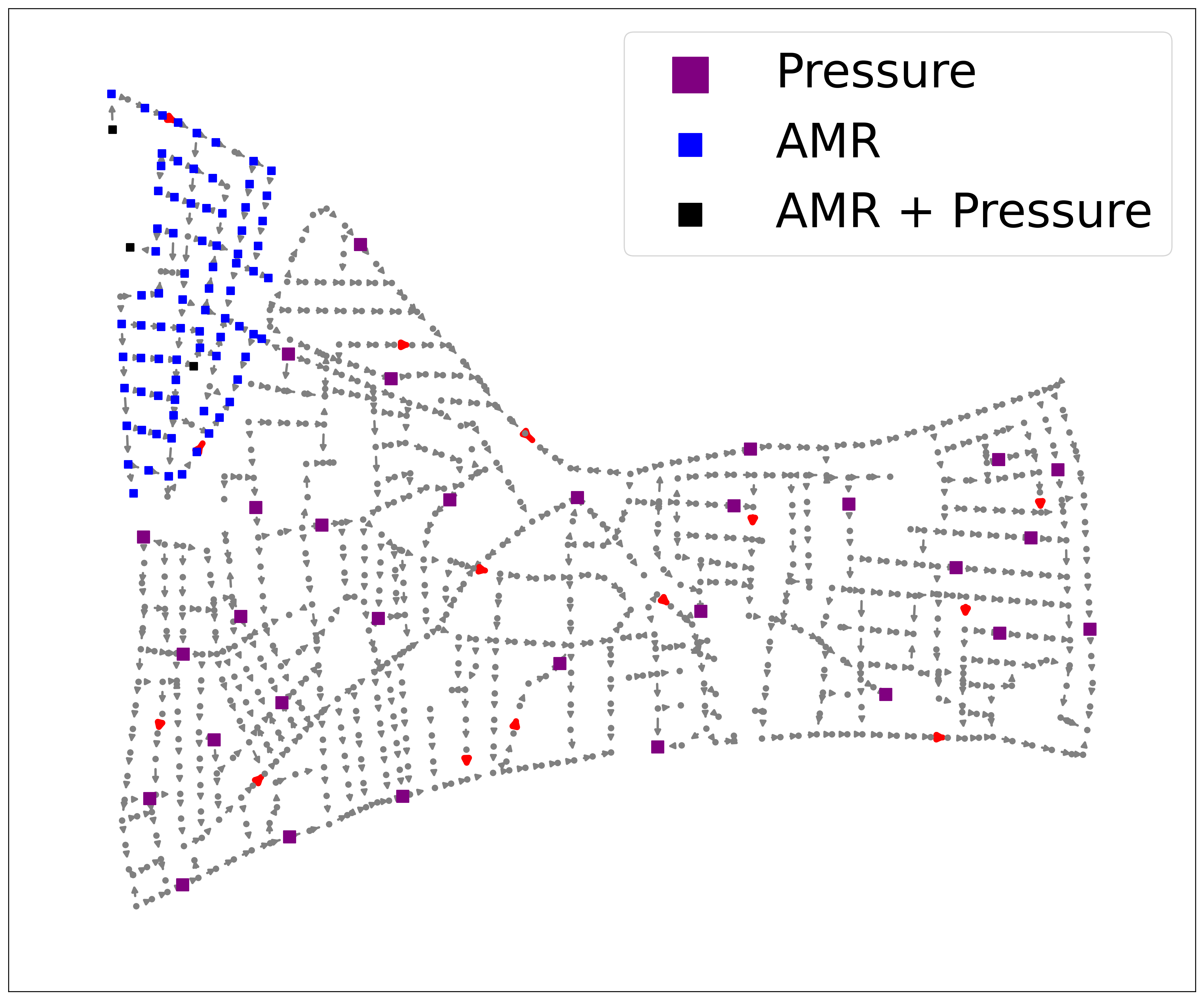}
    \caption{
L-Town network from the \texttt{BattLeDIM 2020} dataset, showing installed sensors (squares).
\texttt{AMR} and \texttt{Pressure} sensors are mounted at nodes; some nodes host both types.
Red triangles mark pipe segments with programmed leaks.
Leaks occur at different times.}
    \label{fig:water-pipe-network}
\end{figure}

Our first demonstration is the detection of sensors affected by leakages in a water-pipe network  \cite[e.g.,][]{Vrachimis-2022,Hinder-2023-model-based}.  
We use the \texttt{BattLeDIM 2020} dataset\footnote{\url{https://battledim.ucy.ac.cy/}}~\cite{Vrachimis-2022} generated with the Water Network Tool for Resilience (\texttt{WNTR}) \citep{Klise-2020water} by simulating water flows in the ``L-Town'' network in Figure~\ref{fig:water-pipe-network} over one year (2018/01/01–2018/12/31).
The network has sensors (``{\tt Pressure}'' and ``{\tt AMR}'') that log every 5 minutes.
Leaks are programmed on 14 pipes (red triangles in Figure~\ref{fig:water-pipe-network}) at different time points. 
Durations vary: some leaks are repaired soon after onset, while others persist much longer (see Figure~\ref{fig:water-leak-ground-truth} in Appendix~\ref{sec:demonstration-leaking-water-pipe-detection}).

We aim to identify sensors affected by a water-leakage event by treating each sensor as a variable.
We use observations from all \texttt{Pressure} sensors (33 in total; time points as rows, sensors as columns).\footnote{We use \texttt{Pressure} sensors because they are evenly distributed across the network, whereas \texttt{AMR} sensors are concentrated in one area.}
At an arbitrary given time point, we form two 7-day windows: one before and one after that point.
Within each window, we downsample to hourly means (12 five-minute readings per hour), yielding 168 hours.
For every hour, we construct a 33-dimensional feature vector by dividing each sensor’s hourly mean by the sensor value at the first point within the same window.
This yields two sample sets of 168 sample vectors, each with 33 variables: the “before” set $\mathbf{X}$ and the “after” set $\mathbf{Y}$.

\begin{figure}[th]
    \centering
    \includegraphics[scale=0.28]{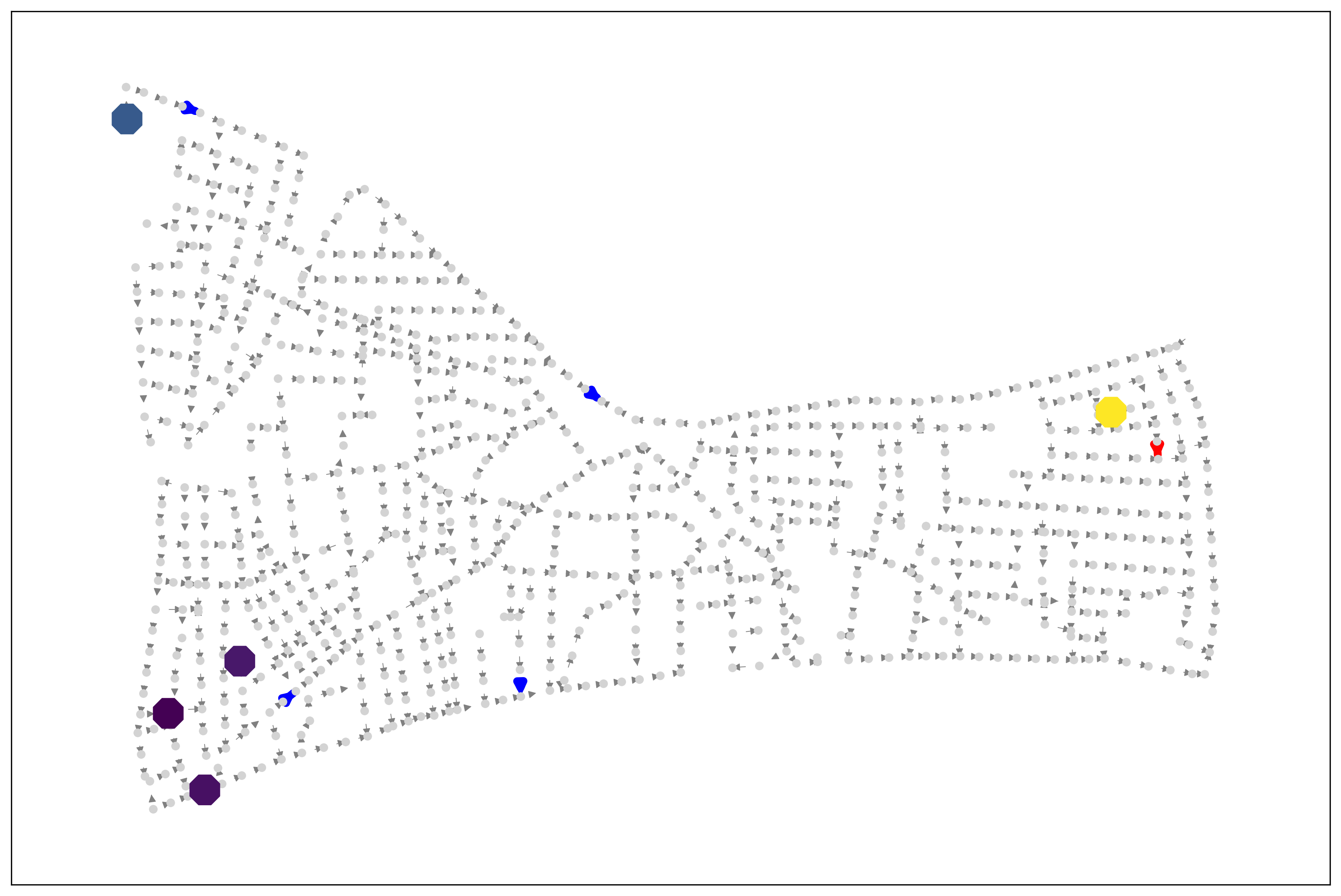}
    \caption{Location of pipe \texttt{p866} (red arrow), where a leak begins at 09:05 on June 1, 2018, and the selected sensors. The yellow circle marks the highest-weight sensor (\texttt{n296}); dark purple circles mark much lower-weight sensors (the darker the lower). Blue triangles indicate other leaking pipes persisting during the case-study windows.}
    \label{fig:case-study-p866-variable-selection-map}
\end{figure}

For demonstration, we set the inspection time to 00:00 on June 1, 2018, and use two seven-day windows: from 00:00 on May 25 to 23:55 on May 31 (before), and from 00:00 on June 1 to 23:55 on June 7 (after). The leak on pipe \texttt{p866} begins at 09:05 on June 1, 2018 (the red triangle in Figure~\ref{fig:case-study-p866-variable-selection-map}). Our goal is to identify the sensors whose readings are most affected by this event without knowing it.

Note that four other pipes are leaking during this window (see the blue triangles in Figure~\ref{fig:case-study-p866-variable-selection-map} and Figure~\ref{fig:water-leak-ground-truth} in Appendix~\ref{sec:demonstration-leaking-water-pipe-detection}). 
These leaks began well before the change point and are in steady state, so they are not expected to materially affect the pressure readings. 
Accordingly, the selection should capture the transient effects triggered by the onset of the {\tt p866} leak.

Algorithm~\ref{alg-enhanced-stability-selection} selected five sensors.
Figure~\ref{fig:case-study-p866-variable-selection-map} shows the selected sensors: the yellow circle marks the sensor (\texttt{n296}) with the highest weight in the variance-importance score vector $\hat{\Pi}$ of Algorithm~\ref{alg-enhanced-stability-selection}, located close to pipe {\tt p866}. 
Indeed, the samples in the two windows visually differ at this sensor (see Figure~\ref{fig:case-study-p866-observations} in Appendix~\ref{sec:demonstration-leaking-water-pipe-detection}).
Thus, Algorithm~\ref{alg-enhanced-stability-selection} identifies the sensor affected by the leak.

\begin{figure}[th]
    \centering
    \includegraphics[width=0.8\linewidth]{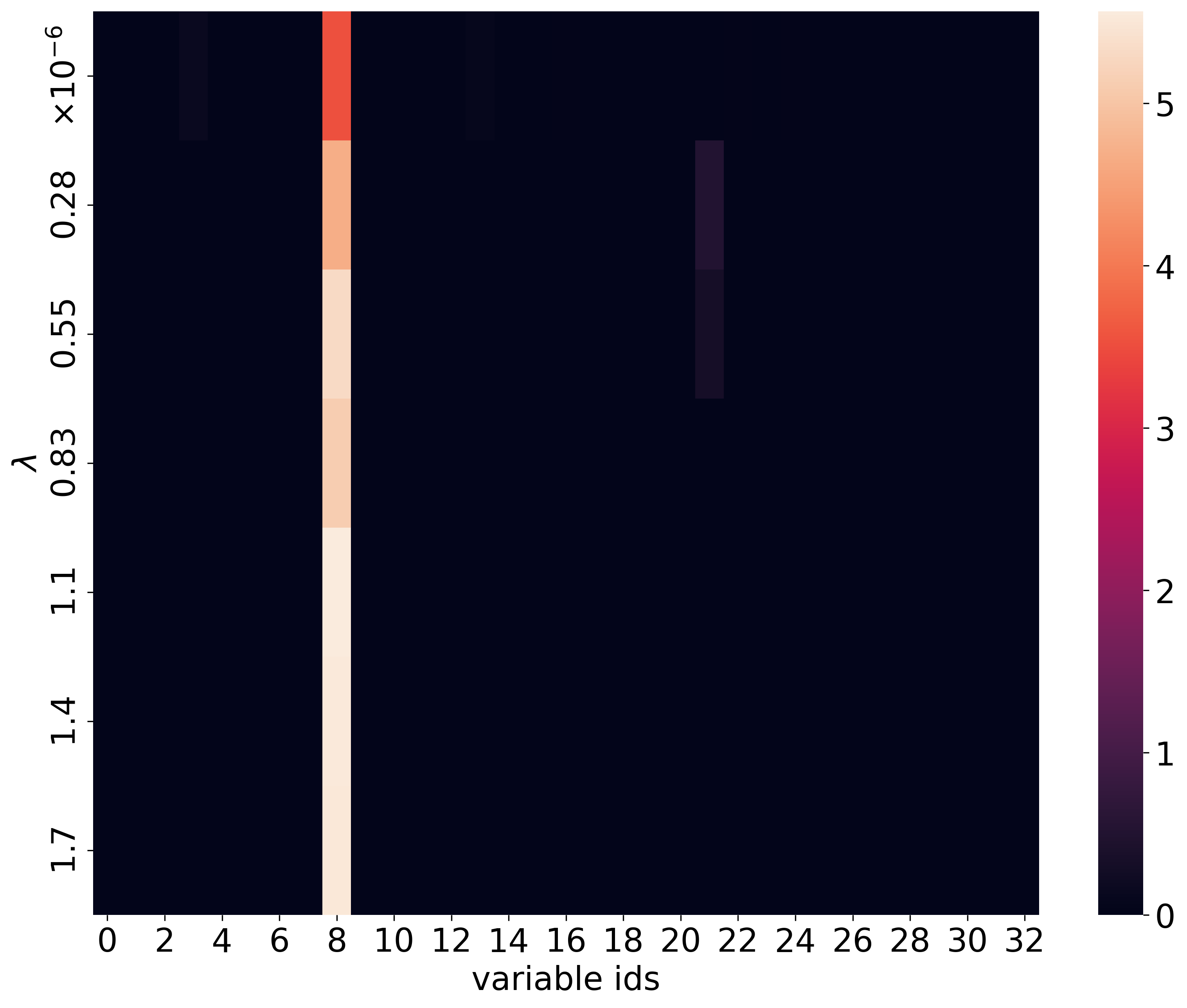}
\caption{
Variable-importance scores from Algorithm~\ref{alg-stability-selection} for the water-pipe experiment (Section~\ref{sec:simulation_water_pipe_network}).
Horizontal axis: variable IDs; vertical axis: regularisation parameter $\lambda$.
Each row shows the variable-importance vector $\hat{\Pi}_\lambda$ for one candidate $\lambda$.
Variable 8 is sensor \texttt{n296}, nearest to the leak.
}

    \label{fig:case-study-p866-variable-selection-hat-pi}
\end{figure}

Figure~\ref{fig:case-study-p866-variable-selection-hat-pi} displays the variable-importance scores $\hat{\Pi}_\lambda$ from Algorithm~\ref{alg-enhanced-stability-selection} across candidate regularization levels $\lambda$.
Variable ID 8 (sensor \texttt{n296}) has consistently high values over a wide range of $\lambda$, indicating that this sensor strongly contributes to distinguishing the “before” and “after” distributions of $\mathbf{X}$ and $\mathbf{Y}$.

\section{Application to Blocked Traffic Data}
\label{sec:simulated_traffic_dataset_body}

\begin{figure}
    \centering
    \includegraphics[width=0.4\linewidth]{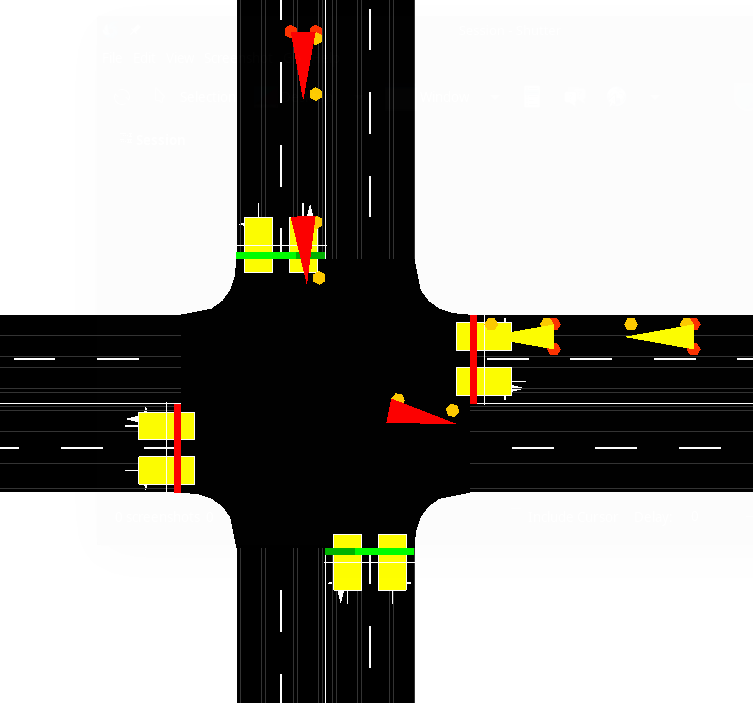}
    \caption{An intersection in the grid network (Section~\ref{sec:simulated_traffic_dataset_body}), showing eight induction-loop sensors (rectangles) and vehicles (triangles).}
    \label{fig:intersection-traffic}
\end{figure}

\begin{figure}[t]
  \centering
  \includegraphics[width=1.0\linewidth]{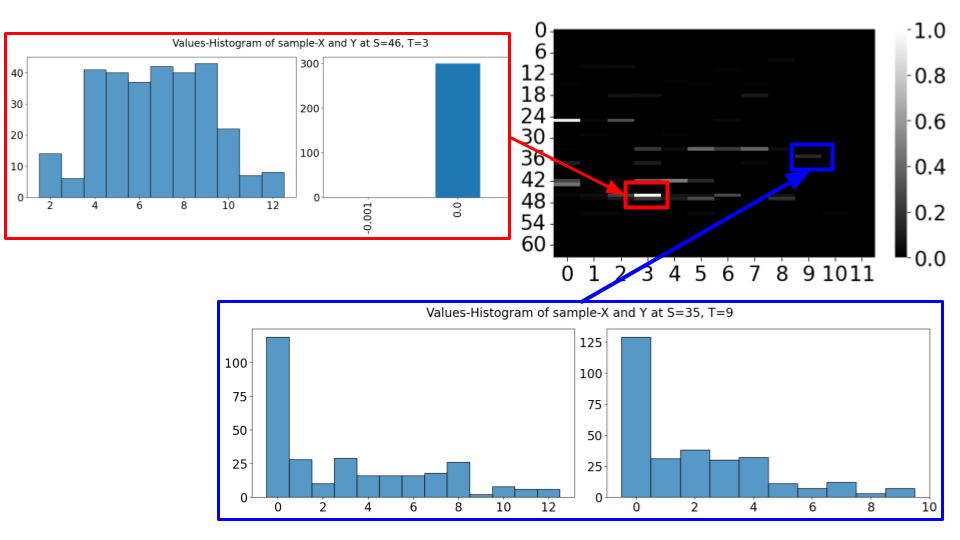}
  \caption{
Top-right heatmap: variable-importance scores $\hat{\Pi}$ (max scaled to 1) on the 64 sensors (vertical) by 12 time intervals (horizontal) from Algorithm~\ref{alg-enhanced-stability-selection}. 
Among the selected variables, red marks the highest score and blue the lowest.
For each highlighted sensor–time pair, the two histograms show the empirical vehicle-count distributions for the baseline (left) and perturbed (right) runs.
The x-axes of the histograms show vehicles per second over a 5-minute interval; for example, 0.02 corresponds to $0.02\times 300=6$ vehicles.
}

  \label{fig:sumo-hist-analysis-body}
\end{figure}

Our second demonstration of two-sample variable selection is the detection of the effects from road blockages in a traffic network, treating each pair of a sensor and a time interval as a variable.

We use the \texttt{SUMO} traffic simulator\footnote{SUMO v1.4.0: \url{https://www.eclipse.org/sumo/}}~\cite{Lopez-2018-microscopic} to model traffic flows in a grid-like network with eight intersections (see Figure~\ref{fig:appendix-grid-network-iilustration} in Appendix~\ref{sec:appendix-simulated_traffic_dataset}) for one hour. Each intersection has eight induction-loop sensors (Figure~\ref{fig:intersection-traffic}), totalling 64 sensors. Sensors record vehicle counts every five minutes, yielding 12 intervals; each run is summarised as a 64\,$\times$\,12 matrix of average counts per interval and sensor.

We consider two scenarios: (i) a baseline with two opposing traffic streams (north west $\leftrightarrow$ south east), and (ii) a perturbed case with two lanes blocked (see Figure~\ref{fig:appendix-grid-network-iilustration} in Appendix~\ref{sec:appendix-simulated_traffic_dataset}) during the first 40 minutes (intervals 0–7). 

Each scenario is run 400 times with distinct random seeds to capture stochastic behaviour, producing two datasets: $\mathbf{X}$ (baseline) and $\mathbf{Y}$ (perturbed), each a set of 64\,$\times$\,12 matrices.

We seek sensor–time pairs whose readings are affected by the perturbation, treating each of the $64\times 12=768$ sensor–time combinations as a variable. We apply Algorithm~\ref{alg-enhanced-stability-selection} with the configuration from Section~\ref{sec:assessment-opt-configuration}, except for kernel length scales: because many entries are zero (no vehicles), the variable-wise median heuristic can return zero length scales; we therefore use the variable-wise mean, which is positive in this setting.

A strict “ground truth” is hard to define here: the closure perturbs flows not only on the blocked links and not only during the 40-minute window, but also on surrounding links and subsequent periods. 
We therefore present an exploratory analysis using the proposed method.

Algorithm~\ref{alg-enhanced-stability-selection} yielded 22 selected variables (sensor–time pairs) out of 64×12 = 768, along with the variable-importance scores $\hat{\Pi}$ from Algorithm~\ref{alg-enhanced-stability-selection}. 
Figure~\ref{fig:sumo-hist-analysis-body} shows $\hat{\Pi}$ as a heatmap and highlights, among the selected variables, the highest- and lowest-scoring pairs (red and blue). 
For each highlighted pair, the accompanying histograms plot the empirical vehicle-count distributions in the two sample sets ${\bf X}$ and ${\bf Y}$.

At the highest-scoring sensor–time pair (red), traffic patterns differ sharply: the baseline histogram shows active flow, whereas the perturbed one shows none—consistent with a nearby closure. An analyst could then inspect adjacent sensors and neighbouring time intervals to confirm. The lowest-scoring selected pair (blue) is also informative: although the two histograms look similar, small shifts at the low and high ends suggest residual impacts near the end of the closure window. We view such flags as hypotheses that a domain expert can verify; in this sense, our method functions as a hypothesis-generation tool for incident forensics.

Temporally (heatmap’s horizontal axis), the selected variables align with the perturbation period: differences in counts emerge from the start of the simulation (index 0) and taper after 40 minutes (around index 8). 

Spatially (heatmap’s vertical axis), interpretation is harder because a road closure perturbs flows across the entire network, not just on the blocked links. 
Still, the selected sensors (e.g., indices 25, 42, 43, 46, 47) align with routes directly affected by the closure. 
Sensor 33 is also selected; it lies on a major corridor likely used for rerouting.

This analysis shows that the method can recover relevant variables—including subtle differences—capturing network-wide flow changes beyond the immediately blocked links.

\section{Conclusion and Future Directions}

This paper makes two main contributions to the emerging literature on two-sample variable selection. One contribution is mathematical, defining the notion of a discriminating set of variables and deriving its key properties, including uniqueness. This lays a foundation for two-sample variable selection as a general methodological framework. Our second contribution is methodological, proposing two practical methods for two-sample variable selection based on regularised MMD-power maximisation. Our methods address regularisation parameter selection by balancing recall (coverage) and precision, using the test power and the p-value of a permutation test on the selected variables as their proxies. The same idea could be used for hyperparameter selection in other two-sample variable selection methods.

There are many potential future directions. 
(i) Test power: While variable selection was our focus, it is also of interest to study, both theoretically and methodologically, whether a two-sample test performed on the selected variables has higher power than the same test performed on all the variables.  
(ii) Scalability: Computing the test power objective~\eqref{eq:regularized-opt} costs a quadratic complexity in the sample size, which can be expensive for large datasets. Incorporating linear-time MMD estimators~\cite{gretton2012optimal} or Fourier-feature approximations~\cite{Jitkrittum2016} would reduce the cost. 
(iii) Theory: Large sample asymptotics of the proposed methods, or two-sample variable selection in general, should be analysed. Conditions for the support consistency in recovering the discriminating variables should be established~\cite{wainwright2019high}. 
(iv) Other regularisers: The Lasso regularisation has many extensions, such as Elastic Net and Group Lasso, which can be used to incorporate known structural properties among variables~\cite{hastie2015statistical}.

\subsection*{Acknowledgement}

This work was done while Kensuke Mitsuzawa was affiliated with EURECOM, and he is currently supported by the French Government, through the NIMML project (ANR-21-CE23-0005-01).
We acknowledge support from Huawei Research and Development.

\bibliographystyle{IEEEtran}
\bibliography{Bibfile}

\vspace{-10mm}

\clearpage

\appendices

\section*{Supplementary Materials}

The supplementary materials are organised as follows.  
 Appendix~\ref{app:proofs} provides the proofs of theoretical results. 
 Appendix~\ref{sec:cats-dogs-demo}  demonstrates two-sample variable selection on face images of cats and dogs. 
 Appendices~\ref{sec:demonstration-leaking-water-pipe-detection} and \ref{sec:appendix-simulated_traffic_dataset} offer supplementary materials on the two experiments in the main body on the water leakage and traffic simulation datasets, respectively.
 Appendix~\ref{sec:length-scale-selection} explains how to select length-scale parameters in the ARD kernel. 
 Appendix~\ref{app:variable_selection} details how to select variables using optimised variable weights.
  Appendix~\ref{sec:reg-param-range-selection} describes how to obtain a set of candidate regularisation parameters. 
  Appendix~\ref{sec:appendix-syn2-result-parameter-choice} investigates the effects of the regularisation parameter on the variable selection performance.
   Appendix~\ref{sec:appendix-ss-analysis} compares different candidate aggregation strategies for   {\tt CV-aggregation}  (Algorithm~\ref{alg-enhanced-stability-selection}).
   Appendix~\ref{sec:appendix-syn2-result} reports additional results for the synthetic data experiments in Section~\ref{sec:empirical_assessment}. 
    Appendix~\ref{sec:linear-estimator-app} performs a preliminary experiment using a linear MMD estimator.  
    Appendix~\ref{sec:mmd-based-optimization} makes a comparison with an alternative regularised objective function where the MMD estimate replaces the ratio objective.

\section{Proofs of Theoretical Results} 
\label{app:proofs}

This section provides proofs of the theoretical results in the main body. 

\subsection*{Notation} 
Let $\mathbb{N}$ be the set of natural numbers, $\mathbb{R}$ be the real line and $\mathbb{R}^D$ with $D \in \mathbb{N}$ be the $D$-dimensional Euclidean space. 
For $x  \in \mathbb{R}^D$, $x^\top$ denotes its transpose and $\| x \| = \sqrt{\sum_{i=1}^D x_i^2}$ its norm, where $x_i$ is the $i$-th element of $x$.
For $D \in \mathbb{N}$, let ${\bf 0}_D := (0,\dots, 0)^\top \in \mathbb{R}^D$ be the $D$-dimensional zero vector.
For $x \in \mathbb{R}$, let $\lfloor x \rfloor$ be its integer part.

For $S \subset \{1, \dots, D\}$, let $|S|$ be its cardinality and $\{1, \dots, D\} \backslash S := \{ d \in \{1, \dots, D\} \mid d \not\in S \}$ be its complement. We may write $S^c := \{1, \dots, D\} \backslash S$.
For a vector $x \in \mathbb{R}^D$ and a probability distribution $P$ on $\mathbb{R}^D$, let $x_S = (x_i)_{i \in S} \in \mathbb{R}^{|S|}$ be the subvector of $x$ with coordinates in $S$, and $P_S$ be the marginal distribution of $P$ on $S$, i.e., $P_S(x_S) := \int P(x) dx_{S^c}$.
When $S = \{d \}$ for $d \in \{1, \dots, D\}$, we may write $P_d = P_S$.
For $U \subset \{1, \dots, D \}$ with $U \cap S = \phi$, let $P_S \otimes P_U$ be the product distribution of the two marginal distributions $P_S$ and $P_U$.

\subsection{Proof of Proposition \ref{prop:problem-formulation} }

\label{sec:prop-problem-form}

\begin{proof}
Write $[D]=\{1,\dots,D\}$. 
We first prove the existence of a discriminating set and the decomposition~\eqref{eq:decomposition}. \\

\noindent
\emph{\bf Existence and Decomposition.}
Let $U\subset [D]$ be maximal (by inclusion) such that, under each distribution,
(i) the variables in $U$ are independent of those in $U^c:=[D]\setminus U$, and
(ii) the marginal distributions on $U$ are identical under the two distributions.
Such a maximal $U$ exists because $[D]$ is finite. Set $S:=U^c$.
Note that $U$ can be the empty set, in which case $S = [D]$.
\begin{itemize}
    \item \emph{No redundant subset for $S$:}
If there were a nonempty $U'\subset S$ whose marginal distributions are identical
under the two distributions and which is independent of $S\setminus U'$ within each
distribution, then $U\cup U'$ would also satisfy (i)–(ii), contradicting the maximality
of $U$. Hence $S$ has no redundant subset.

\item
\emph{Maximality for $S$:}
Let $S'=S\cup U'$ with arbitrary $U'\subset U$ and $U'\neq\emptyset$, so that $S'$ is a strict superset of $S$.
Because $P_U=Q_U$, taking marginals gives $P_{U'}=Q_{U'}$. Because $U$ is
independent of $U^c=S$ under each distribution, $U'$ is independent of $S$ under each distribution.
Therefore, $U'$ is a nonempty redundant subset of $S'$, so $S'$ is not a discriminating set. Since every strict superset $S'\supset S$ is not a discriminating set, $S$ is maximal.

\end{itemize}

Therefore, $S$ satisfies Definition~\ref{def:subset}, proving the existence of a discriminating set.
The decomposition~\eqref{eq:decomposition} follows from the definition of this $S$.
\\ 

\noindent
\emph{\bf Uniqueness.}
Suppose $S_1,S_2\subset [D]$ both satisfy Definition~\ref{def:subset}, and set $S_0:=S_1\cup S_2$.
We show that $S_0$ also satisfies the definition.

Assume for contradiction that $S_0$ has a nonempty redundant subset $U_0$ (identical marginals
on $U_0$ and independence from $S_0\setminus U_0$ within each distribution). At least one of
$U_0\cap S_1$ or $U_0\cap S_2$ is nonempty; say $U_1:=U_0\cap S_1\neq\emptyset$. 
Since $P_{U_0}=Q_{U_0}$, taking marginals gives $P_{U_1}=Q_{U_1}$. Moreover, from
$P_{S_0}=P_{U_0}\otimes P_{S_0\setminus U_0}$ and
$Q_{S_0}=Q_{U_0}\otimes Q_{S_0\setminus U_0}$, marginalizing to $S_1$ yields
\[
P_{S_1}=P_{U_1}\otimes P_{S_1\setminus U_1},\qquad
Q_{S_1}=Q_{U_1}\otimes Q_{S_1\setminus U_1},
\]
so $U_1$ is independent of $S_1\setminus U_1$ under each distribution. Hence $U_1$
is redundant in $S_1$, a contradiction.

Hence, $S_0$ has no redundant subset. By maximality in the definition, we must
have $S_1=S_0$ and similarly $S_2=S_0$, so $S_1=S_2$.
This proves uniqueness.

\end{proof}

\subsection{Proof of Corollary~\ref{coro:p-equal-q}}

\label{sec:proof-p-equal-q}

\begin{proof}

Suppose that the unique discriminating set is the empty set: $S = \emptyset$.  
By Proposition~\ref{prop:problem-formulation}, there exists a subset $U \subset \{1, \dots, D\}$ such that $S = \{1, \dots, D\} \backslash U$ and $P_U = Q_U$, which imply $U = \{1, \dots, D\}$ and thus $P = Q$.  
On the contrary, suppose $P = Q$. Setting $U = \{1, \dots, D\}$ in Proposition~\ref{prop:problem-formulation} implies $S = \emptyset$.

\end{proof}

\subsection{Proof of Proposition \ref{prop:max-mmd-weights}}

\label{sec:proof-mmd-max-weights}

\begin{proof}

Let $X, X' \sim P$ be independent vectors from $P$, and write $X = (X_S, X_U)$ and $X' = (X'_S, X'_U)$. 
Note that, $X_S$ and $X_U$ are independent, and so are $X'_S$ and $X'_U$, because $P = P_S \otimes P_U$ by assumption. 
Likewise,  let $Y, Y' \sim Q$ be independent vectors from $Q$, and write $Y = (Y_S, Y_U)$ and $Y' = (Y'_S, Y'_U)$. 
Then $X_S$ and $X_U$ are independent, and so are $X'_S$ and $X'_U$, because $Q = Q_S \otimes Q_U$ by assumption. 
As $P_U = Q_U$,  we have $X_U$, $X'_U$, $Y_U$ and $Y'_U$ are i.i.d., which implies that $\mathbb{E}[ k_U(X_U, X'_U) ] = \mathbb{E}[ k_U(Y_U, Y'_U) ] = \mathbb{E}[ k_U(X_U, Y_U) ]$.
Using these properties, we have
\begin{align} 
&    {\rm MMD}_k^2(P,Q) \nonumber \\
& = \mathbb{E}[k(X, X')] + \mathbb{E}[k(Y, Y')]   - 2\mathbb{E}[k(X, Y)] \nonumber \\
& = \mathbb{E}[k_S(X_S, X_S') k_U(X_U, X'_U)]  + \mathbb{E}[k_S(Y_S, Y_S') k_U(Y_U, Y'_U)] \nonumber \\
& \quad - 2 \mathbb{E}[k_S(X_S, Y_S) k_U(X_U, Y_U)] \nonumber \\
& = \mathbb{E}[k_S(X_S, X_S') ] \mathbb{E}[ k_U(X_U, X'_U)] \nonumber  \\
& \quad + \mathbb{E}[k_S(Y_S, Y_S')] \mathbb{E}[[ k_U(Y_U, Y'_U)] \nonumber \\
& \quad - 2 \mathbb{E}[k_S(X_S, Y_S)] \mathbb{E}[ k_U(X_U, Y_U)] \nonumber \\
& = \mathbb{E}[ k_U(X_U, X'_U)] \left( \mathbb{E}[k_S(X_S, X_S') ]   + \mathbb{E}[k_S(Y_S, Y_S')] \right. \nonumber \\
& \quad \quad  \left. - 2 \mathbb{E}[k_S(X_S, Y_S)]   \right) \nonumber \\
& = \mathbb{E}[ k_U(X_U, X'_U)] {\rm MMD}_{k_S}^2(P_S,Q_S), \nonumber
\end{align}
which proves the first assertion.
For the second assertion, we have
\begin{align*}
   & \max_{a_U \in \mathbb{R}^{|U|}} \mathbb{E}[ k_U(X_U, X'_U)] \\
   & = \max_{a_U \in \mathbb{R}^{|U|}} \mathbb{E}\left[ \prod_{d \in U} \phi_d \left( \sqrt{ a_d^2 (X_d - X_d')^2  / D} \right) \right] \\
   & \leq \prod_{d \in U} \phi_d \left( 0 \right),
\end{align*}
where the last inequality follows from each $\phi_d$ being a monotonically decreasing function.
Therefore 
\begin{align*}
 &   \max_{a \in \mathbb{R}^D} {\rm MMD}_k^2(P,Q)  \\
 & =   \max_{a_U \in \mathbb{R}^{|U|}} \mathbb{E}[ k_U(X_U, X'_U)] \max_{a_S \in \mathbb{R}^{|S|}} {\rm MMD}_{k_S}^2(P,Q)  \\
 & \leq  \prod_{d \in U} \phi_d \left( 0 \right)    \max_{a_S \in \mathbb{R}^{|S|}} {\rm MMD}_{k_S}^2(P,Q),
\end{align*}
which proves the assertion.
\end{proof}

\subsection{Proof of Proposition \ref{eq:prop-MMD-weights_maximizer}}

\label{sec:proof-mmd-max-weights-maximizer}
 
\begin{proof}

Let $V  \subset U$ be such that for each $d \in V$, the variance of the marginal distribution $P_d\ (= Q_d)$ is not zero. 
Let 
\begin{equation} \label{eq:maxim-700}
a^* := (a^*_S, a^*_{\SC}) \in \arg\max_{a \in \mathbb{R}^D} {\rm MMD}_k^2(P,Q). 
\end{equation}
By Proposition \ref{prop:max-mmd-weights}, we have
\begin{align*}
    & a_S^* \in  \arg\max_{a_S \in \mathbb{R}^{|S|}} {\rm MMD}_{k_S}^2(P_S, Q_S), \\
    & a_\SC^* \in  \arg\max_{a_{\SC} \in \mathbb{R}^{| \SC |} } \mathbb{E}[ k_U(X_U, X'_U)]. 
\end{align*}

For a random variable $Z \in \mathbb{R}$, denote by ${\rm Var}[Z]$ the variance of $Z$. 
Then, by assumption, we have ${\rm Var}[X_i] = {\rm Var}[X'_i] > 0$ for $i \in V\subset T$. 
This implies that
\begin{align*}
    \mathbb{E}[(X_i - X_i')^2] = 2 {\rm Var}[X_i] > 0, 
\end{align*}
where $X_i'$ is an independent copy of $X_i$. 
Therefore, there exists a constant $\varepsilon_i > 0$ such that 
\begin{equation} \label{eq:non-zero-var-718}
{\rm Pr} ( (X_i - X_i')^2 > \varepsilon_i^2 ) > 0.    
\end{equation}

We will show that  $a_i^* = 0$ for $i \in V \subset T$ with proof by contradiction. To this end, suppose $a_i^* \not= 0$. Then,  
\begin{align*}
& \max_{a_{\SC} \in \mathbb{R}^{| \SC |} } \mathbb{E}[ k_U(X_U, X'_U)] \\
& = \mathbb{E} \left[ \prod_{d \in U} \phi_d \left( \sqrt{ (a^*_d)^2 (X_d - X_d')^2  / D} \right) \right] \\
& = {\rm Pr} ( (X_i - X_i')^2 > \varepsilon^2_i )  
\\ & \quad \times \mathbb{E}\left[  \prod_{d \in U} \phi_d \left( \sqrt{ (a^*_d)^2 (X_d - X_d')^2  / D} \right)  \left| (X_i - X_i')^2 > \varepsilon \right. \right] \\
& + {\rm Pr} ( (X_i - X_i')^2 \leq \varepsilon_i^2 ) \\
& \quad \times \mathbb{E}\left[  \prod_{d \in U} \phi_d \left( \sqrt{ (a^*_d)^2 (X_d - X_d')^2  / D} \right)  \left| (X_i - X_i')^2 \leq \varepsilon_i^2 \right. \right] \\
& \stackrel{(A)}{\leq} {\rm Pr} ( (X_i - X_i')^2 > \varepsilon_i^2 ) \phi_i \left( a_i^* \varepsilon_i / \sqrt{D} \right)  \prod_{d \in U \backslash \{ i \}} \phi_d \left( 0 \right)  \\
& \quad + {\rm Pr} ( (X_i - X_i')^2 \leq \varepsilon_i^2 )  \prod_{d \in U} \phi_d \left( 0 \right)  \\
& \stackrel{(B)}{<}  \prod_{d \in U} \phi_d \left( 0 \right) = \max_{a\in \mathbb{R}^{|U|}} \mathbb{E}[k_U(X_U, X'_U)]
\end{align*}
where $(A)$ follows from $\phi_i$ being a monotonically decreasing function (as it is a positive definite function) and $a^*_i > 0$, and $(B)$ from \eqref{eq:non-zero-var-718} and $ \phi_{k_\SC} \left( a_i^* \varepsilon \right / \sqrt{D}) < \phi_i (0)$. 
This contradicts that $a^*_1, \dots, a^*_D$ are a maximiser in \eqref{eq:maxim-700}, implying that the assumption $a_i^* \not= 0$ is false. Therefore $a_i^* = 0$ for $i \in V \subset T$.

\end{proof}

\section{Demonstration: Analysing Cat and Dog Images}
\label{sec:cats-dogs-demo}

\begin{figure}[t]
  \begin{tabular}{cc}
  \begin{minipage}[t]{0.49\hsize}
      \centering
      \includegraphics[scale=0.2]{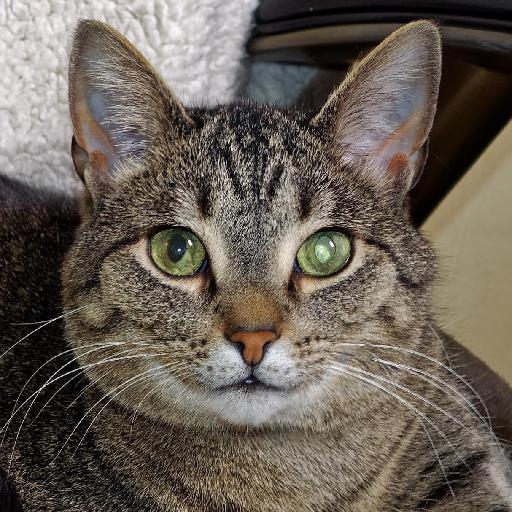}
      \subcaption{flickr\_cat\_000060.jpg}
  \end{minipage} &
  \begin{minipage}[t]{0.49\hsize}
      \centering
      \includegraphics[scale=0.2]{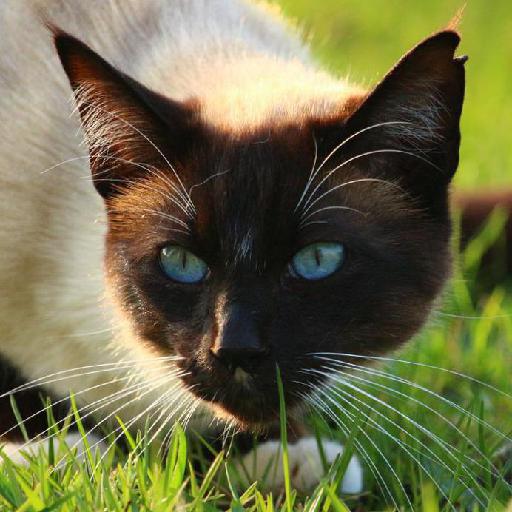}
      \subcaption{pixabay\_cat\_000588.jpg}
  \end{minipage} \\
 
  \begin{minipage}[t]{0.49\hsize}
      \centering
      \includegraphics[scale=0.2]{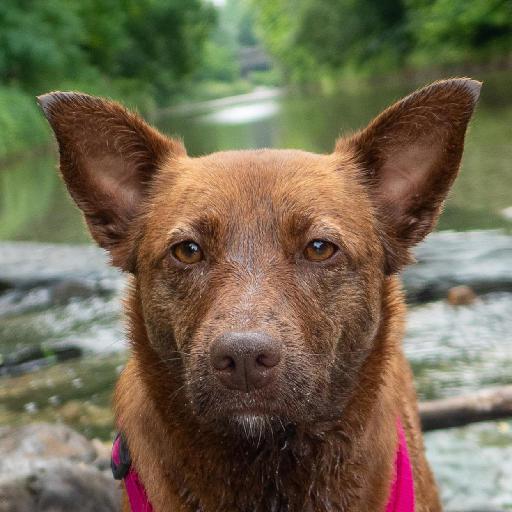}
      \subcaption{flickr\_dog\_000039.jpg}
  \end{minipage} &
  \begin{minipage}[t]{0.49\hsize}
      \centering
      \includegraphics[scale=0.2]{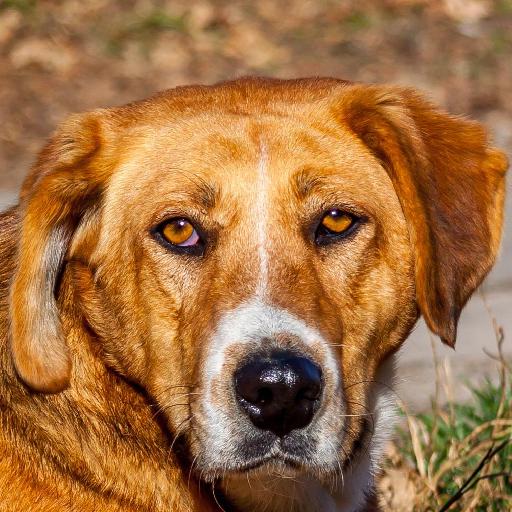}
      \subcaption{pixabay\_dog\_000969.jpg}
  \end{minipage}
  \end{tabular}
  \caption{
  Examples of cats' and dogs' face images from the AFHQ dataset. The caption of each image indicates the file name in the dataset.
  }
  \label{fig:dog-cat-afhq}
\end{figure}

We demonstrate the application of the proposed approaches in analysing image datasets. We use a subset of the AFHQ dataset~\cite{Choi_2020_CVPR} consisting of high-resolution images of cats' and dogs' faces, as described in Figure~\ref{fig:dog-cat-afhq}. The aim here is to select variables (= pixel coordinates) that indicate differences between cats' and dogs' faces.

As preprocessing, we downscale the resolution of each image data to $64 \times 64$ pixels, resulting in a total of $4,096$ dimensions. We convert each pixel's RGB values into a greyscale value ranging from $0$ to $255$.
For our experiments, we randomly selected 1,000 images from the AFHQ dataset containing 5,000 images. 
Here, we optimise the ARD weights using the Adam optimiser using 100 batches in each iteration. 
See Section~\ref{sec:assessment-opt-configuration} for other details.

\begin{figure*}[htbp]
  \begin{tabular}{cc}
  \begin{minipage}[t]{0.49\hsize}
      \centering
      \includegraphics[scale=0.5]{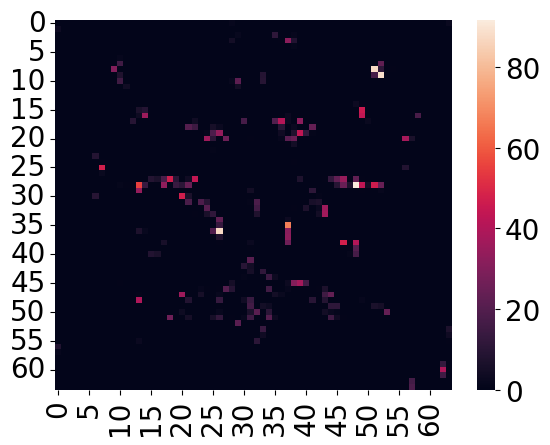}
      \subcaption{Optimised ARD weights obtained by {\tt mmd-baseline}}
  \end{minipage} &
  \begin{minipage}[t]{0.49\hsize}
      \centering
      \includegraphics[scale=0.5]{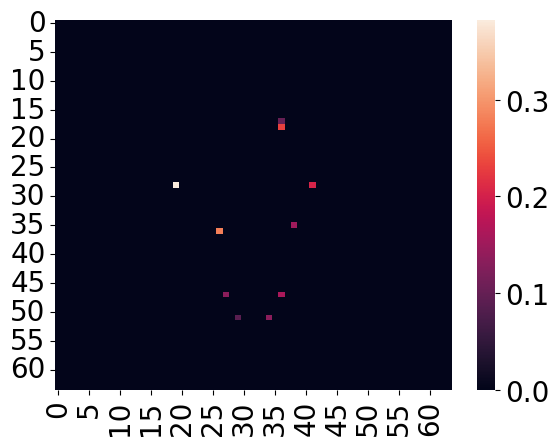}
      \subcaption{Optimised ARD weights obtained by {\tt model-selection}}
  \end{minipage} \\
 
  \begin{minipage}[t]{0.49\hsize}
      \centering
      \includegraphics[scale=0.5]{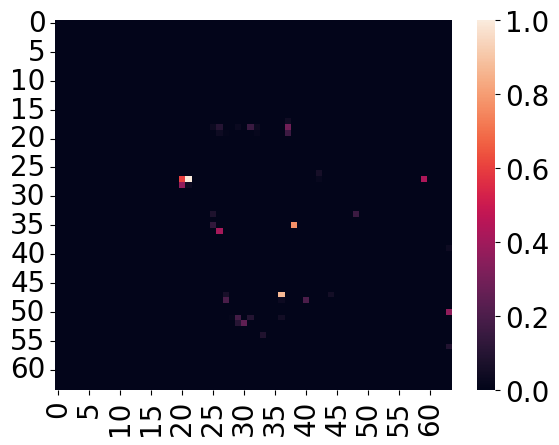}
      \subcaption{The score matrix obtained from {\tt CV-aggregation} }
  \end{minipage} &
  \begin{minipage}[t]{0.49\hsize}
  \end{minipage}
  \end{tabular}
  \caption{
  Score matrices obtained with (a) {\tt mmd-baseline}, (b) {\tt model-selection} and (c) {\tt CV-aggregation}.
  }
  \label{fig:dog-cat-variables}
\end{figure*}

Figure~\ref{fig:dog-cat-variables} describes the score matrices obtained with {\tt mmd-baseline}, {\tt model-selection} and {\tt CV-aggregation}. As {\tt CV-aggregation} performs the best in our other assessments, we examine the variables selected by {\tt CV-aggregation}. Figure~\ref{fig:dog-cat-afhq-cv-agg-masked} shows some cats' and dogs' face images on which the selected variables (pixel positions) are highlighted as yellow dots.

By examining Figure~\ref{fig:dog-cat-afhq-cv-agg-masked}, we can observe that the selected variables capture discrepancies between cats' and dogs' faces in mainly four areas: 1) the shape of the left eye, 2) the shape of the right eye, 3) the shape from the nose to the chin, and 4) the position of the forehead. Regarding eye shape, cats' eyes are typically wide and oval, while dogs' eyes tend to be elongated and elliptical. 
In addition to the shapes and positions of facial parts, colour depths (i.e., greyscale values from $0$ to $255$) may also contribute to discrepancies between cats' and dogs' faces. For example, cats' eyes tend to be transparent (resembling glass), while dogs' eyes are darker. Differences between cats and dogs in the shape from the nose to the chin are also noticeable. A cat's nose is small and pointy, and its chin shapes a gentle curve, while a dog has a larger and rounded black-blob-shaped nose and a more acute chin.

\begin{figure*}[htbp]
  \begin{tabular}{cc}
  \begin{minipage}[t]{0.49\hsize}
      \centering
      \includegraphics[scale=0.4]{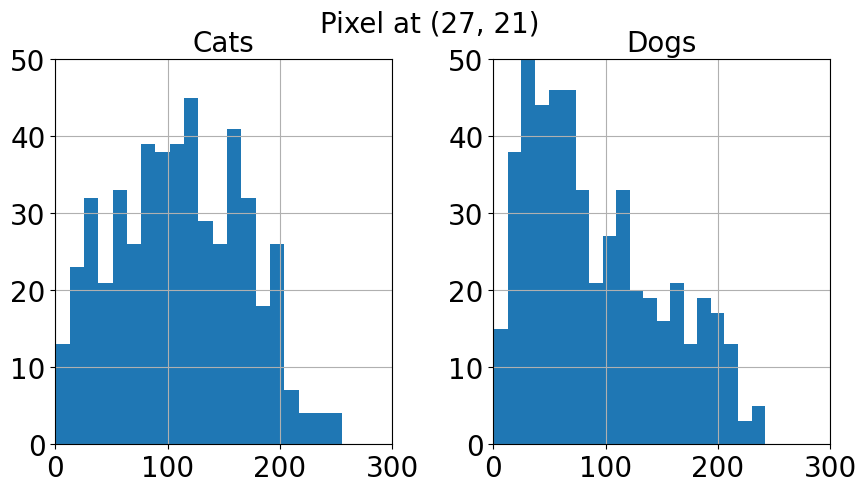}
  \end{minipage} &
  \begin{minipage}[t]{0.49\hsize}
      \centering
      \includegraphics[scale=0.4]{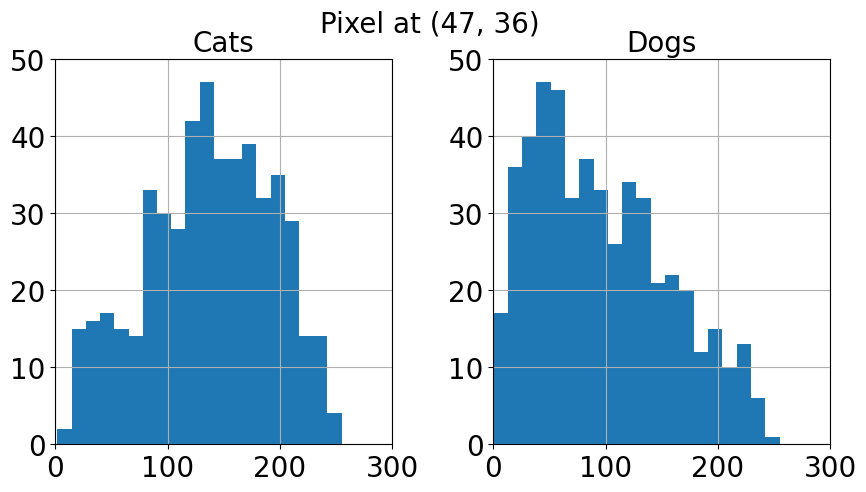}
  \end{minipage}
  \end{tabular}
  \caption{
  Histograms of greyscale values for the pixel at the 27th row and the 21st column (left two histograms) and those at the 47th row and 36th row (right two histograms).
  }
  \label{fig:histogram-cat-and-dog}
\end{figure*}

For further analysis, we compare the histograms of greyscale values in the datasets at specific pixel positions: the pixel at the 27th row and the 21st column (located on the upper left of the left eye) and at the pixel at the 47th row and the 36th column (located on the right side of the nose). These pixels have the highest and second scores given by {\tt CV-aggregation}. Figure~\ref{fig:histogram-cat-and-dog} shows the histograms at these pixels. Examining the histogram of the pixel at the 27th row and the 21st column, we observe that dog images have higher frequencies in low greyscale values than cat images. Since low greyscale values indicate darker colours, this may indicate the presence of dark iris colours in dogs' eyes. Similarly, for the pixel at the 47th row and the 36th column, dog images have higher frequencies in low greyscale values (dark colours). This pixel is often associated with a part of the nose, implying differences in the nose shapes and colours between cats and dogs.

\section{Supplementary Materials for Leaking Water Pipe Detection in Section~\ref{sec:simulation_water_pipe_network}}
\label{sec:demonstration-leaking-water-pipe-detection}

\begin{figure}[th]
    \includegraphics[width=0.95\linewidth]{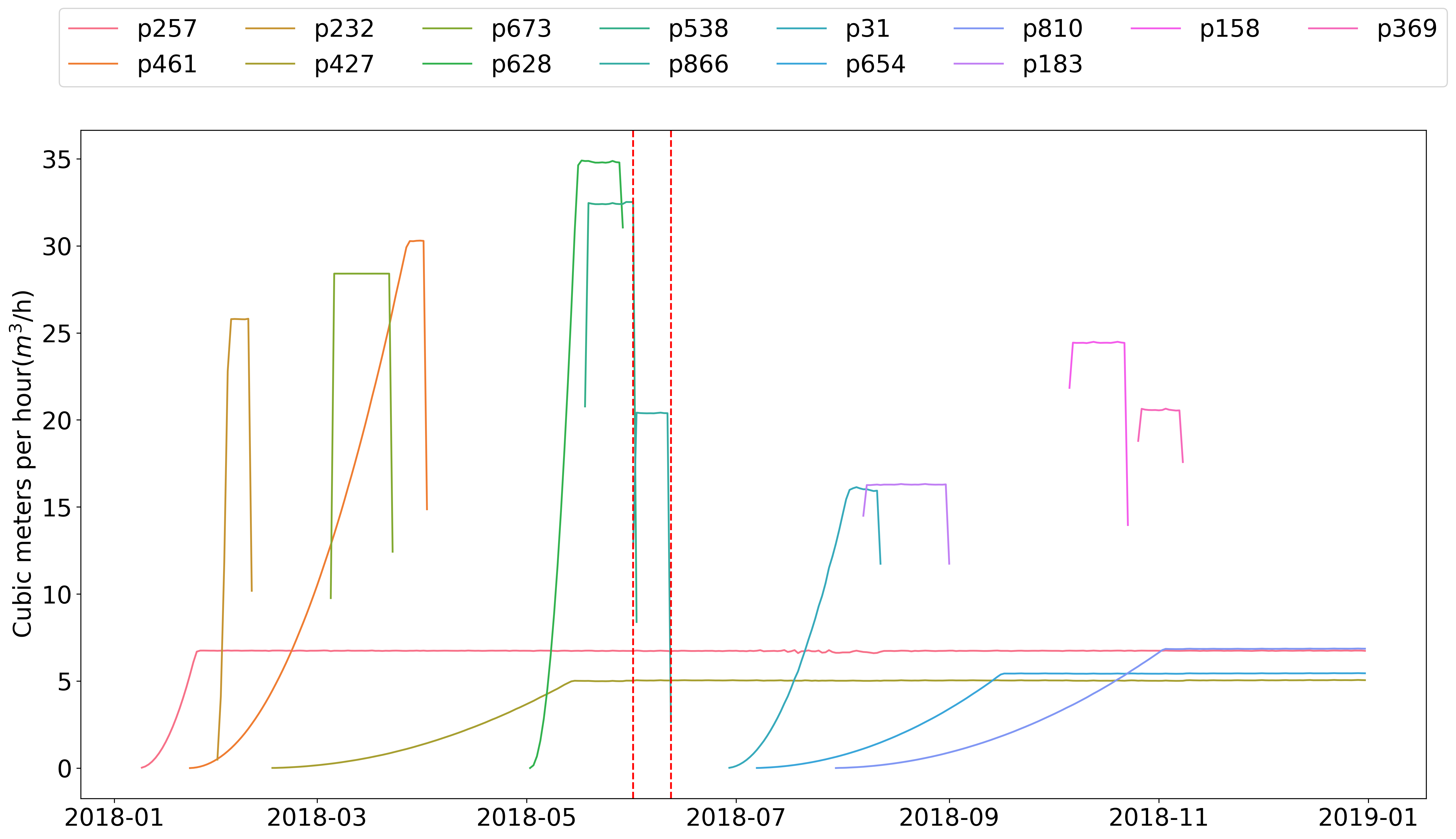}
\caption{
Fourteen leaking pipes in the \texttt{BattLeDIM 2020} dataset.
The legend lists the pipe IDs with leaks.
The horizontal axis is time; the vertical axis is leakage rate (per hour). 
Vertical red dashed lines highlight pipe \texttt{p866}, the focus of our case study.
}
\label{fig:water-leak-ground-truth}
\end{figure}

\begin{figure}[th]
    \centering
    \begin{minipage}[t]{0.9\hsize}
        \includegraphics[scale=0.3]{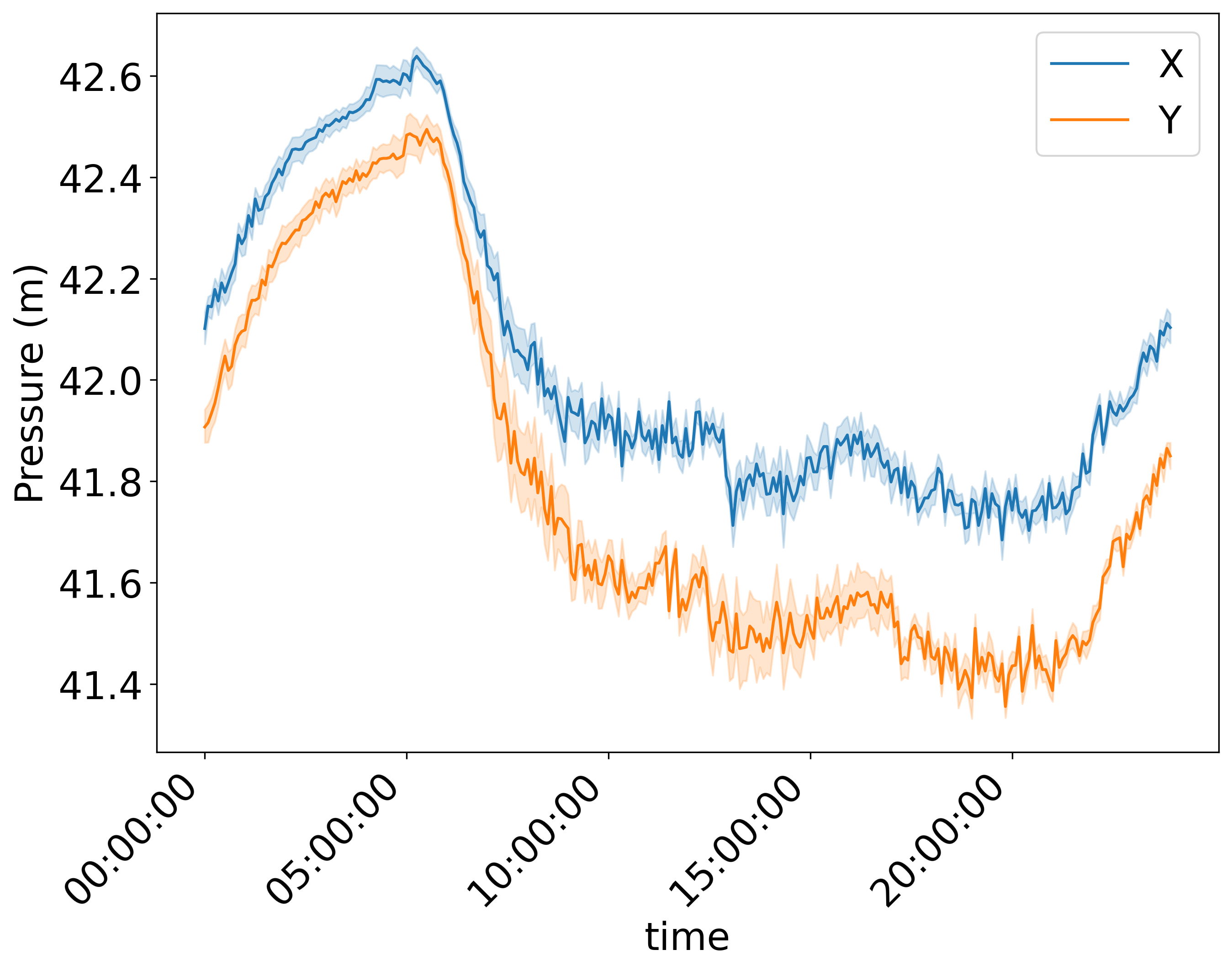}
    \end{minipage} \\
    \begin{minipage}[t]{0.9\hsize}
        \includegraphics[scale=0.3]{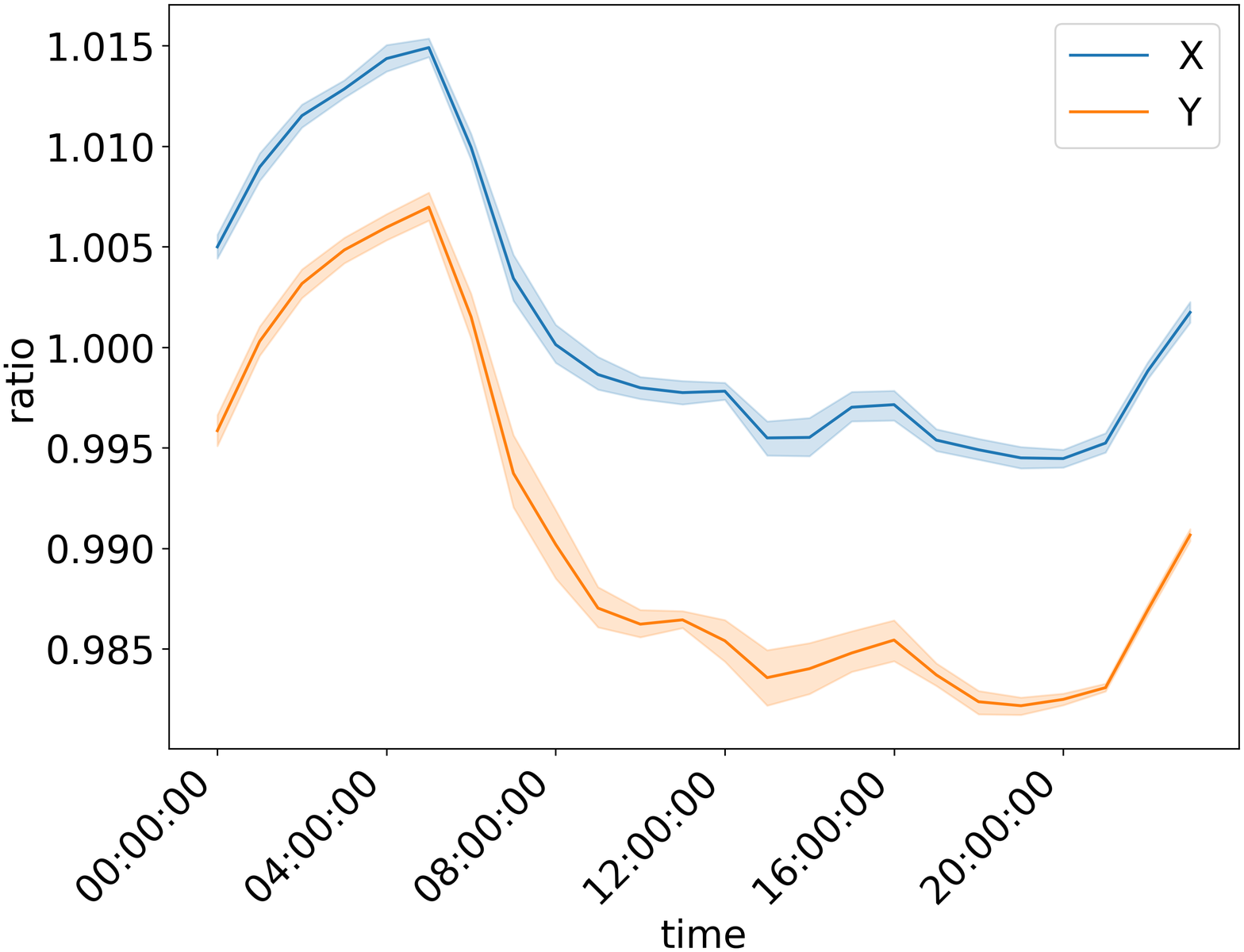}
    \end{minipage}

    \caption{
        Time-series observations from the selected sensor \texttt{n296}.
        The upper panel shows the raw pressure measurements spanning seven days.
        The lower panel shows the converted feature representation.
        Both plots aggregate the data by time interval, with the solid line indicating the mean across the seven days and the shaded area representing the standard deviation around that mean.
    }
    \label{fig:case-study-p866-observations}
\end{figure}

We provide supplementary materials for the experiments on the analysis of leaking water data in Section~\ref{sec:simulation_water_pipe_network}.
Figure~\ref{fig:water-leak-ground-truth} describes leakage events at different time points.
Durations vary: some leaks are repaired soon after they occur, while others persist for much longer.
Figure~\ref{fig:case-study-p866-observations} (top) shows the two 7-day subsequences at sensor \texttt{n296} (before vs.\ after the change point).
The bottom panel shows the corresponding aggregated-then-normalised features for $\mathbf{X}$ and $\mathbf{Y}$.
Both views clearly reveal differences between the two samples and their underlying observation sequences.

\section{Supplementary Materials for the Traffic Data Experiment in Section~\ref{sec:simulated_traffic_dataset_body}}
\label{sec:appendix-simulated_traffic_dataset}

\begin{figure*}[th]
    \begin{minipage}[t]{0.49\hsize}
      \centering
      \includegraphics[keepaspectratio, scale=0.23]{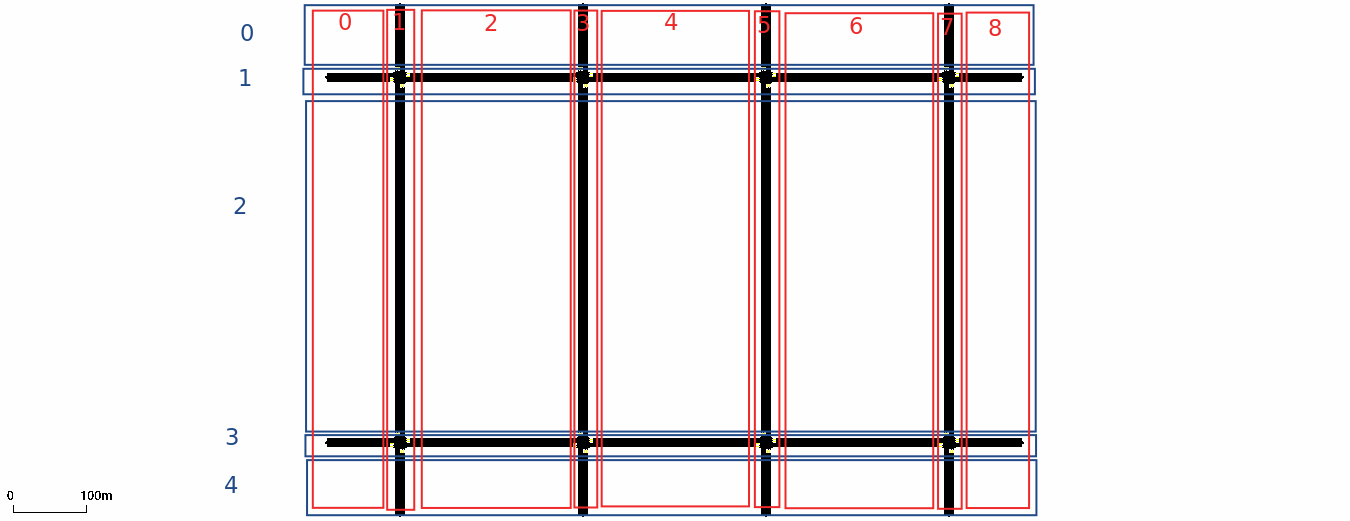}
      \subcaption{The grid road network.}
      \label{grid-network}
    \end{minipage}
    \hfill
    \begin{minipage}[t]{0.49\hsize}
      \centering
      \includegraphics[keepaspectratio, scale=0.2]{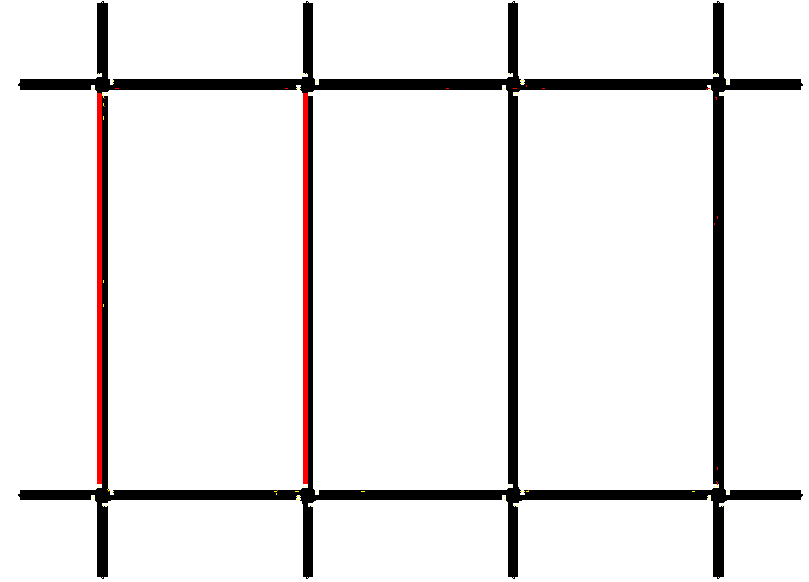}
      \subcaption{The two closed lanes in the grid network, highlighted in red.}
      \label{grid-network-block}
    \end{minipage}

    \vspace{1cm} 

    \begin{minipage}[t]{1.0\hsize}
      \centering
      \includegraphics[keepaspectratio, scale=0.25]{figures/appendix/sumo-assessments/zooming-intersection.png}
      \subcaption{An intersection in the grid network, with eight sensors (induction loop detectors) indicated by rectangles and vehicles by triangles.}
      \label{zooming-intersection}
    \end{minipage}

   \caption{
    Illustrations of the traffic road network used in the experiments of Section~\ref{sec:simulated_traffic_dataset_body}. 
    Figure~\ref{grid-network} describes the grid network. Each edge (= road) and each intersection are named according to the row index (blue) and the column index (red). 
    In each scenario, there are two groups of vehicles, one travelling from the left-upper edge (row 1, column 0) to the right-bottom edge (row 3, column 8) and the other travelling in the opposite direction (from the right-bottom to the left-upper). Figure~\ref{grid-network-block} highlights the two lanes blocked in scenario $Q$ in red. Figure~\ref{zooming-intersection} describes an intersection, where two sensors are located at the end of two lanes, resulting in eight sensors for each intersection. 
    }
   \label{fig:appendix-grid-network-iilustration}
\end{figure*}

\begin{figure*}[h]
    \centering
    \includegraphics[width=140mm]{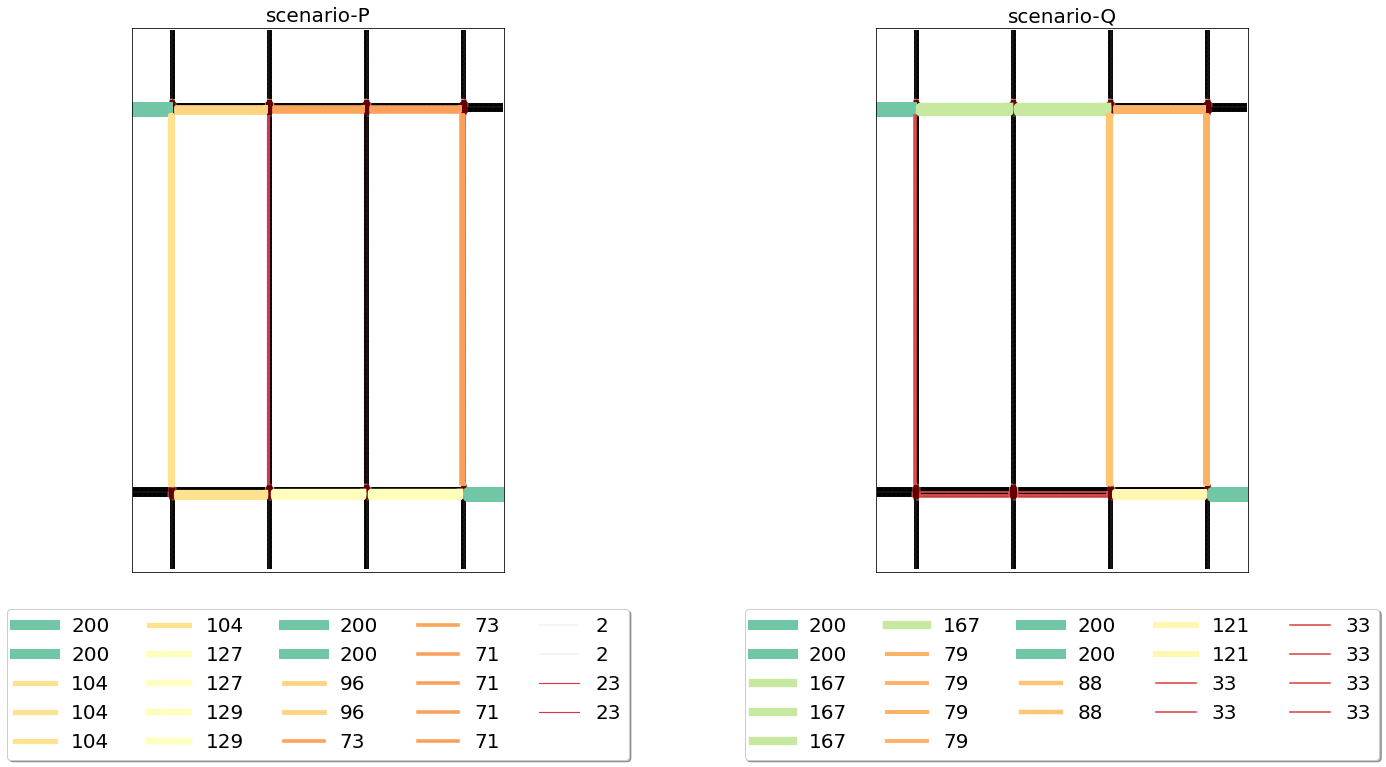}
    \caption{
   The number of vehicles passing each road in one simulation of the base scenario (left) and the perturbed one (right). Each edge's line width and colour indicate the corresponding number of vehicles shown in the legend box.
    }
    \label{fig:trajectory-sample}
\end{figure*}

Figure~\ref{fig:appendix-grid-network-iilustration} describes the grid network used in our traffic experiments. Figure~\ref{grid-network} labels roads and intersections by row (blue) and column (red); traffic flows in two opposing streams (left-upper to right-bottom and vice versa). Figure~\ref{grid-network-block} highlights the two lanes blocked in the perturbed scenario. Figure~\ref{zooming-intersection} zooms into an intersection, where two sensors per incoming lane yield eight sensors per intersection.
Figure~\ref{fig:trajectory-sample} compares a single simulation without road closures (left) to one with two blocked lanes (right); edge width and colour encode the number of vehicles per road (see legend).

\label{sec:appendix-sumo-generation-process}

\section{Kernel Length Scale Selection}
\label{sec:length-scale-selection}

\subsection{Variable-wise Median Heuristic}
\label{sec:dimension-wise-median}

We describe the variable-wise median heuristic, a method for selecting the length scales $\gamma_1, \dots, \gamma_D$ in the ARD kernel~\eqref{eq:ard-kernel-matrix}. This method extends the standard median heuristic used in the kernel literature \cite[e.g.,][]{garreau2018large}.

For each $d = 1, \dots, D$, we set $\gamma_d$ as the median of pairwise distances of the $d$-th variable values in the datasets. To describe the heuristic more precisely, let ${\bf X}  = \{X_1, \dots, X_n \} \subset \mathbb{R}^D$ and ${\bf Y} = \{ Y_1, \dots, Y_m \} \subset \mathbb{R}^D$. For each $i=1,\dots, n$, write $X_i = (x^i_1, \dots, x^i_D )^\top \in \mathbb{R}^D$, i.e., $x_d^i \in \mathbb{R}$ is the $d$-th entry of the vector $X_i$. Similarly, for $j = 1,\dots, m$, let $Y_j = (y_1^j, \dots, y_D^j)^\top \in \mathbb{R}^D$ with $y_d^j$ denoting the $d$-th entry of $Y_j$. 
Then, for each $d = 1,\dots, D$, define $L_d \subset \mathbb{R}$ as the set of the $d$-th variable values from ${\bf X}$ and ${\bf Y}$, i.e., 

$$
L_{d} := \{ x_{d}^i \mid  i = 1,\dots, n \} \cup \{ y_{d}^j \mid j = 1,\dots, m \},
$$

We then set $\gamma_d$ as the median of pairwise distances between elements in $L_d$:

\begin{equation} \label{eq:median-heuristic}
\gamma_{d}^2 := {\rm median}\left\{  (z - z')^2 \mid z, z' \in L_{d}, \quad z \not= z'     \right\}.
\end{equation}

By setting the length scale $\gamma_d$ in this way, the hope is that the scaled differences  

\begin{equation} \label{eq:ratio-dist-length}
\{ (x_{d}^i - y_{d}^j)^2 / \gamma_{d}^2 \mid i = 1,\dots, n,\ j = 1,\dots, m  \}    
\end{equation}
would have approximately a similar level of variability across different dimensions $d = 1,\dots, D$.

\subsubsection*{Minimum Value Replacement}

If the majority of the elements in $L_d$ take the same value (e.g., $x_d^i = y_d^j = 0$ for many of $i = 1, \dots, n$ and $j = 1, \dots, m$), many of the pairwise differences in \eqref{eq:median-heuristic} become zero, so their median $\gamma_d$ may be zero. This causes a problem because the length scale should be positive by definition. We propose the following procedure to address this issue. First, compute the $D$ length scales $\gamma_1, \dots, \gamma_D$ using the variable-wise median heuristic, and let $\gamma_{\rm min} > 0$ be the minimum among the {\em positive} computed length scales. Then, if the median~\eqref{eq:median-heuristic} becomes zero for the $d$-th variable, we set $\gamma_d = \gamma_{\rm min}$.

\subsection{Variable-wise Mean Heuristic}

In the application to traffic simulation data in Section~\ref{sec:simulated_traffic_dataset_body}, we found that the variable-wise median heuristic makes many length scales $\gamma_1, \dots, \gamma_D$ zero. This is caused by the data vectors ${\bf X} = \{ X_1, \dots, X_n \}$ and ${\bf Y} = \{  Y_1, \dots, Y_m \}$ being sparse in this application, i.e., each data vector has many zero entries. Accordingly, as explained above, the zero length scales are replaced by the minimum of the positive length scales. However, we found that optimisation of the ARD weights using the resulting length scales becomes unstable. The reason is that the replaced length scale, say $\gamma_d$, may be much smaller than some of the pairwise distances $(x^i_d - y_d^j)$ in the data. 

For example, suppose for simplicity that $n = m = 3$, $x_d^1 = x_d^2 = 0$, $x_d^3 = 2.0$, and $y_d^1 = y_d^2 = y_d^3 = 0$. Then the set of pairwise squared distances in \eqref{eq:median-heuristic} is $\{ 0, 0, 0, 0, 0, 0, 0, 0, 0, 0,  4, 4, 4, 4, 4 \}$. Then the median \eqref{eq:median-heuristic} is zero, and thus $\gamma_d$ is set as the minimum $\gamma_{
\rm min}$ of the positive length scales. If this minimum is, for example, $\gamma_{\rm min} = 0.1$, then the ratio \eqref{eq:ratio-dist-length} for $i = 3$ and $j = 1$ becomes $(x_d^3 - y_d^1)^2 / \gamma_{\rm min}^2 = 4/ 0.1^2 = 400$. Since this value appears in the exponent of the ARD kernel  \eqref{eq:ard-kernel-matrix}, the kernel value becomes close to $0$ for the initial ARD values $a_1 = \cdots = a_D = 1$, and thus the optimisation of the ARD weights becomes unstable.  

Therefore, for such sparse data, we suggest using the variable-wise {\em mean} heuristic, i.e., setting $\gamma_d$ by replacing the median \eqref{eq:median-heuristic} by the mean of the pairwise distances. In the above example, by taking the mean of $\{ 0, 0, 0, 0, 0, 0, 0, 0, 0, 0,  4, 4, 4, 4, 4 \}$, we have $\gamma_d^2 = 1.33$, so $(x_d^3 - y_d^1)^2 / \gamma_d^2 = 4/ 1.33 = 3.00$; hence the resulting ARD kernel would not collapse to zero and the optimisation of the ARD weights becomes stable. Therefore, we use this variable-wise mean heuristic in the traffic simulation experiment in Section \ref{sec:appendix-simulated_traffic_dataset}.

\section{Variable Selection using the Optimised Weights}
\label{app:variable_selection}

We explain how to select variables using the optimised ARD weights $a^* = (a^*_1, \dots, a^*_D)$,  the solution of \eqref{eq:regularized-opt}. One way is to set a threshold $\pi_{\rm thr} \geq 0$ and select variables whose weights are above the threshold: $\hat{S} := \{ d \in \{1, \dots, D\} \mid  a^*_d > \pi_{\rm thr} \}$. The question is how to set the threshold. Our preliminary experiments revealed that the use of a fixed threshold, such as $\pi_{\rm thr} = 0$ and $\pi_{\rm thr} = 0.1$, does not work well. The reason is that the range of the ARD weights changes drastically depending on the given dataset. For example, the maximum and minimum of the ARD weights can be $10^{-3}$ and $10^{-7}$, respectively, for one dataset, while they can be $10^2$ and $10^{-1}$ for another dataset.  In either case, however, the ARD weights distribution indicates each variable's relative importance.

Consequently, we use the following data-driven method for determining the threshold $\pi_{\rm thr}$ based on the histogram of optimised ARD weights. The idea is to set the threshold as the smallest local minimum in the histogram. For instance, if $D = 5$, $a^*_1 = 0.01$, $a^*_2 = 0.02$, $a^*_3 = 0.02$, $a^*_4 = 0.03$, $a^*_5 = 0.1$,  we set $\pi^*_{\rm thr} = 0.03$, and the 5th variable is selected as $a^*_5 = 0.1 > 0.03$.  
Algorithmically, the method first constructs a histogram of the optimised ARD weights (with 100 bins as a default setting). Subsequently, it identifies bins with zero frequency. Among these bins, the one with the smallest value is selected as the threshold $\pi_{\rm thr}$.

\section{Selecting Candidate Regularisation Parameters}
\label{sec:reg-param-range-selection}

\begin{algorithm}[tb]
  \caption{Generating Candidate Regularisation Parameters}
  \label{alg-hyperparameter_range_selection}
  \textbf{Input:} 
  $\lambda_{\rm lower}$: the lower bound (default value: 0.01). $N_\lambda$: the number of candidate parameters (default value: 6). $({\bf X}, {\bf Y})$: a pair of data. $N_{\rm history}$: the number required for the stopping condition (default value: 3). 

  \algohline
  \textbf{Output:} $\Lambda$: a set of candidate regularisation parameters.
  \algohline 
  \begin{algorithmic}[1]
    \State $\lambda = \lambda_{\rm lower}$
    \State $S_{\rm history} \gets []$ \Comment{Initialize an empty list}
    \State While { \textsc{StopCriteria}($\hat{S}_{\lambda}, S_{\rm history}, N_{\rm history}$) == False}; do
            \State \ \ \ \ Obtain $a_{\lambda} \in \mathbb{R}^D$ by solving \eqref{eq:regularized-opt} using $\lambda$ and $({\bf X}, {\bf Y})$.
        \State \ \ \ \ Obtain $\hat{S}_{\lambda}$ using $a_{\lambda}$ as in Section~\ref{sec:variable_selection}.
        \State \ \ \ \ \Call{AppendToList}{$S_{\text{history}}, \hat{S}_{\lambda}$}
        \State \ \ \ \ Update($\lambda$).
    \State End While
    \State $\lambda_{\rm upper} = \lambda$.
    \State ${\rm step}_{\lambda} = (\lambda_{\rm upper} - \lambda_{\rm lower}) / (N_{\lambda} - 1)$.
    \State $\Lambda = \{\lambda_{\rm lower}, \lambda_{\rm lower} + {\rm step}_{\lambda}, \lambda_{\rm lower} + 2 {\rm step}_{\lambda} \dots, $ 
    \par \ \ \ \ $\lambda_{\rm lower} + (N_{\lambda} - 1){\rm step}_{\lambda}, \lambda_{\rm upper}\}$.
\end{algorithmic}
\algohline
  \begin{algorithmic}[1]
  \State Function StopCriteria$(\hat{S}_{\lambda}, S_{\rm history}, N_{\rm history})$\{
  \If{$|\hat{S}_{\lambda}| == 1$}
    \State Return \textbf{True}
  \ElsIf{$S_{\text{history}}[-1] == ,\dots, == S_{\text{history}}[-N_{\rm history}]$}
    \State Return \textbf{True}
  \Else
    \State Return \textbf{False}
  \EndIf
  \State \}
  \end{algorithmic}

  \algohline

  \begin{algorithmic}[1]
  \State Function Update($\lambda_{\rm previous}$) \{
    \If{$0 < \lambda_{\rm previous} < 1.0$}
      \State Return $2 \lambda_{\rm previous}$
    \Else
      \State Return $\lambda_{\rm previous} + 0.5$
    \EndIf
  \State \}
  \end{algorithmic}
\end{algorithm}

For the methods for selecting the regularisation parameter $\lambda$ described below, one needs to specify the possible range of $\lambda$ and set candidates from this range. 

Let $\lambda_{\rm lower}$ be the lower bound of the range (we set $\lambda_{\rm lower} = 0.01$ as a default setting). 
Let $\hat{S}_{\lambda}$ denote the set of selected variables using the regularisation parameter $\lambda$. 
We set the upper bound $\lambda_{\rm upper}$ as the smallest $\lambda$ that makes the number of selected variables to be one, i.e., $| \hat{S}_\lambda | = 1$, or such that any value $\tilde{\lambda}$ larger than $\lambda$ does not change the selected variables, i.e., $\hat{S}_\lambda = \hat{S}_{\tilde{\lambda}}$ for any $\tilde{\lambda} > \lambda$.

Once the upper bound $\lambda_{\rm upper}$ is found, we set candidate parameters as follows. Let $N_\lambda$ be the number of candidate parameters, and let ${\rm step}_{\lambda} = (\lambda_{\rm upper} - \lambda_{\rm lower}) / (N_{\lambda} - 1)$. Then we define candidate parameters as $\{\lambda_{\rm lower}, \lambda_{\rm lower} + {\rm step}_{\lambda}, \dots, \lambda_{\rm lower} + {\rm step}_{\lambda} (N_{\lambda} - 1), \lambda_{\rm upper}\}$.

Algorithm \ref{alg-hyperparameter_range_selection} describes the procedure.
In the algorithm, $S_{\rm history}$ is initialised as an empty list. 
The while-iteration from lines 3 to 7 continues as long as the function \textsc{StopCriteria} returns False. \textsc{StopCriteria} checks two conditions: 1) if the number of selected variables $\hat{S}_\lambda$ is one, i.e., $| \hat{S}_\lambda | = 1$; and 2) if the selected variables do no change for the previous $N_{\rm history}$ iterations. In the latter (Line 4 of \textsc{StopCriteria}), $S_{\rm history}[-1]$ refers to the set of variables in the last iteration, and $S_{\rm history}[-N_{\rm history}]$ is that in the $N_{\rm history}$-th previous iteration.

In the main algorithm, line 4 optimises the ARD weights $a_\lambda$ by numerically solving \eqref{eq:regularized-opt} using the current value of the regularisation parameter $\lambda$ and data $({\bf X}, {\bf Y})$. Based on these ARD weights, line 5 selects variables ${\hat S}_\lambda$ as explained in Section \ref{sec:variable_selection}. In line 6, the function  \textsc{AppendToList} appends ${\hat S}_\lambda$ to the list $S_{\rm history}$, which is used in  \textsc{StopCriteria}. In line 7, the function \textsc{Update} increases the regularisation parameter $\lambda$ for the next iteration. 
If $0 < \lambda < 1$, the function \textsc{Update} increases $\lambda$ by multiplying $2$; if $1 < \lambda$, it adds $0.5$ to $\lambda$. We have designed this procedure to speed up the search, as we observed that the upper bound is often above $1$ in our preliminary analysis.

Once the upper bound $\lambda_{\rm upper}$ is determined by \textsc{StopCriteria}, the algorithm outputs a set of $N_{\lambda}$ candidate parameters in Lines 10 and 11.

\section{Effects of the Regularisation Parameter}
\label{sec:appendix-syn2-result-parameter-choice}

We study how the choice of the regularisation parameter $\lambda$ in \eqref{eq:regularized-opt} affects the variable selection performance. To this end, we set $\lambda$ as each of the candidate values from $10^{-3}, 10^{-3 + 0.25}, 10^{-3 + 0.5}, \dots, 10^{1.25}, 10^1$, optimise the ARD weights by numerically solving \eqref{eq:regularized-opt} (see Section~\ref{sec:assessment-opt-configuration} for details), and perform variable selection as in Appendix~\ref{app:variable_selection}.  Figure~\ref{fig:synthetic2-lambda-comparisons-precision-recall}  describes the F score, Precision and Recall for each value of $\lambda$ and each setting, where the means and standard deviations are obtained from the results of 10 independently repeated experiments.  

One can observe that the optimal value of $\lambda$ varies depending on the setting of distributions $P$ and $Q$. For example, the highest precision is attained with $\lambda = 10^{-1.5}$ for the ``Narrower variances'' setting, but it is attained with $\lambda = 1.0$ for the other settings. The dependence of the best regularisation parameter on the setting of $P$ and $Q$ highlights the difficulty of manually selecting an appropriate regularisation parameter. This fact motivates the proposed approaches of Algorithms~\ref{alg-hyperparameter_selection}~and~\ref{alg-enhanced-stability-selection}. 

\begin{figure*}[t]
  \begin{minipage}[b]{1.0\hsize}
      \centering
      \includegraphics[scale=0.3]{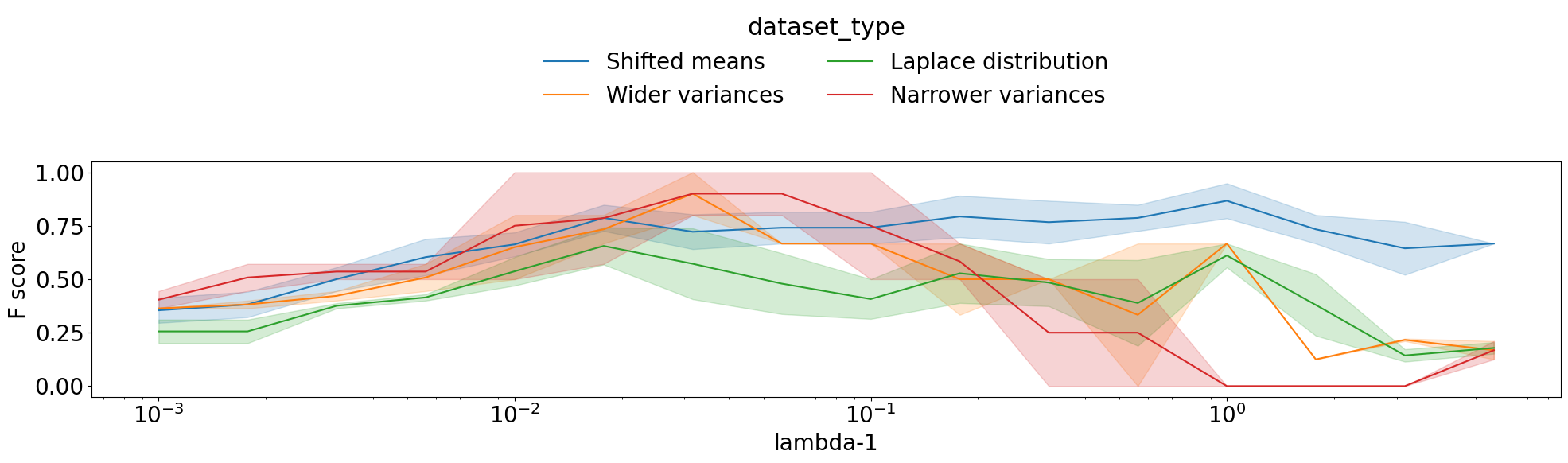}
      \label{syn2-f-score-comparison}
  \end{minipage}

  \begin{minipage}[b]{1.0\hsize}
      \centering
      \includegraphics[scale=0.3]{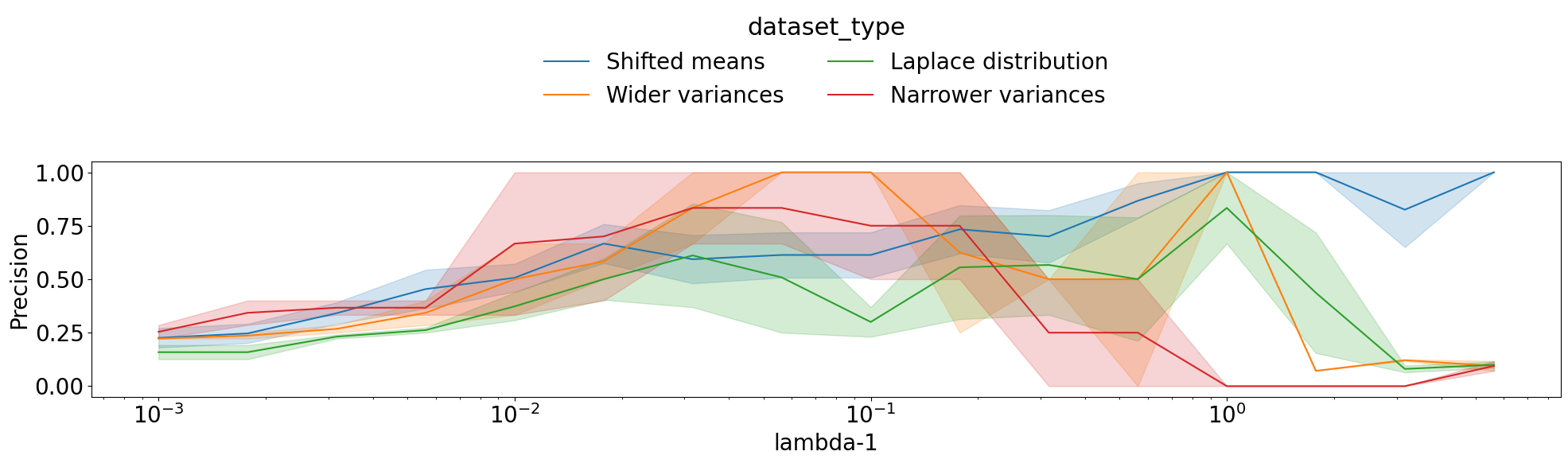}
  \end{minipage}

  \begin{minipage}[b]{1.0\hsize}
      \centering
      \includegraphics[scale=0.3]{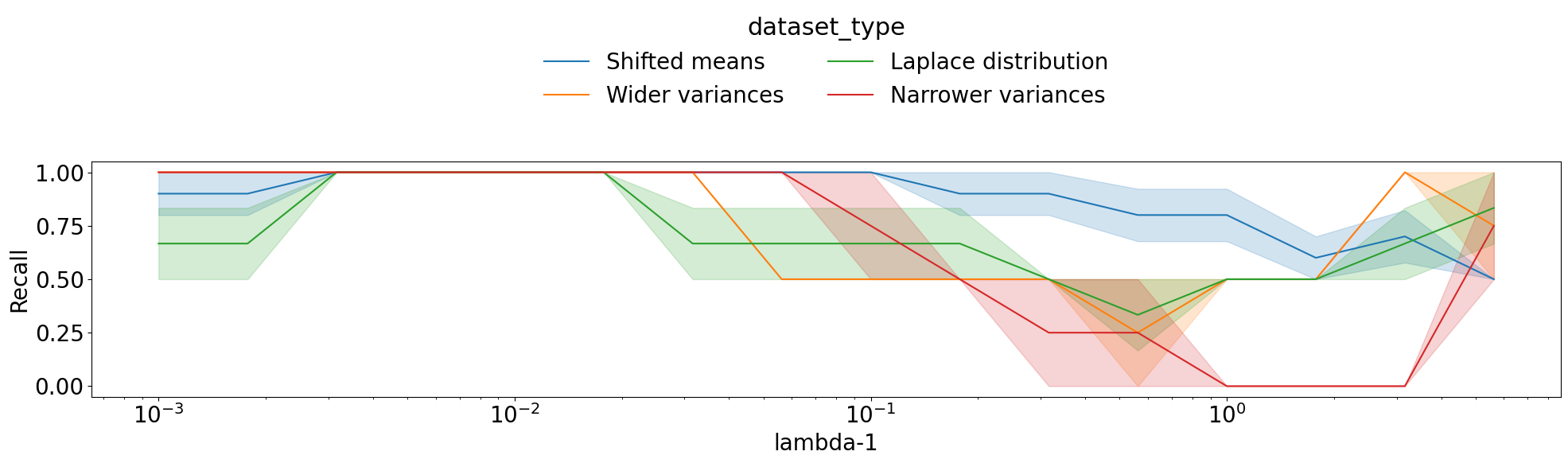}
  \end{minipage}
  \caption{
Effects of the regularisation parameter on the variable selection performance, discussed in Appendix~\ref{sec:appendix-syn2-result-parameter-choice}. The top, middle and bottom figures show the F score, Precision and Recall for each setting of distributions $P$ and $Q$, where the horizontal axis indicates the regularisation parameter $\lambda$ used.
}
  \label{fig:synthetic2-lambda-comparisons-precision-recall}
\end{figure*}

\section{Comparing Different CV Aggregation Strategies}
\label{sec:appendix-ss-analysis}

In the process of developing Algorithm~\ref{alg-enhanced-stability-selection} ({\tt CV-aggregation}), there were several candidate ways of computing the aggregation score vector $\hat{\Pi}_\lambda = (\hat{\Pi}_{\lambda, 1}, \dots, \hat{\Pi}_{\lambda, D} )^\top \in \mathbb{R}^D$ for each candidate regularisation parameter $\lambda$. Here, we compare these different choices and how we have arrived at our choice for Algorithm~\ref{alg-enhanced-stability-selection}. 
We use the notation for Algorithm~\ref{alg-enhanced-stability-selection} in Section~\ref{sec:stability-selection} in the following. 

First, there are the following five ways of defining $\hat{\Pi}_\lambda$ based on the selected variables $S_\lambda^i \subset \{1, \dots, D
\}$, where $i = 1, \dots, K$:
\begin{enumerate}
    \item ``{\tt plane-variable}'': $\hat{\Pi}_{\lambda, d} = \frac{1}{K} \sum_{i=1}^{K} I\{ d \in \hat{S}_{\lambda}^i \}$ for ${d = 1,\dots, D}$.    
    \item ``{\tt p-value-variable}'': ${\hat{\Pi}_{\lambda, d} = \frac{1}{K} \sum_{i=1}^{K} (1 - p_\lambda^i)\ I\{ d \in \hat{S}_{\lambda}^i \}}$ for ${d = 1,\dots, D}$.
    \item ``{\tt test-power-variable}'':  ${ \hat{\Pi}_{\lambda, d} = \frac{1}{K} \sum_{i=1}^{K} \ell_{\rm val} (a_\lambda^i )\ I \{ d \in \hat{S}_{\lambda}^i \} }$ for ${d = 1,\dots, D}$.
    \item ``{\tt p-value-filter-variable}'': ${\hat{\Pi}_{\lambda, d} = \frac{1}{K} \sum_{i=1}^{K} I(p_{\lambda}^i < 0.05) I \{ d \in \hat{S}_i^{\lambda} \}}$ for ${d = 1,\dots, D}$. 
    \item ``{\tt p-value-filter-test-power-variable}'': ${\hat{\Pi}_{\lambda, d} = \frac{1}{K} \sum_{i=1}^{K} I( p_\lambda^i < 0.05 ) \ell_{\rm val} (a_\lambda^i )  I \{ d \in \hat{S}_i^{\lambda} \}}$ for ${d = 1,\dots, D}$. 
\end{enumerate}
We also have the following five ways based on the normalised ARD weights $\tilde{a}_\lambda^i$, where $i = 1, \dots, K$:
\begin{enumerate}
    \item ``{\tt plane-ard}'': 
    $\hat{\Pi}_{\lambda} = \frac{1}{K} \sum_{i=1}^{K} \tilde{a}_\lambda^i$.
    \item ``{\tt p-value-ard}'': 
    $\hat{\Pi}_\lambda = \frac{1}{K} \sum_{i=1}^{K} (1 - p^i_{\lambda}) \tilde{a}_\lambda^i$.
    \item ``{\tt test-power-ard}'': 
    $\hat{\Pi}_\lambda = \frac{1}{K} \sum_{i=1}^{K} \ell_{\rm val} (a_\lambda^i ) \tilde{a}_\lambda^i$.
    \item ``{\tt p-value-filter-ard}'': 
    ${\hat{\Pi}_\lambda~:=~\frac{1}{K} \sum_{i=1}^{K}  I( p_\lambda^i < 0.05 ) \tilde{a}_\lambda^i}$.   
    \item ``{\tt p-value-filter-test-power-ard}'': 
      $\hat{ \Pi }_\lambda~=~\frac{1}{K} \sum_{i=1}^K I( p_\lambda^i < 0.05 ) \ell_{\rm val} (a_\lambda^i ) \tilde{a}_\lambda^i$.
\end{enumerate}
Note that the last approach is the one used in Algorithm \ref{alg-enhanced-stability-selection}.

\begin{figure*}[htbp]
  \centering
  \includegraphics[scale=0.2]{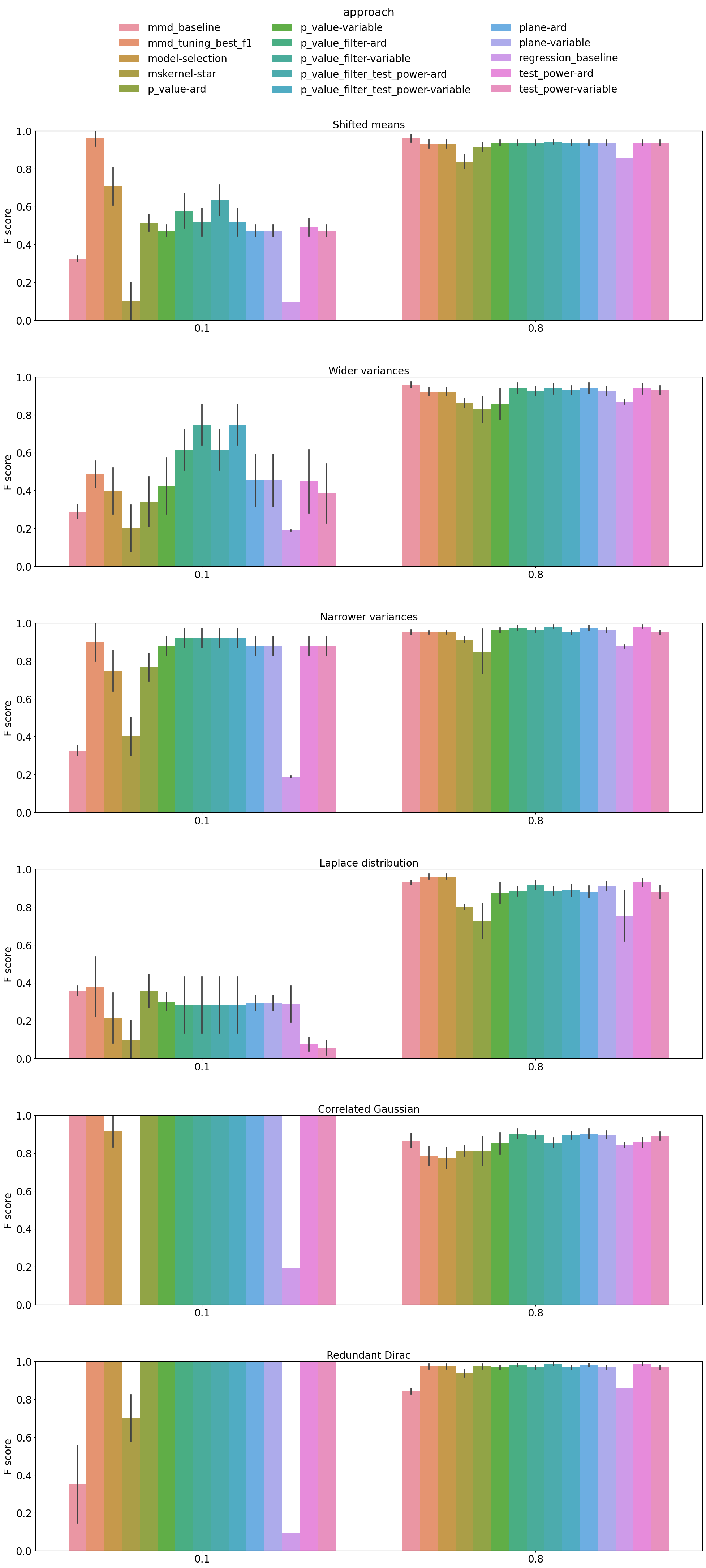}
  \caption{
  Comparison of the ten candidate aggregation methods for Algorithm~\ref{alg-enhanced-stability-selection} and the other baseline approaches in Appendix~\ref{sec:appendix-ss-analysis}.  Note that "{\tt p-value-filter-test-power-ard}'' is the one adopted in Algorithm~\ref{alg-enhanced-stability-selection} in the main body of the paper.
  }
  \label{fig:appendix-subsampling-comparison}
\end{figure*}

Figure~\ref{fig:appendix-subsampling-comparison} describes the results of these candidate ways for computing the score vector $\hat{\Pi}_\lambda$ in Algorithm~\ref{alg-enhanced-stability-selection} and the other baseline methods. It shows F scores obtained in the same way as in Section~\ref{sec:synthetic_data_assessment}, for two settings of the rate of the ground-truth variables $S$,  $\rho = 0.1$ and $\rho = 0.8$.

One can observe that ``{\tt  p-value-filter-test-power-ard}'' exhibits the highest stability among the ten candidate methods for Algorithm~\ref{alg-enhanced-stability-selection}. Consequently, we have decided to adopt this way for Algorithm~\ref{alg-enhanced-stability-selection}.

\section{Supplementary to Section~{\ref{sec:empirical_assessment}}}
\label{sec:appendix-syn2-result}

\subsection{Experiments with with Higher $\rho$}

\label{sec:exp-with-higher-low}

\begin{figure}[t]
    \centering
    \includegraphics[scale=0.3]{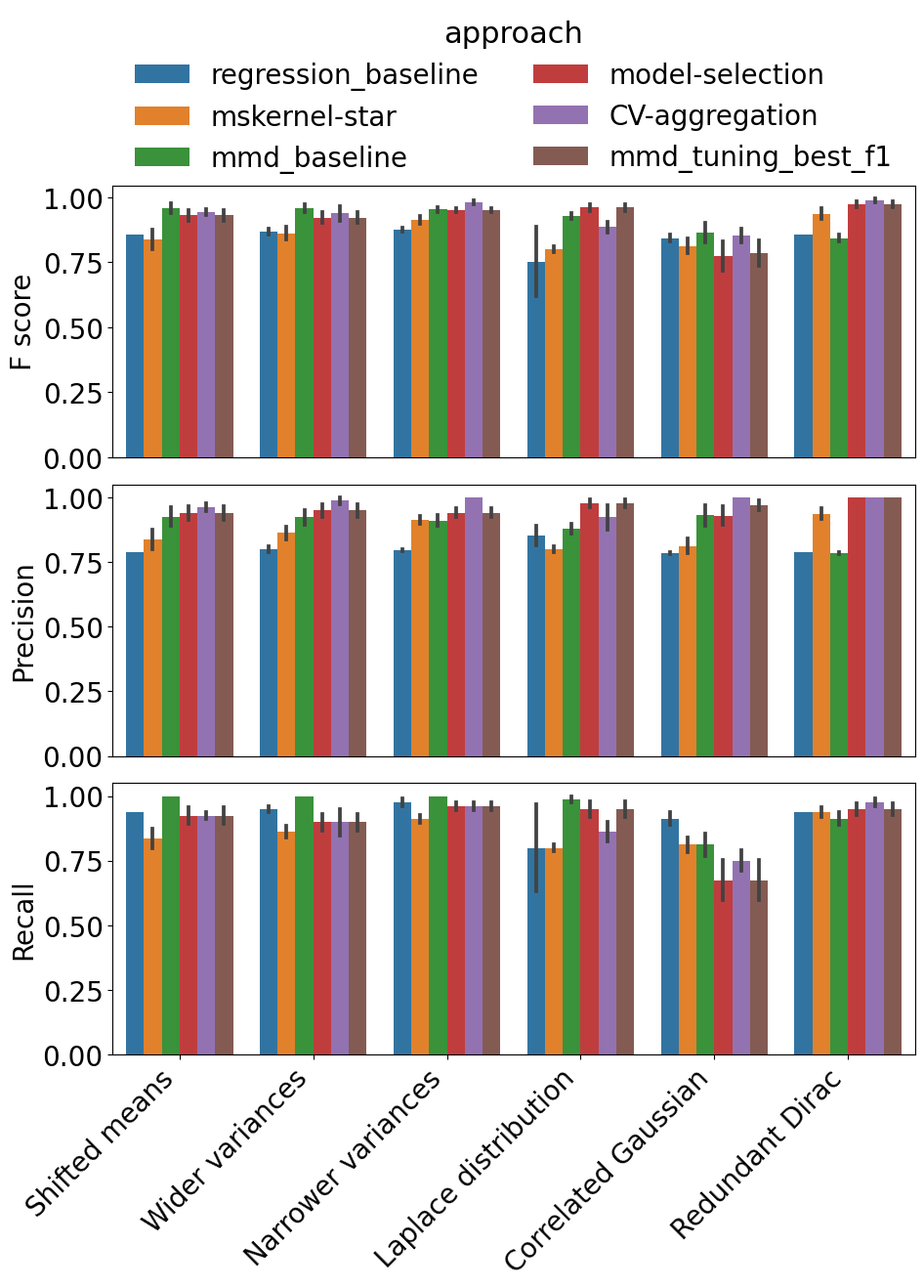}
    \caption{
    The results of the same experiments as Section~\ref{sec:synthetic_data_assessment} with $\rho = 0.8$,  discussed in Appendix \ref{sec:appendix-syn2-result}.
    }
    \label{fig:synthetic2-f-precision-recall-0.8}
\end{figure}

We perform the same experiments as in Section~\ref{sec:empirical_assessment}, with the rate $\rho$ of ground-truth variables $S$ being modified to $0.8$, so that $|S| = 20 \times 0.8 = 16$. Figure~\ref{fig:synthetic2-f-precision-recall-0.8} shows the results. Compared with the results for $\rho = 0.1$ in Section~\ref{sec:empirical_assessment}, all the approaches yield reasonably high scores across different settings of distributions $P$ and $Q$. The relative ease of the setting $\rho = 0.8$ can be attributed to the high number of ground-truth variables $16$. For example, if one selects all the variables, i.e., $\hat{S} = \{1, \dots, D\}$, then the Recall is $1$ and the Precision is $0.8$, so the F score is 0.89. However, in practice, one cannot take such a strategy because the value of $\rho$ is typically unknown. Given that we do not use the information of true $\rho$ for variable selection, the obtained scores can be regarded as reasonably high. Among the approaches considered, the proposed methods, {\tt model-selection} and {\tt CV-aggregation}, yield stably high scores across the different distribution settings.

\subsection{Assessing the Effects of Dimensionality}
\label{sec:various_feature_size_assessment}

We assess the effects of the dimensionality $D$, i.e., the total number of variables, on the variable selection performance.
We consider the same setting as Section~\ref{sec:empirical_assessment}, with $P$ and $Q$ being $D$-dimensional distributions in the following way. 

Let $P = \mathcal{N}({\bf 0}_D, \Sigma)$, where the covariance matrix $\Sigma \in \mathbb{R}^{D \times D}$ is diagonal with $D$ diagonal elements being 
$$
(\underbrace{1, 1, \dots, 1}_{10}, \underbrace{2, 2, \dots, 2}_{10}, \dots, \underbrace{D/10, D/10, \dots, D/10}_{10})
$$  
We define the other distribution $Q$ by defining how its random vector $Y \sim Q$ is generated. Let $S \subset \{1, 2, \dots, D \}$ be ground-truth variables with $\rho = 0.1$, so that $|S| = D \times 0.1$.
Let $\tilde{X}_S \sim \mathcal{N}({\bf 0}_{ |S| }, \Sigma_S)$, where $\Sigma_S \in \mathbb{R}^{ |S| \times |S| }$ is the submatrix of $\Sigma$ restricted to the variables $S$. Let $Z_S \sim \mathcal{N}( {\bf 0}_{|S|}, I_{ |S| })$. Then we define $Y_S = \tilde{X}_S \odot Z_S$, where $\odot$ denotes the element-wise product. Let $Y_{ \{ 1, \dots, D \} \backslash S } \sim \mathcal{N}( 0_{ D - |S| },  \Sigma_{ \{1, \dots, D\} \backslash S } )$. Finally, define $ Y = (Y_S, Y_{ \{ 1, \dots, D \} \backslash S }) \in \mathbb{R}^D$, and $Q$ is defined as the distribution of $Y$. 

We consider various values for the dimensionality: $D = 20, 50, 100, 200, 300, 400, 500$. We fix the sample size $N = 200$. For each value of $D$ and each method, we experiment as in Section~\ref{sec:empirical_assessment} and repeat it 10 times to derive evaluation scores' means and standard deviations.

\begin{figure}[t]
    \centering
    \includegraphics[scale=0.38]{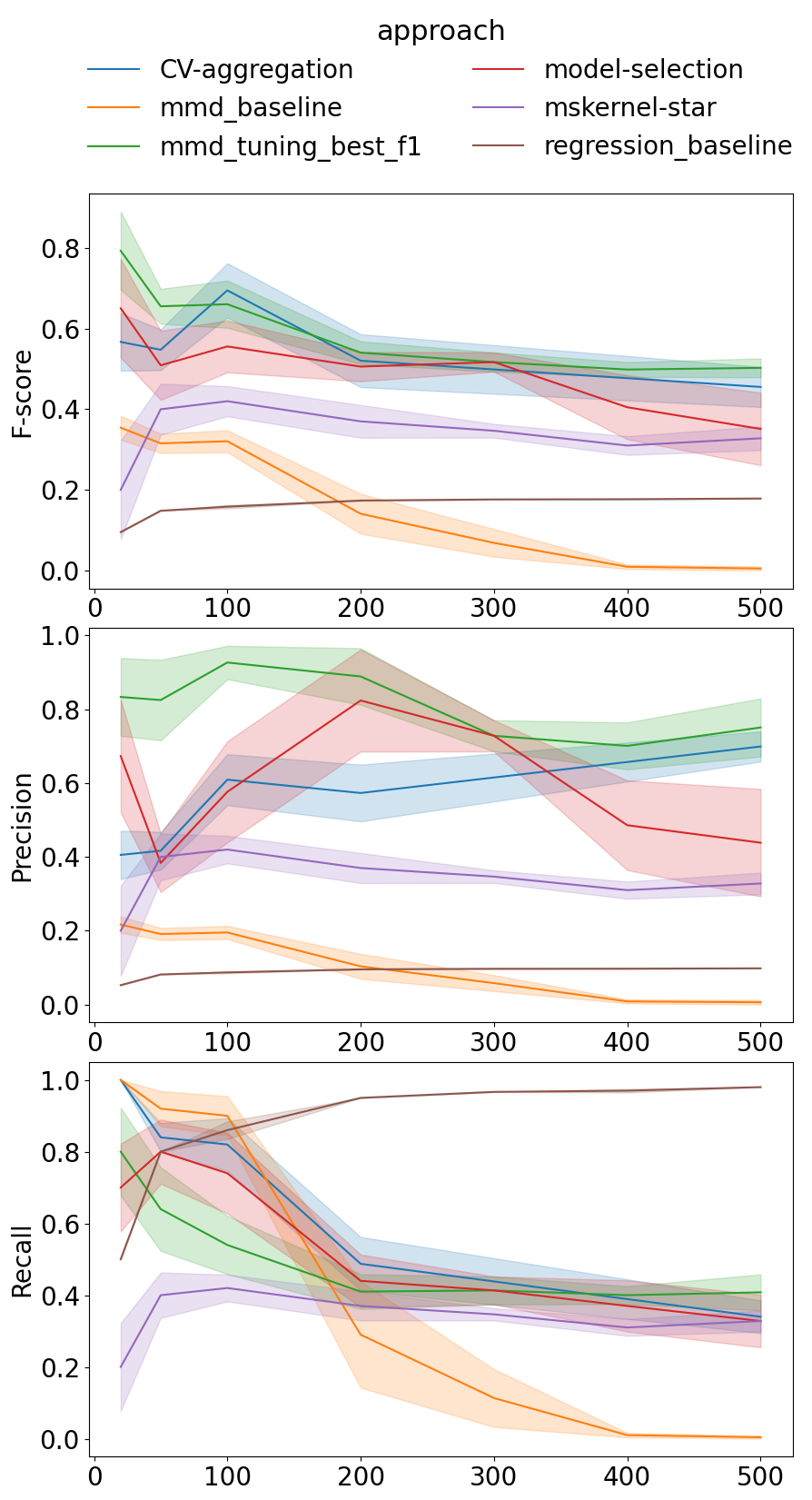}
    \caption{
The results of the experiments on assessing the effects of the dimensionality $D$ in Appendix~\ref{sec:various_feature_size_assessment}. The top, middle and bottom figures show the F score, Precision and Recall for each method and each $D$, with the mean and standard deviation computed from 10 independently repeated experiments. The horizontal axis indicates the dimensionality $D$.
}
    \label{fig:feature_size_comparison}
\end{figure}

Figure~\ref{fig:feature_size_comparison}  shows the results. As a general trend, all the approaches, except the regression baseline, experience a drop in their F scores as the dimensionality $D$ increases. This drop can be attributed to the variable selection task becoming harder for higher $D$. The decrease of the F score appears to be milder for {\tt CV-aggregation} than the other methods, and the {\tt CV-aggregation} attains similar F scores for the largest $D$ as {\tt mmd-tuning-best-f1}, which is not implementable in practice and provides the best possible performance of {\tt model-selection}.

When comparing {\tt mmd-baseline} with the proposed methods ({\tt CV-aggregation} and {\tt model-selection}, we observe that {\tt mmd-baseline} experiences a drastic drop in the F score as the dimensionality $D$ increases, eventually resulting in the zero F score for $D = 500$. In contrast, the proposed methods yield reasonable F scores even for $D = 500$. This difference can be attributed to the regularisation approach used in the proposed methods, which proves to be more effective for higher dimensionality $D$. 

\section{Variable Detection with Linear MMD Estimator} 
\label{sec:linear-estimator-app}

A linear unbiased MMD estimator, proposed by Gretton et al.~\cite{JMLR:v13:gretton12a}, is a computationally efficient alternative to the quadratic unbiased MMD estimator~\eqref{eq:mmd-unbiased-est}. The quadratic estimator's computational complexity is $O(n^2)$ for sample size $n = m$, which can be costly for large $n$. The linear MMD estimator's complexity is $O(n)$, thus offering a reduced computational cost, while its variance is larger than the quadratic estimator. Here, we compare the linear and quadratic estimators when used in the proposed methods for variable selection.

Given ${\bf X} = \{X_1, \dots, X_n \} \subset \mathbb{R}^D$ and ${\bf Y} = \{Y_1, \dots, Y_n \} \subset \mathbb{R}^D$   with sample size $n$ being odd,  the linear MMD estimator is defined as 
\begin{equation*}
    {\rm MMD}^2_l({\bf X}, {\bf Y}) := \frac{2}{n} \sum_{i=1}^{n/2} h( (X_{2i-1}, Y_{2i-1}), (X_{2i}, Y_{2i})) 
\end{equation*}
where $h( (X, Y), (X', Y')) :=  k( X, X' ) + k(Y, Y') - k(X, Y')  - k(X', Y)$.
Assuming that $n$ is divisible by $4$,  an unbiased estimator of its variance, which corresponds to \eqref{eq:var-unbiased-est} for the quadratic estimator, is given by \cite[Eq.~(7)]{gretton2012optimal}:
\begin{align*}
    \check{\sigma}^2 := \frac{4}{n} \sum_{\ell = 1}^{n/4}   
& \left[  h( (X_{4\ell-3}, Y_{4\ell-2}), (X_{4\ell-3}, Y_{4\ell-2})) \right. \\ 
& \left. - h( (X_{4\ell-1}, Y_{4\ell}), (X_{4\ell-1}, Y_{4\ell}) ) \right]^2 
\end{align*}
Then, the ratio objective corresponding to \eqref{eq:430} is defined as 
$$
\frac{ {\rm MMD}^2_l({\bf X}, {\bf Y}) }{ \sqrt{ \check{\sigma}^2 + C } }
$$
Using this ratio objective in the regularised objective in \eqref{eq:regularized-opt}, one can use the linear estimator in the proposed methods.

\begin{figure*}[t]
    \begin{minipage}[t]{0.49\hsize}
      \centering
      \includegraphics[keepaspectratio, scale=0.32]{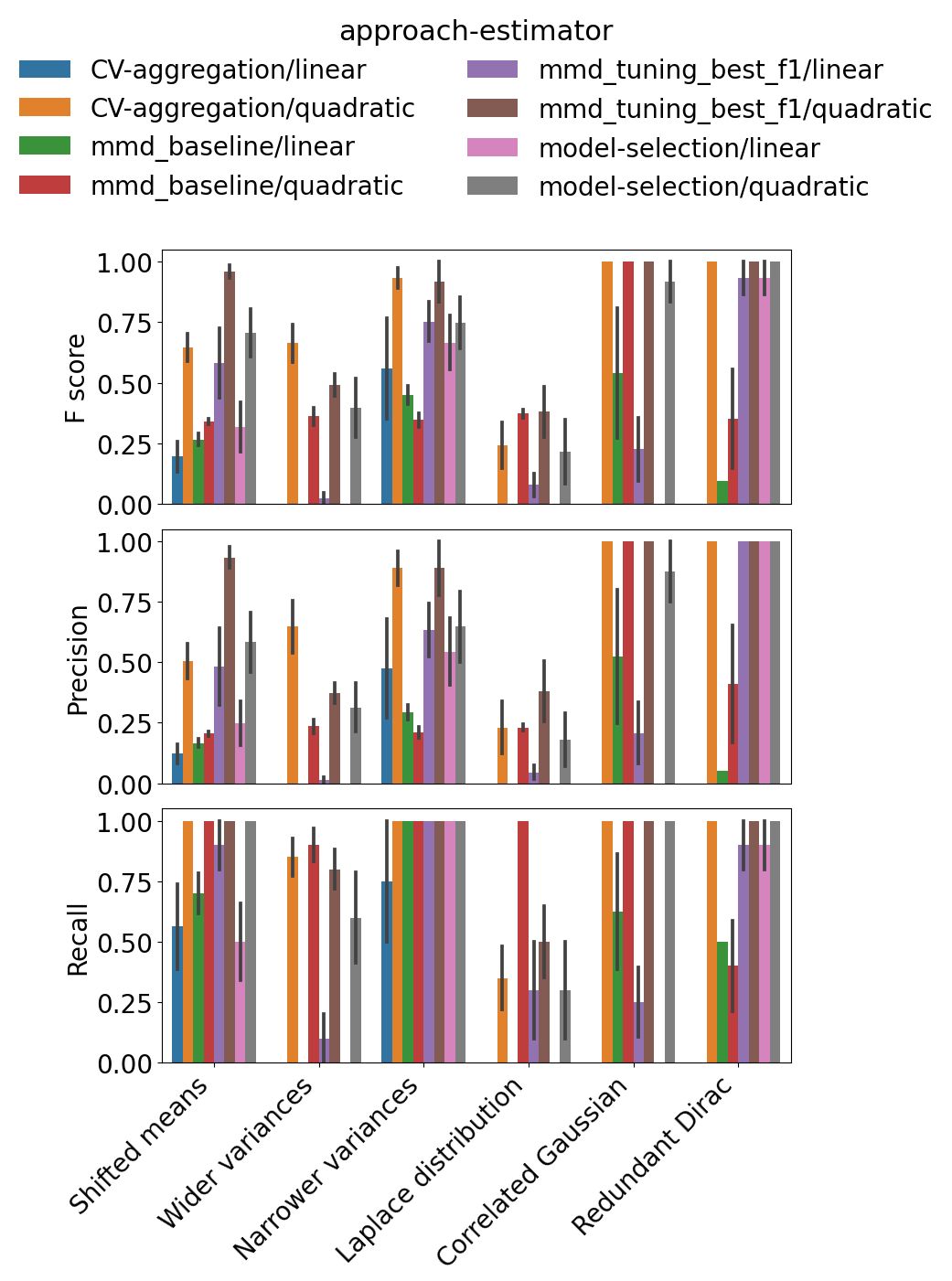}
      \subcaption{$\rho=0.1$}
    \end{minipage}
    \hfill
    \begin{minipage}[t]{0.49\hsize}
      \centering
      \includegraphics[keepaspectratio, scale=0.31]{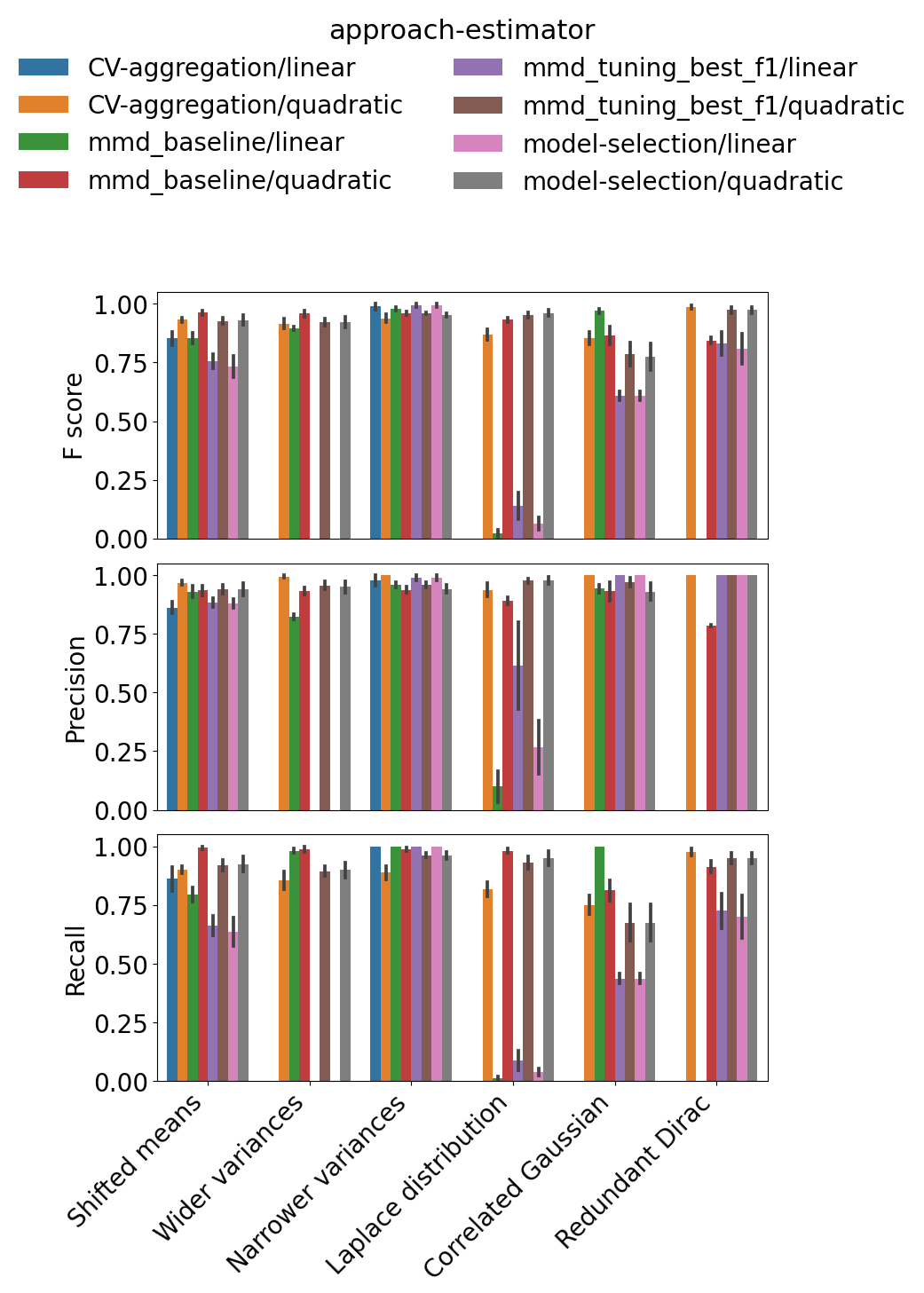}
      \subcaption{$\rho=0.8$}
    \end{minipage}
    \caption{
    Comparison of the linear and quadratic MMD estimators when used in the proposed approaches ({\tt CV-aggregation} and {\tt model-selection}) and other baselines ({\tt mmd-baseline} and {\tt mmd-tuning-best-f1}) in Appendix~\ref{sec:linear-estimator-app}. The left and right figures are the results for $\rho = 0.1$ and $\rho = 0.8$, respectively. The top, middle and bottom rows show the F score, Precision and Recall for each method and each setting of distributions $P$ and $Q$, with the means and standard deviations obtained from 10 independently repeated experiments.  Note that the linear estimator often leads to the complete failure of variable detection, in which case no bar is shown. 
    }
    \label{fig:linear-estimator-comparison}
\end{figure*}

We perform the same experiments as Section~\ref{sec:synthetic_data_assessment} and make a comparison with the proposed methods using the linear estimator. Figure~\ref{fig:linear-estimator-comparison} shows the results of the F score, Precision and Recall. For all the settings of distributions $P$ and $Q$, the evaluation scores for the linear estimator are significantly lower than the quadratic estimator. The significant drop in the evaluation scores occurs even for the rate $\rho = 0.8$ of ground-truth variables, which is easier than $\rho = 0.1$. Specifically, in the ``Wider variances'' setting, the scores for the linear estimator are consistently low, approaching $0.0$. 
For {\tt mmd-tuning-best-f1}, the linear estimator consistently performs worse than the quadratic estimator. Since {\tt mmd-tuning-best-f1} provides the highest possible F score for {\tt model-selection}, the results indicate that the linear estimator is not suitable for accurate variable selection, when the sample size $n$ is not too large so that the quadratic estimator is applicable.

\section{Alternative MMD-based Objective}
\label{sec:mmd-based-optimization}

\begin{figure*}[t]
  \begin{minipage}[t]{0.45\textwidth}
    \centering
    \includegraphics[width=80mm]{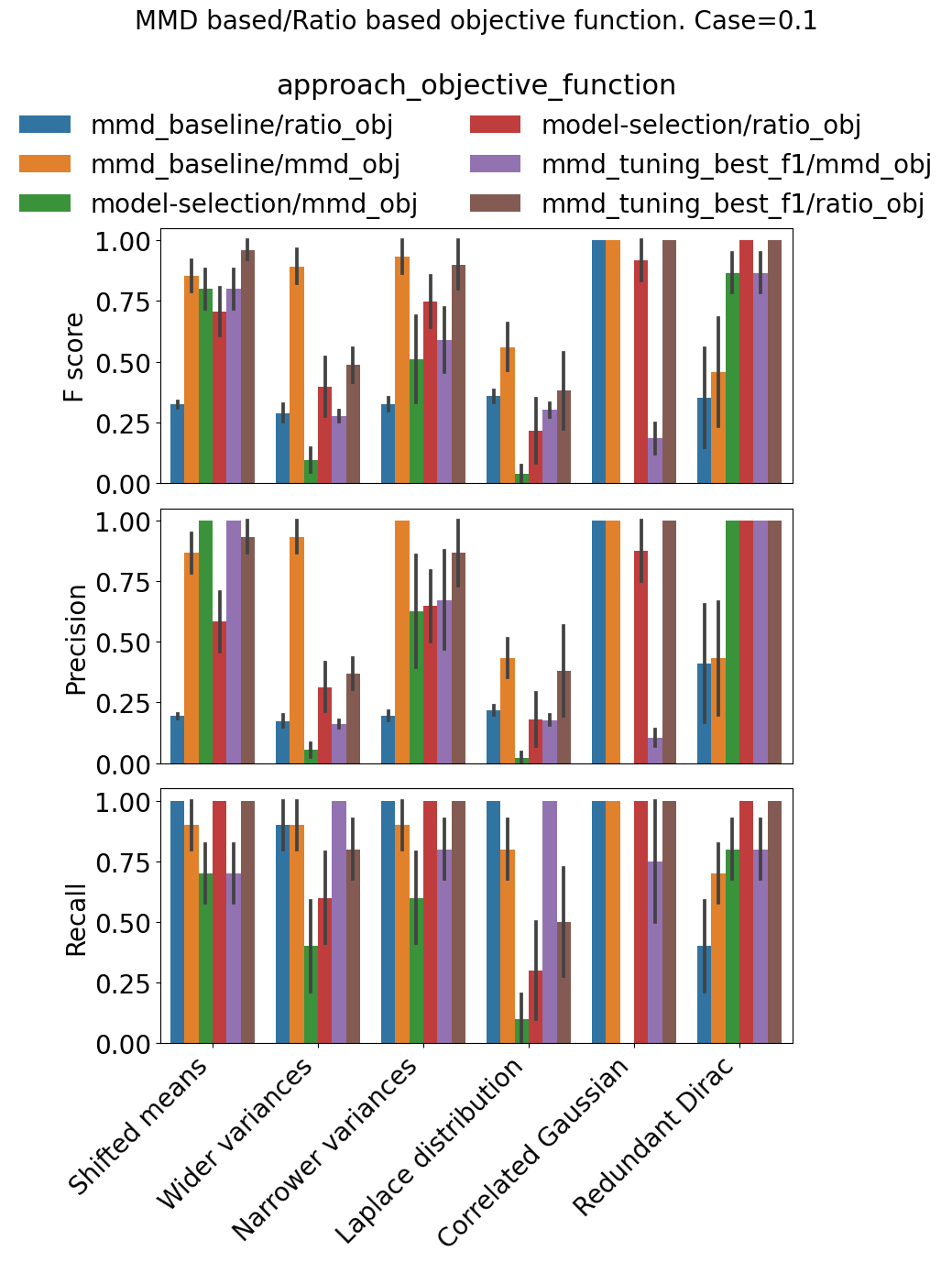}
  \end{minipage}%
  \begin{minipage}[t]{0.45\textwidth}
    \centering
    \includegraphics[width=80mm]{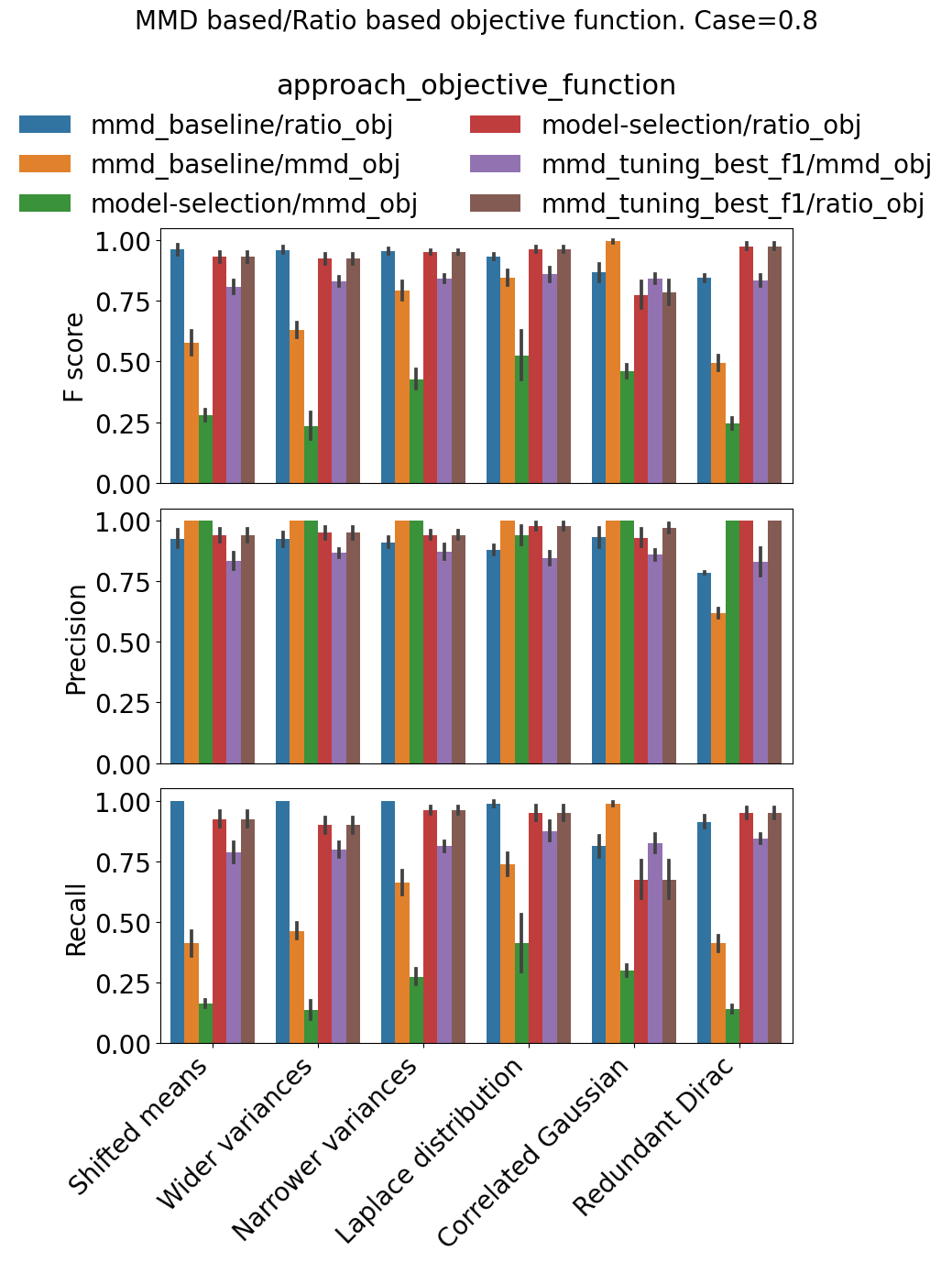}
  \end{minipage}
  \caption{Results of comparative experiments between the ratio-based objective ({\tt ratio\_obj}) and the MMD objective-based objective ({\tt mmd\_obj}) in Appendix \ref{sec:mmd-based-optimization}. The left figure describes the results for $\rho = 0.1$ and the right for $\rho = 0.8$. Note that the evaluation scores of {\tt model\_selection/mmd\_obj} for ``Correlated Gaussian'' with $\rho = 0.1$ are all zero, hence no bars.}
  \label{fig:mmd-and-test-power-based-optimizations-f-and-recall}
\end{figure*}

\begin{figure}[t]
  \centering
  \includegraphics[width=80mm]{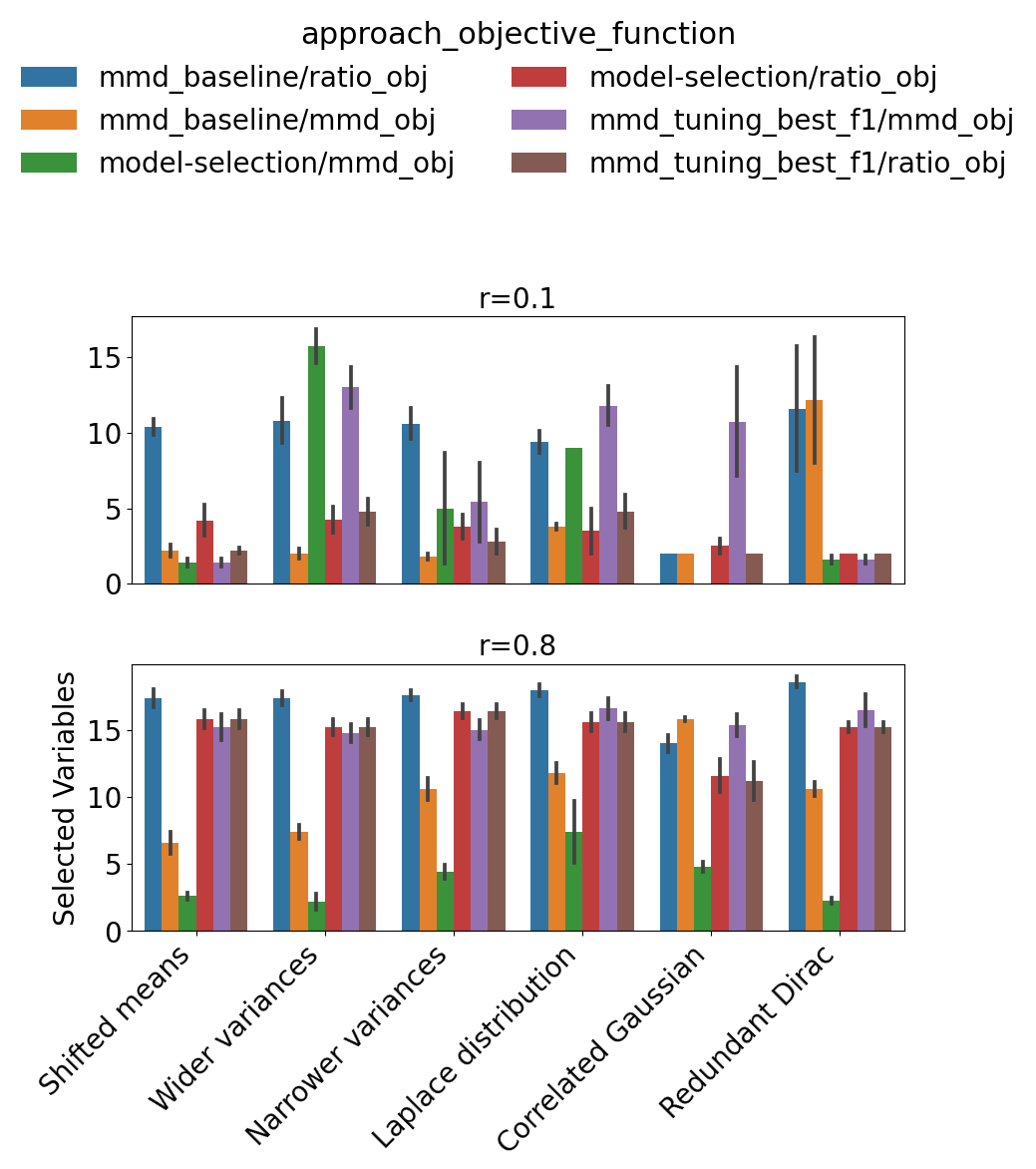}
  \caption{
  The number of selected variables in the comparative experiments between the ratio-based and MMD-based objective functions in Appendix~\ref{sec:mmd-based-optimization}. The top figure shows the results with $\rho = 0.1$, where the number of ground-truth variables is $20 \times 0.1 = 2$. The bottom figure is for $\rho = 0.8$, where the true number is $20 \times 0.8 = 16$.}
  \label{fig:mmd-and-test-power-based-optimizations-precision-n-variables}
\end{figure}

Here, we compare the proposed objective function \eqref{eq:regularized-opt} for optimising the ARD weights with an alternative objective function where the ratio objective \eqref{eq:430} is replaced by an unbiased MMD estimate $\widehat{\rm MMD}^{2}_U( {\bf X}, {\bf Y} )$ in \eqref{eq:mmd-unbiased-est}, i.e., 
\begin{equation} \label{eq:MMD-objective-app}
    \min_{ a \in \mathbb{R}^D } - \log \widehat{\rm MMD}^{2}_U( {\bf X}, {\bf Y} ) +  \lambda \sum_{d=1}^{D} |a_{d}|
\end{equation}
Historically, before Gretton et al.~\cite{gretton2012optimal} proposed the ratio objective as an objective function for hyper-parameter tuning in the MMD, Sriperumbudur et al.~\cite{kernel-choise} discussed that one could use the MMD itself as an objective function for hyper-parameter tuning of MMD. The latter approach optimises kernel parameters to maximise the MMD. It corresponds to removing the denominator in the ratio objective \eqref{eq:430}, thus resulting in \eqref{eq:MMD-objective-app}. Gretton et al.~~\cite{gretton2012optimal} showed that maximising the ratio objective yields higher test power than the MMD maximisation. We will investigate whether this property translates to the variable selection task, i.e., whether the ratio-based objective \eqref{eq:regularized-opt} leads to better variable selection performance than the MMD-based objective \eqref{eq:MMD-objective-app}.

We perform the same experiments as in Section~\ref{sec:synthetic_data_assessment} for both $\rho = 0.1$ and $\rho = 0.8$ (the constant $\rho$ represents the ratio of the number of ground-truth variables over the total number of variables). Here, we compare the performance of Algorithm \eqref{alg-hyperparameter_selection} ({\tt model-selection}), {\tt mmd-baseline} and {\tt mmd-tuning-best-f1}, which use the ratio-based objective \eqref{eq:regularized-opt}, and their MMD-based objective versions using \eqref{eq:MMD-objective-app}.

Figure ~\ref{fig:mmd-and-test-power-based-optimizations-f-and-recall}  describes the Precision, Recall and F scores for each setting and each method, obtained from 10 independently repeated experiments. 
The MMD-objective versions tend to yield lower values of the evaluation criteria than the ratio-objective versions. In particular, the former yields significantly lower Recall than the latter for $\rho = 0.8$. While the MMD-based objective yields better Precision for some settings, this improvement in Precision is enabled by sacrificing the Recall. 
Figure~\ref{fig:mmd-and-test-power-based-optimizations-precision-n-variables} shows the number of selected variables for each method and each setting. It is evident that the MMD objective results in much fewer variables than the ground-truth variables for $\rho = 0.8$.  
Based on these observations, we conclude that the ratio objective is more appropriate than the MMD objective for variable selection in two-sample testing.

\end{document}